\documentclass[10pt,journal,compsoc]{IEEEtran}
\usepackage{times}
\usepackage{epsfig}
\usepackage{enumitem}
\usepackage{graphicx}
\usepackage{amsmath}
\usepackage{amssymb}
\usepackage{multirow}
\usepackage{colortbl}
\usepackage{color}
\usepackage{threeparttable}
\usepackage{pifont}
\usepackage{booktabs}
\usepackage{adjustbox}
\usepackage{subfig}
\usepackage{caption}
\usepackage[misc]{ifsym}
\usepackage{xcolor,colortbl}
\usepackage{makecell}
\usepackage{tikz}
\usepackage[edges]{forest}
\usepackage{longtable}
\usepackage{mathrsfs}
\usepackage{xltabular}
\usepackage[hyphens]{url}

\newcommand*\circled[1]{\tikz[baseline=(char.base)]{
            \node[shape=circle,fill=black,text=white,draw,inner sep=0.5pt] (char) {#1};}}

\newcommand*{\belowrulesepcolor}[1]{%
  \noalign{%
    \kern-\belowrulesep 
    \begingroup 
      \color{#1}%
      \hrule height\belowrulesep 
    \endgroup 
    \vspace{-0.03mm}
  }%
} 
\newcommand*{\aboverulesepcolor}[1]{%
  \noalign{%
  \vspace{-0.03mm}
    \begingroup 
      \color{#1}%
    \endgroup 
    \kern-\aboverulesep 
  }%
}
\newcommand{\vspacefigtext}{\vspace{-3mm}}

\newcommand{\contentawarebig}{Content-Aware Editing\xspace}
\newcommand{\contentawareuppercase}{Content-aware editing\xspace}
\newcommand{\contentfreebig}{Content-Free Editing\xspace}
\newcommand{\contentfreeuppercase}{Content-free editing\xspace}
\newcommand{\contentawaresmall}{content-aware editing\xspace}
\newcommand{\contentawareshortsmall}{content-aware\xspace}
\newcommand{\contentfreesmall}{content-free editing\xspace}

\newcommand{\editingalgorithmsmall}{editing\xspace}
\newcommand{\editingalgorithmbig}{Editing\xspace}

\newcommand{\inversionfeaturesuppercase}{Inversion clue\xspace}
\newcommand{\contentawareinjectionbig}{Content-Aware Injection}
\newcommand{\contentfreeinjectionbig}{Content-Free Injection}

\newcommand{\texturalspaceadapterbig}{Textual Space Adapter}
\newcommand{\imagespaceadapterbig}{Latent Space Adapter}

\newcommand{\concat}{~+~}

\newcommand{\objectmanipulationshort}{TOM}
\newcommand{\attributemanipulationshort}{TAM}

\newcommand{\spatialtransformationshort}{TST}
\newcommand{\inpaintingshort}{TI}

\newcommand{\stylechangeshort}{TSC}
\newcommand{\imagetranslationshort}{TIT}

\newcommand{\subjectcustomizationshort}{TSDC}
\newcommand{\attributecustomizationshort}{TADC}

\newcommand{\gimageshort}{GNI}
\newcommand{\gmaskshort}{GM}
\newcommand{\gtextshort}{GST}
\newcommand{\ginstructionshort}{GI}
\newcommand{\guishort}{GUI}

\newcommand{\tuningshort}{IT}
\newcommand{\forwardspaceshort}{IF}

\newcommand{\normalshort}{EN}
\newcommand{\attentionshort}{EA}
\newcommand{\blendingshort}{EB}
\newcommand{\scorebasedshort}{ES}
\newcommand{\optimizationshort}{EO}

\newcommand{\imageconcatenation}{SIC}
\newcommand{\latentblending}{SLI}
\newcommand{\imageadapter}{SIA}
\newcommand{\texturalspaceadapter}{STSA}
\newcommand{\latentspaceadapter}{SLSA}

\newcommand{\imageconcatenationshort}{SIC}
\newcommand{\latentblendingshort}{SLB}
\newcommand{\imageadaptershort}{SIA}
\newcommand{\texturalspaceadaptershort}{STSA}
\newcommand{\latentspaceadaptershort}{SLSA}

\newcommand{\mycellcolor}{\cellcolor{gray!10}}

\definecolor{figorange}{RGB}{228,130,47}
\definecolor{figred}{RGB}{255,0,0}
\definecolor{figgreen}{RGB}{0,176,80}

\usepackage{xspace}
\usepackage{arydshln}

\makeatletter
\DeclareRobustCommand\onedot{\futurelet\@let@token\@onedot}
\def\@onedot{\ifx\@let@token.\else.\null\fi\xspace}
\def\eg{\emph{e.g}\onedot} 
\def\ie{\emph{i.e}\onedot}

\usepackage[linesnumbered,lined,boxed,commentsnumbered,ruled]{algorithm2e}
\usepackage[pagebackref,breaklinks=true,colorlinks,citecolor=green,urlcolor=blue,linkcolor=blue,bookmarks=false]{hyperref}

\ifCLASSOPTIONcompsoc
  \usepackage[nocompress]{cite}
\else
  \usepackage{cite}
\fi
\ifCLASSINFOpdf
\else
\fi
\begin{document}

\title{A Survey of Multimodal-Guided Image Editing with Text-to-Image Diffusion Models}

\author{Xincheng Shuai, Henghui Ding, Xingjun Ma, Rongcheng Tu,\\ Yu-Gang Jiang,~\IEEEmembership{Fellow,~IEEE}, Dacheng Tao,~\IEEEmembership{Fellow,~IEEE}
\IEEEcompsocitemizethanks{\IEEEcompsocthanksitem Xincheng Shuai, Henghui Ding, Xingjun Ma, and Yu-Gang Jiang are with Fudan University, Shanghai, China. E-mail: henghui.ding@gmail.com

\IEEEcompsocthanksitem Rongcheng Tu and Dacheng Tao are with Nanyang Technological University, Singapore.
}
}

\markboth{Journal of \LaTeX\ Class Files,~Vol.~14, No.~8, August~2015}%
{Shell \MakeLowercase{\textit{et al.}}: Bare Advanced Demo of IEEEtran.cls for IEEE Computer Society Journals}

\IEEEtitleabstractindextext{
\begin{abstract}
Image editing aims to edit the given synthetic or real image to meet the specific requirements from users. It is widely studied in recent years as a promising and challenging field of Artificial Intelligence Generative Content (AIGC). Recent significant advancement in this field is based on the development of text-to-image (T2I) diffusion models, which generate images according to text prompts. These models demonstrate remarkable generative capabilities and have become widely used tools for image editing. T2I-based image editing methods significantly enhance editing performance and offer a user-friendly interface for modifying content guided by multimodal inputs. In this survey, we provide a comprehensive review of multimodal-guided image editing techniques that leverage T2I diffusion models. First, we define the scope of image editing from a holistic perspective and detail various control signals and editing scenarios. We then propose a unified framework to formalize the editing process, categorizing it into two primary algorithm families. This framework offers a design space for users to achieve specific goals. Subsequently, we present an in-depth analysis of each component within this framework, examining the characteristics and applicable scenarios of different combinations. Given that training-based methods learn to directly map the source image to target one under user guidance, we discuss them separately, and introduce injection schemes of source image in different scenarios. Additionally, we review the application of 2D techniques to video editing, highlighting solutions for inter-frame inconsistency. Finally, we discuss open challenges in the field and suggest potential future research directions. We keep tracing related works at \url{https://github.com/xinchengshuai/Awesome-Image-Editing}.

\end{abstract}
}

\maketitle

\IEEEdisplaynontitleabstractindextext

\IEEEpeerreviewmaketitle

\section{Introduction}
\label{sec:introduction}
\IEEEPARstart{W}{ith} the development of cross-modal datasets~\cite{LAION-5B,LAION-400M,RedCaps,Wit,Conceptual-Captions,MeViS,YFCC100M} and generative frameworks~\cite{GAN,VAE,FM,EBM,DDPM}, emerging large-scale text-to-image (T2I) models~\cite{StableDiffusion,Imagen,DALLE-2} enable human beings to create desired images, leading to Artificial Intelligence Generative Content (AIGC) era in computer vision. Most of these works are based on diffusion models~\cite{DDPM}, a popular generative framework that is widely studied. Recently, numerous works explore the applications of these diffusion-based models in other fields, like image editing~\cite{PnP,MasaCtrl,pix2pix-zero,Imagic,Anydoor,RIE}, 3D generation / editing~\cite{DreamFusion,GaussianEditor,InstructNeRF2NeRF}, video generation / editing~\cite{VDM,Imagen-Video,FLATTEN,StyleCrafter} and so on. Unlike image generation, editing aims for secondary creation, which modifies desired elements in source image and retains semantic-unrelated contents. 
There is room for further improvement in quality and applicability, making editing still a promising and challenging task. 
In this work, we provide a comprehensive review of multimodal-guided image editing techniques that leverage T2I diffusion models.

\begin{figure*}[!t]
	\centering
    \includegraphics[width=1\linewidth]{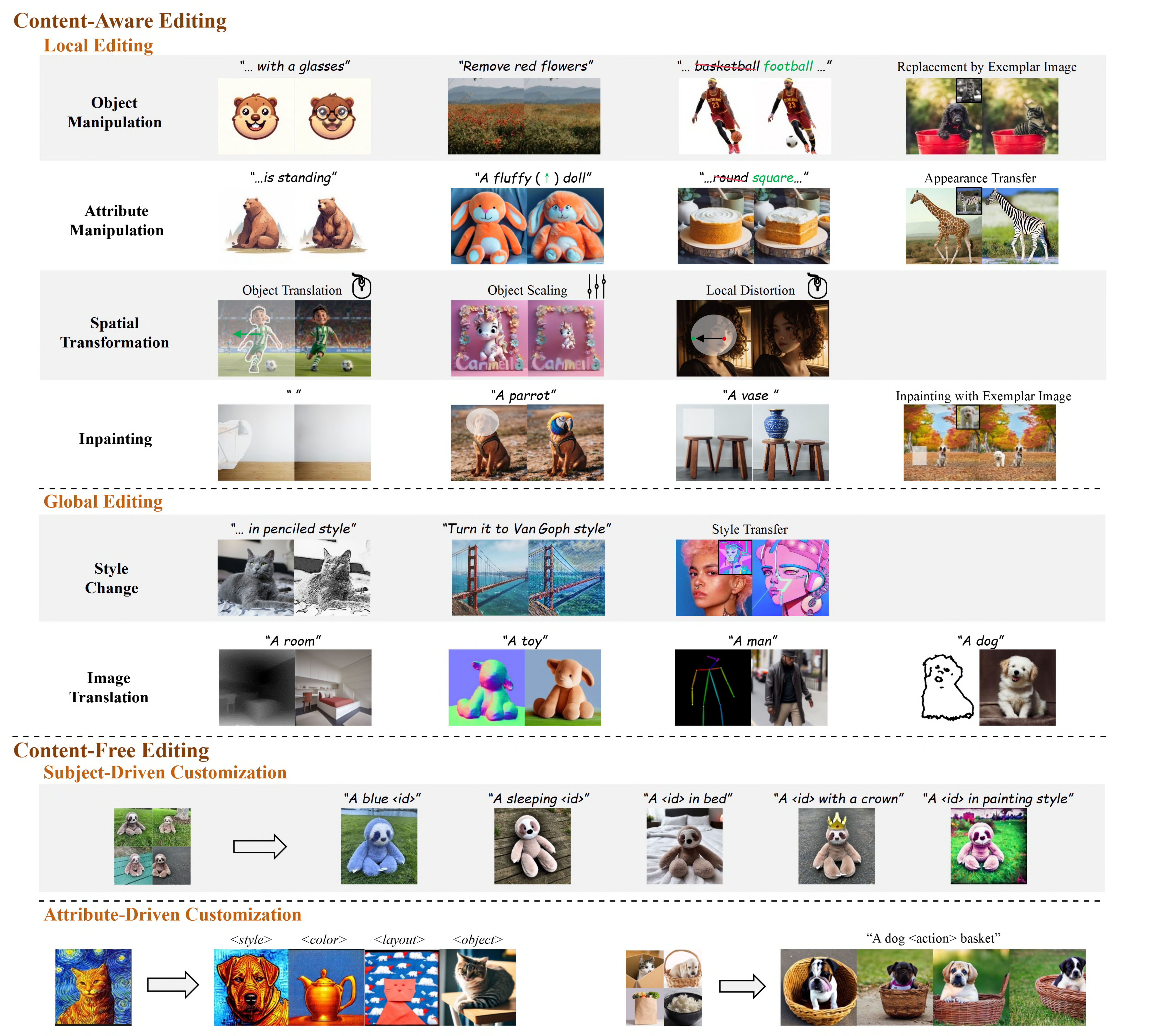}
    \vspace{-20pt}
    \caption{\textbf{Editing tasks meeting our definition}. We categorize editing tasks into content-aware and content-free groups, and enumerate several source-target pairs along with corresponding control signals for each scenario. The sample images are from~\cite{Inversion-Free,Ledits++,Paint-by-Example,P2P,Cross-Image-Attention,DesignEdit,Drag-Diffusion,InST,ControlNet,DreamBooth,MATTE,Reversion}.}
	\label{fig:editing-tasks}
 
\end{figure*}

\newcommand{\taxonomycaption}{Multimodal-Guided Image Editing with T2I Diffusion Models}
\tikzstyle{node-cfg}=[
    rectangle,
    draw=brown!60!black,
    sharp corners,
    text opacity=1,
    inner sep=3pt, 
    text=black,
    fill=yellow!5,
    fill opacity=.4, 
    line width=1pt,
    font=\tiny,
    align=left
]
\begin{figure*}[t!]
    \centering
    \resizebox{1.0\textwidth}{!}{
        \begin{forest}
            forked edges,
            for tree={
                grow=east,
                reversed=true,
                anchor=base west,
                parent anchor=east,
                child anchor=west,
                base=left,
                font=\tiny,
                rectangle,
                draw=brown!60!black,
                sharp corners,
                align=left,
                minimum width=2em,
                edge+={darkgray, line width=1pt},
                s sep=2.8pt,
                inner xsep=1.8pt,
                inner ysep=2.7pt,
                line width=0.7pt,
                par/.style={rotate=90, child anchor=north, parent anchor=south, anchor=center},
            },
            where level=1{text width=5.8em,font=\tiny,}{},
            where level=2{text width=4.9em,font=\tiny,}{},
            [
                \taxonomycaption, par
                [
                    Inversion Algorithm (\S~\ref{sec:inversion algorithms})
                    [
                        Tuning-Based \\Inversion (\S~\ref{sec:tuning-based inversion space})
                        [
                           \textbf{(\romannumeral 1) Textual Space:} TI~\cite{TI} {,} DreamArtist ~\cite{DreamArtist} {,} 
                           UniTune~\cite{UniTune} {,} HiPer~\cite{HiPer} {,} XTI~\cite{P+} {,} NeTI~\cite{NeTI} {,}  ProSpect~\cite{Prospect} {,}  \\ MATTE~\cite{MATTE} {,}  Concept Decomposition~\cite{Concept-Decomposition} {,} Reversion~\cite{Reversion} {,}  ADI~\cite{ADI} {,}  Lego~\cite{Lego} 
                           \\
                           \textbf{(\romannumeral 2) Model Space:} DreamBooth~\cite{DreamBooth} {,} He \emph{et al.}~\cite{Data-Perspective} {,} DCO~\cite{DCO} {,}  FaceChain-SuDe~\cite{Facechain-Sude} {,} CustomDiffusion\cite{Custom-Diffusion} {,} \\ 
                           Cones2~\cite{Cones2} {,} Cones~\cite{Cones} {,}  SVDiff~\cite{Svdiff} {,} LoRA~\cite{Lora} {,}  ANOVA~\cite{ANOVA} {,} CatVersion~\cite{CatVersion}{,} Break-a-Scene~\cite{Break-A-Scene} {,} Clic~\cite{Clic}
                           \\
                           \textbf{(\romannumeral 3) Image Adapter Space:} InST~\cite{InST} {,}  DisenBooth~\cite{Disenbooth} {,} Cai \emph{et al.}~\cite{Decoupled-Textual-Embeddings} {,} ViCo~\cite{VICO}
                           , node-cfg, text width=25.9em
                        ]             
					]
                    [
                        Forward-Based \\ Inversion (\S~\ref{sec:foward-based inversion space})
                        [
                           \textbf{(\romannumeral 1) DDIM Inversion:} NTI~\cite{NTI} {,} PTI~\cite{PTI} {,} StyleDiffusion~\cite{StyleDiffusion} {,} KV Inversion~\cite{KV-Inversion} {,} NPI~\cite{NPI} {,} ProxEdit~\cite{ProxEdit}{,} \\FPI~\cite{FPI} {,} AIDI~\cite{AIDI} {,} EDICT~\cite{EDICT} {,} BDIA~\cite{BDIA} {,} PnP Inversion~\cite{PnP-Inversion} {,} pix2pix-zero~\cite{pix2pix-zero} {,} ReGeneration~\cite{ReGeneration} {,} DPL~\cite{DPL}
                           \\
                           \textbf{(\romannumeral 2) DDPM Inversion:} CycleDiffusion~\cite{CycleDiffusion} {,} DDPM Inversion~\cite{Edit-Friendly}
                            , node-cfg, text width=25.9em
                        ]       
                    ]
				]
				[
                    \editingalgorithmbig Algorithm (\S~\ref{sec:editing-algorithm})
                    [
                        Attention-Based \\ \editingalgorithmbig Algorithm\\ (\S~\ref{sec:attention-based image editing})
                        [
                           \textbf{(\romannumeral 1) Manipulation in Cross-Attention Mechanism:} P2P~\cite{P2P} {,} Custom-Edit~\cite{Custom-Edit}{,} Object-Shape Variation~\cite{Localize-Object-Shape} {,} FoI~\cite{FOI}
                           \\
                       \textbf{(\romannumeral 2) Manipulation in Self-Attention Mechanism:}  PnP~\cite{PnP} {,} FPE~\cite{FPE} {,} PhotoSwap~\cite{PhotoSwap} {,} DreamMatcher~\cite{DreamMatcher} {,} \\ TF-ICON~\cite{TF-ICON} {,} HD-Painter~\cite{HD-Painter} {,} DesignEdit~\cite{DesignEdit} {,} MasaCtrl~\cite{MasaCtrl} {,} TIC~\cite{TIC} {,} Cross-Image Attention~\cite{Cross-Image-Attention} {,} \\ StyleInjection~\cite{StyleInjection} {,}  Z\({}^{\mbox{*}}\)~\cite{Z-star} 
                            , node-cfg, text width=25.9em
                        ] 
                    ]
                    [
                        Blending-Based \\ \editingalgorithmbig Algorithm\\ (\S~\ref{sec:interpolation-based image editing})
                        [
                           \textbf{(\romannumeral 1) Blending in Spatial Space:} 
                           BLD~\cite{Blended-Latent-Diffusion} {,} High-Resolution Blended Diffusion~\cite{High-Resolution-Image-Editing}{,} Differential Diffusion~\cite{Differential-Diffusion} {,} \\ DreamEdit~\cite{OIR} {,} OIR~\cite{OIR} {,} DiffEdit~\cite{Diff-Edit} {,} Watch Your Steps~\cite{Watch-Your-Step} {,} Zone~\cite{Zone} {,}
                           PFB-Diff~\cite{Pfb-diff} {,} \\
                           Tuning-Free Image Customization~\cite{Tuning-Free-Image-Customization}
                           \\
                           \textbf{(\romannumeral 2) Blending in Semantic Space:}  Imagic~\cite{Imagic} {,} Forgedit~\cite{Forgedit} {,} PTI~\cite{PTI} {,} Wu~\emph{et al.}~\cite{Uncovering-Disentanglement} {,} DAC~~\cite{DAC}
                            , node-cfg, text width=25.9em
                        ]    
                    ]
                    [
                        Score-Based \editingalgorithmbig \\ Algorithm\\ (\S~\ref{sec:score-based image editing})
                        [
                           \textbf{(\romannumeral 1) Multi-Noise Guidance:} SEGA~\cite{SEGA} {,} LEDITS~\cite{Ledits}{,} LEDITS++~\cite{Ledits++} {,} Stable Artist~\cite{Stable-Artist}
                           \\
                           \textbf{(\romannumeral 2) Multi-Expert Guidance:}  SINE~\cite{SINE} {,} 
                           DCO~\cite{DCO} 
                           \\
                           \textbf{(\romannumeral 3) Energy Function Guidance:}  NMG~\cite{NMG}{,} Self-Guidance~\cite{Self-Guidance} {,} DragonDiffusion~\cite{Dragon-Diffusion} {,} MagicRemover~\cite{MagicRemover} {,} \\ FreeControl~\cite{FreeControl} {,} DiffEditor~\cite{DiffEditor}
                            , node-cfg, text width=25.9em
                        ]    
                    ]
                    [
                        Optimization-Based \\ \editingalgorithmbig Algorithm\\ (\S~\ref{sec:optimization-based image editing})
                        [
                           \textbf{(\romannumeral 1) Optimization under Image-Level Loss:} RDM~\cite{Region-Aware}
                           \\
                           \textbf{(\romannumeral 2) Optimization under Feature-Level Loss:}  DragDiffusion~\cite{Drag-Diffusion} {,} Pick-and-Draw~\cite{Pick-and-Draw} {,} EBM~\cite{EBM} 
                           \\
                           \textbf{(\romannumeral 3) Optimization under Score Distillation Loss:}  DDS~\cite{DDS} {,} CDS~\cite{CDS}  {,} Ground-a-Score~\cite{Ground-A-Score}
                            , node-cfg, text width=25.9em
                        ]    
                    ]
                ]
                [
                    Training-Based Image \\ Editing (\S~\ref{sec:end-to-end image editing})
                    [
                        Content-Aware Editing \\  (\S~\ref{sec:content-aware-editing})
                        [
                            \textbf{(\romannumeral 1) Instruction-Based Image Editing:} InstructPix2Pix~\cite{InstructPix2Pix}  {,} MagicBrush~\cite{MagicBrush}{,} HIVE~\cite{HIVE}  {,} InstructDiffusion~\cite{InstructDiffusion}{,} \\ Emu-Edit~\cite{Emu-Edit}  {,} MGIE~\cite{MGIE}{,} SmartEdit~\cite{SmartEdit} 
                           \\
                           \textbf{(\romannumeral 2) Image Inpainting:}  Imagen Editor~\cite{Imagen-Editor} {,} SmartBrush~\cite{SmartBrush} {,} PowerPaint~\cite{PowerPaint} {,} PbE~\cite{Paint-by-Example}  {,} ObjectStitch~\cite{ObjectStitch} {,} \\ RIC~\cite{Reference-Based-Composition} {,} PhD~\cite{PHD} {,} AnyDoor~\cite{Anydoor}
                           \\
                           \textbf{(\romannumeral 3) Image Translation:}  ControlNet~\cite{ControlNet} {,} T2I-Adapter~\cite{T2I-Adapter}  {,} SCEdit~\cite{SCEdit} {,} UniControl~\cite{UniControl} {,} Cocktaik~\cite{Cocktail} {,} \\ Uni-ControlNet~\cite{Uni-ControlNet} {,} CycleNet~\cite{CycleNet} {,} CycleGAN-Turbor~\cite{CycleGAN-Turbo}
                            , node-cfg, text width=25.9em
                        ]
                    ]
                    [
                        Content-Free Editing \\ (\S~\ref{sec:content-free-editing})
                        [
                            \textbf{(\romannumeral 1) Subject-Driven Customization:} Taming~\cite{Taming}  {,} InstantBooth~\cite{InstantBooth} {,} E4T~\cite{E4T}  {,} ProFusion~\cite{Enhance-Detail} {,} \\ FastComposer~\cite{FastComposer} {,}  PhotoMaker~\cite{PhotoMaker} {,} Photoverse~\cite{Photoverse} {,} InstantID~\cite{InstantID} {,} ELITE~\cite{ELITE} {,} BLIP-Diffusion~\cite{Blip-Diffusion} {,} \\ Domain-Agnostic~\cite{Domain-Agnostic} {,} UMM~\cite{UMM}  {,} SuTI~\cite{Subject-Driven-Diffusion} {,} Subject-Diffusion~\cite{Subject-Diffusion} {,} Instruct-Imagen~\cite{Instruct-Imagen}
                           \\
                           \textbf{(\romannumeral 2) Attribute-Driven Customization:}  ArtAdapter~\cite{ArtAdapter} {,}  DreamCreature~\cite{DreamCreature} {,} Lee \emph{et al.}~\cite{Language-Informed} {,} pOps~\cite{pops}
                            , node-cfg, text width=25.9em
                        ]
                    ]
                ]
                [
                    Extension in Video Editing \\ (\S~\ref{sec:video})
                    [
                        Utilization of Motion \\ Prior (\S~\ref{sec:motion-prior})
                        [
                            Dreamix~\cite{Dreamix} {,}  AnyV2V~\cite{AnyV2V} {,} Gen-1~\cite{Gen-1} {,}  FlowVid~\cite{FlowVid} 
                            , node-cfg, text width=25.9em
                        ]
                    ]
                    [
                       Cross-Frame Attention \\  (\S~\ref{sec:cross-frame})
                       [
                           TAV~\cite{Tune-a-Video} {,} EAV~\cite{Edit-A-Video} {,}  Video-P2P~\cite{Video-P2P} {,}  FateZero~\cite{FateZero}
                           , node-cfg, text width=25.9em
                       ]
                    ]
                    [
                       Propagation of Image \\ Information (\S~\ref{sec:propagation}) 
                       [
                            TEXT2LIVE~\cite{Text2LIVE} {,} VidEdit~\cite{VidEdit}  {,} StableVideo~\cite{StableVideo} {,} Shape-Aware NLA~\cite{Shape-Aware-NLA}{,} DiffusionAtlas~\cite{DiffusionAtlas} {,} \\ TokenFlow~\cite{TokenFlow} {,}  Fairy~\cite{Fairy} {,} VidToMe~\cite{VidToMe} {,} FLATTEN~\cite{FLATTEN}
                            , node-cfg, text width=25.9em
                        ]
                    ]
                ]
        ]       
        \end{forest}
        }
    \caption{\textbf{Organization of the survey}.}
    \label{fig:organization}
\end{figure*}
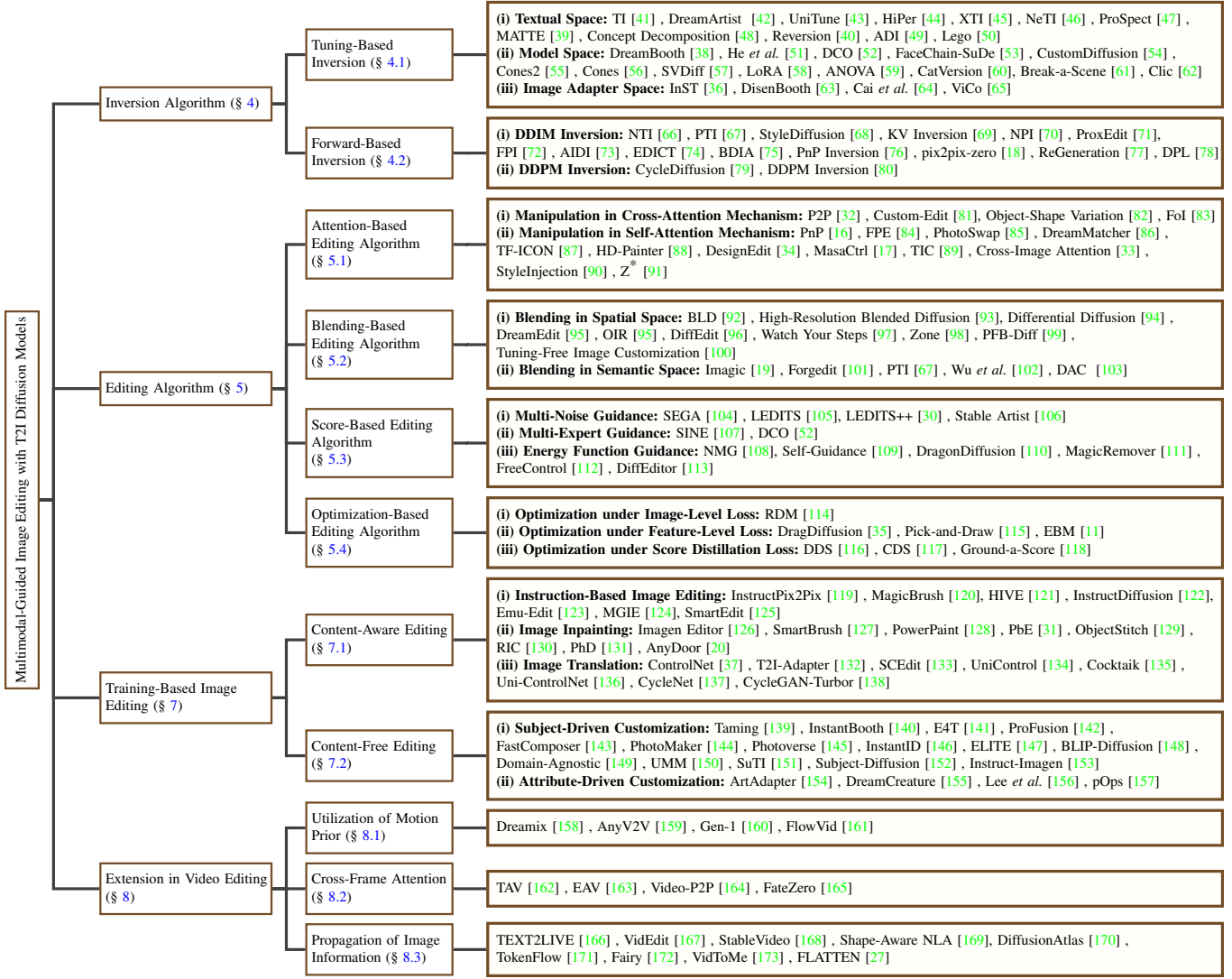

There are surveys~\cite{comprehensive-survey1,controllable-generation-survey,super-resolution-survey,medical-survey,editing-survey} reviewing state-of-the-art diffusion-based methods from different aspects, such as image restoration~\cite{restoration-survey}, super-resolution~\cite{super-resolution-survey}, medical image analysis~\cite{medical-survey}, \emph{etc}. Compared with these surveys, we focus on techniques in the field of image editing. Two concurrent works~\cite{editing-survey,controllable-generation-survey} are related to our survey. Among them, ~\cite{editing-survey} introduces the application of diffusion models in image editing and categories relevant papers based on their learning strategies. Compared with it, we discuss the topic from a novel and holistic perspective, and propose a unified framework to formalize the editing process. We find that the interpretation of editing is limited and incomplete in previous literature~\cite{PnP,editing-survey,NTI,P2P}. These works restrict the scope of preserved concepts and intend to reconstruct maximal amount of details from source image. However, this common setting excludes the maintenance of some high-level semantics, like identity, style and so on. To address this issue, we first provide a rigorous and comprehensive definition of editing, and include more relevant studies like~\cite{DreamBooth,Break-A-Scene,ControlNet,InstantID} in this survey. \figurename~\ref{fig:editing-tasks} illustrates various scenarios meeting our definition. It is worth noting that some generation tasks like customization~\cite{TI,Custom-Diffusion} and conditional generation with image guidance~\cite{ControlNet,UniControl} all conform to our discussion scope. These tasks are discussed in another concurrent work~\cite{controllable-generation-survey} that focuses on controllable generation. Secondly, we integrate reviewed methods into an unified framework, which separates the editing process into two algorithm families, \ie, inversion and editing algorithms. In~\cite{editing-survey}, a similar framework is introduced to unify the methods that do not need training or test-time fine-tuning. Differently, our framework is more versatile for the generalized editing scenarios in discussion. Meanwhile, the framework provides a design space for users to combine proper techniques depending on their specific purposes. Experiments in the survey demonstrate the characteristics of different combinations along with their applicable scenarios. Furthermore, we also investigate the extension of 2D approaches~\cite{P2P,SDEdit} in video editing\cite{FateZero,VidToMe} and concentrate on their solutions of temporal inconsistency, completing the missing part in research area.

We conduct an extensive survey of over three hundred papers, and examine the essence and internal logic of existing methods. This survey mainly focuses on studies based on T2I diffusion models~\cite{StableDiffusion,Imagen,StableDiffusionXL}. In Section~\ref{sec:preliminaries}, diffusion model and techniques in T2I generation are introduced, offering a basic theoretical background. In Section~\ref{sec:problem-formulation}, we give the definition of image editing and discuss several important aspects, such as user guidance of different modalities, editing scenarios, and some qualitative and quantitative metrics for evaluation. Meanwhile, we formalize the proposed unified framework to integrate existing methods. Next, the major components of our framework are discussed in Section~\ref{sec:inversion algorithms}
and Section~\ref{sec:editing-algorithm} respectively. Inversion algorithm captures concepts to be preserved from source images, while \editingalgorithmsmall algorithm aims to reproduce visual elements under user guidance, achieving both content consistency and semantic fidelity. In Section~\ref{sec:design-space}, we examine the different combinations of inversion and \editingalgorithmsmall algorithms, and investigate their characteristics and applicable scenarios, thereby guiding users to choose proper methods for distinct targets. Since training-based approaches~\cite{Anydoor,IP-Adapter,InstructDiffusion,InstructPix2Pix} learn to directly transform the source image to target one, we discuss these works in Section~\ref{sec:end-to-end image editing} and elaborate injection schemes of source image in different tasks. Section~\ref{sec:video} introduces the extension of image editing in video domain. Due to the scarcity of video data, directly applying image-domain methods often leads to inter-frame inconsistencies. This section discusses several solutions in existing works~\cite{TokenFlow,Dreamix,Video-P2P,Text2LIVE}. Finally, we discuss the unresolved challenges in Section~\ref{sec:future}, offering potential future research directions. \figurename~\ref{fig:organization} demonstrates the organization of our work and categorizes reviewed papers in each section.

\section{Preliminaries}
Herein we introduce some fundamental aspects of diffusion models and text-to-image generation, providing a basic theoretical
background to facilitate the understanding of the survey.
\label{sec:preliminaries}
\subsection{Denoising Diffusion Probabilistic Models}
Recently, Denoising Diffusion Probabilistic Models (DDPMs) \cite{DDPM} have been widely explored in image generation and related tasks~\cite{StableDiffusion,MasaCtrl,zhao2022closer,SegRefiner,InstructNeRF2NeRF,NTI,InstructPix2Pix,Text2LIVE} because of the powerful capability of modeling complicated distribution. DDPMs consist of two $T$-step Markov processes. From $0$ to $T$, forward process iteratively introduces independent Gaussian noise into image. As the number of iterations approaches $T$, the noisy image becomes to pure Gaussian noise. Reversely, backward process transforms the noise back to clean image. From $T$ to $0$, diffusion models predict the injected noise in current step and denoise to infer the next latent.

\noindent$\bullet$
\textbf{Forward \& Backward Processes in DDPM}. Given the input image $\mathbf{z}_0$ and step $1 \leq t \leq T$, the forward process of DDPM $F_{fw}^{DP}$ is formalized as:
\begin{equation}
\begin{aligned}\label{eq:ddpm-forward-step}
\mathbf{z}_t&=F_{fw}^{DP}(\mathbf{z}_{t-1},t-1) \\
&=\sqrt{1-\beta_t} \mathbf{z}_{t-1}+\sqrt{\beta_t} \epsilon_t,
\end{aligned}
\end{equation}
where $\beta_t$ is the hyperparameter specified by variance scheduler, and $\epsilon_t \sim \mathcal{N}(0, \mathbf{I})$ is injected noise. Furthermore, $\mathbf{z}_t$ can also be directly derived from $\mathbf{z}_0$:
\begin{equation}
\begin{aligned}\label{eq:ddpm-forward}
\mathbf{z}_t=\sqrt{\bar{\alpha}_t} \mathbf{z}_0 +\sqrt{1-\bar{\alpha}_t} \epsilon_t^0,
\end{aligned}
\end{equation}
where $\bar{\alpha}_t=\prod_{i=1}^t \alpha_i$ and $\alpha_t=1-\beta_t$. $\epsilon_t^0$ is introduced Gaussian noise in $t$ step. Reversely, backward process  $F_{bw}^{DP}$ denoises the latent to clean image, which is formalized as:
\begin{equation}
\begin{aligned}\label{eq:ddpm-reverse}
\mathbf{z}_{t-1}&=F_{bw}^{DP}(\mathbf{z}_t,t)\\
&=\frac{1}{\sqrt{\alpha_t}}\left(\mathbf{z}_{t}-\frac{\beta_t}{\sqrt{1-\bar{\alpha}_t}} \varepsilon_\theta\left(\mathbf{z}_{t}, t\right)\right)+\sigma_t \hat{\epsilon}_t,
\end{aligned}
\end{equation}
where $\varepsilon_\theta\left(\mathbf{z}_{t}, t\right)$ is the estimation of injected noise, and $\theta$ represents network parameters. $\sigma_t$ and $\hat{\epsilon}_t$ are hyperparameter and stochastic component in backward process respectively.

For training of noise estimation network $\varepsilon_\theta$, the study~\cite{DDPM} proposes to minimize variational bound~\cite{VAE} of negative log-likelihood, where the goal is solving:
\begin{equation}
\begin{aligned}\label{eq:ddpm-loss}
\underset{\theta}{\arg \min } E_{ \mathbf{z}_0 \sim p_{data},t \sim \mathcal{U}(1, T), \epsilon \sim \mathcal{N}(\mathbf{0}, \mathbf{I})}\left[\lambda_t \left\|\varepsilon_{\theta}\left(\mathbf{z}_t, t\right)-\epsilon\right\|^2\right],
\end{aligned}
\end{equation}
where $p_{data}$ is the distribution of training data. $\mathbf{z}_t$ is obtained through Eq.~\ref{eq:ddpm-forward}, and $\lambda_t$ is time-variant weighting factor.

\noindent$\bullet$
\textbf{Inversion \& Sampling Processes in DDIM}. Denoising Diffusion Implicit Model (DDIM)~\cite{DDIM} accelerates sampling in DDPMs, where sampling process $F_{bw}^{DI}$ is given as:
\begin{equation}
\begin{aligned}\label{eq:ddim-reverse}
\mathbf{z}_{t-1}&=F_{bw}^{DI}(\mathbf{z_t},t)\\
&=\sqrt{\bar{\alpha}_{t-1}}\frac{\mathbf{z}_t-\sqrt{1-\bar{\alpha}_t}\varepsilon_\theta(\mathbf{z}_t,t)}{\sqrt{\bar{\alpha}_t}}+\sqrt{1-\bar{\alpha}_{t-1}}\varepsilon_\theta(\mathbf{z}_t,t).
\end{aligned}
\end{equation}
Rearranging Eq.~\ref{eq:ddim-reverse} derives the DDIM inversion process $F_{fw}^{DI}$:
\begin{equation}
\begin{aligned}\label{eq:ddim-forward inversion}
\mathbf{z}_{t}&=F_{fw}^{DI}(\mathbf{z}_{t-1},t-1)\\
&=\sqrt{\bar{\alpha}_{t}}\frac{\mathbf{z}_{t-1}-\sqrt{1-\bar{\alpha}_{t-1}}\varepsilon_\theta(\mathbf{z}_t,t)}{\sqrt{\bar{\alpha}_{t-1}}}+\sqrt{1-\bar{\alpha}_{t}}\varepsilon_\theta(\mathbf{z}_t,t),
\end{aligned}
\end{equation}
which is widely used in editing to invert real image to latent space.

\noindent$\bullet$
\textbf{Conditional Generation}. For controllable generation, classifier-guidance~\cite{classifier-free} introduces a noise-dependent classifier to steer the backward process. Classifier-free guidance~\cite{classifier-guidance} provides a solution without auxiliary classifier. It combines conditional noise estimate $\varepsilon_\theta(\mathbf{z}_t,t,\mathcal{C})$ with unconditional term $\varepsilon_\theta(\mathbf{z}_t,t,\varnothing)$ in backward process, where $\mathcal{C}$ and $\varnothing$ indicate condition and null condition respectively. The predicted noise in classifier-free guidance is expressed as:
\begin{equation}
\begin{aligned}\label{eq:classifier-free guidance}
\tilde{\varepsilon}_\theta\left(\mathbf{z}_t, t, \mathcal{C}, \varnothing, \omega \right)=\omega \cdot \varepsilon_\theta\left(\mathbf{z}_t, t, \mathcal{C}\right)+(1-\omega) \cdot \varepsilon_\theta\left(\mathbf{z}_t, t, \varnothing\right),
\end{aligned}
\end{equation}
where $\omega$ is guidance scale, and larger $\omega$ enhances the impact of condition $\mathcal{C}$ in final result.

\noindent$\bullet$
\textbf{Score Function}. From score-based perspective~\cite{SDE,NCSM}, diffusion processes~\cite{DDPM} can be generalized as representation of stochastic differential equation (SDE), which is formalized as.
\begin{equation}
\begin{aligned} \label{eq:sde-forward}
d \mathbf{z}=f(\mathbf{z}, t)+g(\mathbf{z})d \mathbf{w},
\end{aligned}
\end{equation}
where $f(\mathbf{z}, t)$ and $g(\mathbf{z})$ are drift coefficient and diffusion coefficient respectively, and $\mathbf{w}$ indicates Brownian motion. The reverse process of Eq.~\ref{eq:sde-forward} is given as:
\begin{equation}
\begin{aligned}\label{eq:sde-backward}
d \mathbf{z}=\left(f(\mathbf{z}, t)-g(\mathbf{z})^2 \nabla_\mathbf{z} \log p_t(\mathbf{z})\right) d t+g(\mathbf{z}) d \overline{\mathbf{w}},
\end{aligned}
\end{equation}
where $\nabla_\mathbf{z} \log p_t(\mathbf{z})$ is score function of $p_t(\mathbf{z})$, defined as gradient of the log of noisy marginal distribution probability. Furthermore, conditional score function is given as:
\begin{equation}
\begin{aligned} \label{eq:conditional-score-function-decomposition}
\nabla_{\mathbf{z}} \log p_t\left(\mathbf{z} \mid \mathcal{C}\right)=&\nabla_{\mathbf{z}} \log \left(\frac{p_t\left(\mathcal{C} \mid \mathbf{z}\right) p_t\left(\mathbf{z}\right)}{p(\mathcal{C})}\right) \\
&\propto \nabla_{\mathbf{z}} \log p_t\left(\mathbf{z}\right)+  \nabla_{\mathbf{z}} \log p_t\left(\mathcal{C} \mid \mathbf{z}\right),
\end{aligned}
\end{equation}
where $\log p_t\left(\mathcal{C} \mid \mathbf{z}\right)$ represents the posterior probability of $\mathcal{C}$.

According to related literature~\cite{classifier-free,EDM}, the estimated noise in DDPMs can be expressed as following equivalent form:
\begin{equation} \label{eq:noise-score-version}
\begin{aligned}
\varepsilon_\theta(\mathbf{z}_t,t,\mathcal{C})=-\sqrt{1-\bar{\alpha}_t} \nabla_{\mathbf{z}_t} \log p_t(\mathbf{z}_t \mid \mathcal{C}), 
\end{aligned}
\end{equation}
which facilitates the interconversion between score function and predicted noise.

\subsection{Text-to-Image Generation}
Recently, numerous large-scale models~\cite{StableDiffusion,Imagen,DALLE-2,StableDiffusionXL,GLIDE} are developed to achieve text-driven generation, which also empower other related arenas~\cite{Imagic,VDM,Drag-Diffusion,StableVideo,Magic3D,GRES,MVDream,VSD}. In order to facilitate the discussion, we introduce several fundamental aspects of T2I generation in following.

\noindent$\bullet$
\textbf{Text Encoder}. Text encoder~\cite{CLIP,T5XXL,BERT} is one of the indispensable components in current T2I models, which extracts expressive semantic features from text input. Among them, CLIP~\cite{CLIP} jointly trains text encoder and image encoder with 400 million text-image pairs to embed text and image in a multimodal space. For large language models (LLMs), T5~\cite{T5XXL} and BERT~\cite{BERT} are trained on text-only corpus, which are equipped with powerful comprehension and reasoning abilities. As indicated in literature~\cite{Imagen}, these models exhibit dissimilar effects in image generation. To handle with text input, text encoder converts each discrete word to an unique token, which is used to retrieve the token embedding in learned look-up table. Finally, these word-level features are further processed to obtain the text embedding that encompasses holistic semantic.

\noindent$\bullet$
\textbf{Attention Mechanism}. Transformer block~\cite{Transformer} is default built-in module in existing large-scale models because of it's scalability. Specifically, cross-attention enables the communication between image features and text embedding, facilitating the text-guided generation, while self-attention is responsible for capturing the spatial correlation of image patches. The process of attention mechanism is formalized as:
\begin{equation}
\begin{aligned}\label{eq:attention}
\mathrm{Attention}(\mathcal{Q},\mathcal{K},\mathcal{V})&=\mathrm{Softmax}(\frac{\mathcal{Q}\mathcal{K}^T}{\sqrt{d}})\mathcal{V} \\
&=\mathcal{A}\mathcal{V}, \\
\end{aligned}
\end{equation}
where, $\mathcal{Q}$, $\mathcal{K}$ and $\mathcal{V}$ are query, key and value respectively, and $d$ indicates hidden dimension. $\mathcal{A}$ is attention map. In self-attention, these variables are all calculated from image features through corresponding projection weights. Differently, cross-attention obtains $\mathcal{K}$ and $\mathcal{V}$ from text features for cross-modal computation.

\noindent$\bullet$
\textbf{Text-to-Image Model}. Empowered by sophisticated text encoders~\cite{CLIP,T5XXL}, there are various T2I models, like StableDiffusion~\cite{StableDiffusion}, Imagen~\cite{Imagen}, DALLE-2~\cite{DALLE-2},  GLIDE~\cite{GLIDE} and so on~\cite{StableDiffusionXL,PixArt-alpha}. Slightly different from Eq. \ref{eq:ddpm-loss}, the training objective of these models is expressed as:
\begin{equation}
\begin{aligned}\label{eq:ddpm-conditional-loss}
E_{ (\mathbf{z}_0,\mathcal{C}) \sim p_{data},t \sim \mathcal{U}(1, T), \epsilon \sim \mathcal{N}(\mathbf{0}, \mathbf{I})}\left[\lambda_t \left\|\varepsilon_{\theta}\left(\mathbf{z}_t, t, \mathcal{C} \right)-\epsilon \right\|^2\right], 
\end{aligned}
\end{equation}
where $\mathcal{C}$ is specified as text condition. Among these models, since the code of StableDiffusion is open-source, it is widely studied in following works~\cite{DreamFusion,TokenFlow,P2P}. StableDiffusion builds on Latent Diffusion Model (LDM)~\cite{StableDiffusion}, which employs autoencoder~\cite{VAE,VQGAN,VQVAE} to project input image to lower-dimensional space for computing efficiency. Meanwhile, it adopts U-Net~\cite{U-Net} as the architecture of noise estimator and operates in latent space.

\subsection{Notation}
We list commonly used notations in \tablename~\ref{tab:conception_notation} for facilitating understanding of the survey. In remaining parts, we will indicate condition $\mathcal{C}$ in formula if necessary to signify the inclusion of conditional inputs, such as $F_{bw}^{DI}(\mathbf{z}_t,t,\mathcal{C})$. Meanwhile, in cases where no differentiation is needed, we unify the forward (inversion) / backward (sampling) process of DDPM and DDIM as $F_{fw}$ / $F_{bw}$.

\begin{table}
\centering
\small
\caption{\textbf{Commonly used notations in this survey}.}
\vspace{-3mm}
\label{tab:conception_notation}
\renewcommand{\arraystretch}{1.8}
\begin{adjustbox}{width=0.46\textwidth}
\begin{tabular}{c c}
\toprule[0.1em]
\textbf{Notations}  & Descriptions \\ 
\midrule[0.1em]
$\mathbf{z}_0$ / $\mathbf{z}_t$ / $\mathbf{z}_T$  & Noisy latent in $0$ / $t$ / $T$ step. \\
$\tilde{\varepsilon}_\theta$ / $\varepsilon_\theta$  & Noise estimation with / without classifier-free guidance.\\
$F_{fw}$ / $F_{bw}$  & Forward (Inversion) / Backward (Sampling) process.\\
$F_{fw}^{DP}$ / $F_{bw}^{DP}$  & Forward / Backward process in DDPM.\\
$F_{fw}^{DI}$ / $F_{bw}^{DI}$  & Inversion / Sampling process in DDIM.\\
\hline
$S_I$ / $G$ / $\mathbf{z}_0^e$ & Source image set / Guidance set / Edited image. \\
$\mathbf{z}_t^s$ / $\mathbf{z}_t^e$ & Noisy latent of source / editing image in $t$ step. \\
$\Phi_I$ / $\mathcal{C}_I$ & \inversionfeaturesuppercase~/ Source prompt. \\
$F_{inv}$ / $F_{edit}$ & Inversion / \editingalgorithmbig algorithm. \\
$F_{inv}^{T}$ / $F_{inv}^F$ & Tuning-based / Forward-based inversion algorithm. \\
\makecell[c]{$F_{edit}^{Norm}$ / $F_{edit}^{Attn}$ / $F_{edit}^{Blend}$ \\ $F_{edit}^{Score}$ / $F_{edit}^{Optim}$} & \makecell[c]{Normal / Attention-based / Blending-based \\ Score-based  / Optimization-based \editingalgorithmsmall algorithm.}\\
\bottomrule
\end{tabular}
\end{adjustbox}
\end{table}

\section{Problem Formulation}
\label{sec:problem-formulation}
Before investigating advanced methods, we first present our definition of image editing and introduce several critical aspects, like multimodal user guidance and editing scenarios involved in our topic. Furthermore, we provide a detailed introduction to the proposed unified framework. \tablename~\ref{tab:paper-list} illustrates these elements of representative approaches.

\begin{table*}[!th]\footnotesize
    \centering
    \footnotesize
    \caption{\textbf{Summarization of representative works}. We use abbreviations starting with \textbf{``T''} to notify editing scenarios: \textbf{o}bject \textbf{m}anipulation (\objectmanipulationshort) / \textbf{a}ttribute \textbf{m}anipulation (\attributemanipulationshort) / \textbf{s}patial \textbf{t}ransformation (\spatialtransformationshort) / \textbf{i}npainting (\inpaintingshort) / \textbf{s}tyle \textbf{c}hange (\stylechangeshort) / \textbf{i}mage \textbf{t}ranslation (\imagetranslationshort) / \textbf{s}ubject-\textbf{d}riven \textbf{c}ustomization (\subjectcustomizationshort) / \textbf{a}ttribute-\textbf{d}riven customization (\attributecustomizationshort). For conditions, we use abbreviations starting with \textbf{``G''}: \textbf{s}tatic \textbf{t}ext (\gtextshort) / \textbf{i}nstruction (\ginstructionshort) / \textbf{n}atural \textbf{i}mage (\gimageshort) / \textbf{m}ask (\gmaskshort) / \textbf{u}ser \textbf{i}nterface (\guishort). For training-free methods, we use abbreviations starting with \textbf{``I''} and \textbf{``E''} to denote inversion and editing algorithms respectively: \textbf{t}uning-based (\tuningshort) / \textbf{f}orward-based (\forwardspaceshort) inversion, \textbf{n}ormal (\normalshort) / \textbf{a}ttention-based (\attentionshort) / \textbf{b}lending-based (\blendingshort) / \textbf{s}core-based (\scorebasedshort) / \textbf{o}ptimization-based (\optimizationshort) editing. For training-based methods, we use abbreviations starting with \textbf{``S''} to indicate injection schemes: \textbf{i}mage \textbf{c}oncatenation (\imageconcatenationshort) / \textbf{l}atent \textbf{b}lending (\latentblendingshort) / \textbf{i}mage \textbf{a}dapter (\imageadaptershort) / \textbf{t}extual \textbf{s}pace \textbf{a}dapter (\texturalspaceadaptershort) / \textbf{l}atent \textbf{s}pace \textbf{a}dapter (\latentspaceadaptershort).} \label{tab:paper-list}
    \vspace{-3mm}

    \scalebox{0.86}{
\begin{tabular}{p{0.12\textwidth}p{0.16\textwidth}p{0.16\textwidth}p{0.20\textwidth}p{0.4\textwidth}}
    \toprule
\belowrulesepcolor{gray!30!}
\rowcolor{gray!30!}\multicolumn{5}{c}{\textbf{Training-Free Approaches (Section~\ref{sec:inversion algorithms},~\ref{sec:editing-algorithm})}} \\ \aboverulesepcolor{gray!30!} \midrule

\belowrulesepcolor{gray!15!}
\rowcolor{gray!15!} Inversion Algorithm & \editingalgorithmbig Algorithm & Editing Scenario & Guidance Set & Method \\ \aboverulesepcolor{gray!15!} \midrule

\multirow{8}{*}{\tuningshort} & \multirow{4}{*}{\normalshort} & \objectmanipulationshort \concat \attributemanipulationshort &  \gtextshort & \cite{UniTune,HiPer}\\

& & \subjectcustomizationshort &  \gtextshort & \noindent\cite{TI,DreamArtist,Cones2,P+,NeTI,DreamBooth,Data-Perspective,Facechain-Sude,Custom-Diffusion,Cones,Svdiff,Lora-Image,ANOVA,CatVersion,Break-A-Scene,Clic,Disenbooth,Decoupled-Textual-Embeddings} \\

& & \attributecustomizationshort &  \gtextshort & \noindent\cite{Prospect,MATTE,Concept-Decomposition,Reversion,ADI,Lego,StyleDrop} \\

& \mycellcolor & \mycellcolor \objectmanipulationshort \concat \attributemanipulationshort & \mycellcolor  \ginstructionshort & \mycellcolor \cite{FOI} \\
&\multirow{-2}{*}{\mycellcolor \attentionshort} & \mycellcolor \subjectcustomizationshort & \mycellcolor  \gtextshort& \mycellcolor \cite{DreamMatcher,VICO} \\

& \multirow{2}{*}{\blendingshort} &  \objectmanipulationshort \concat \attributemanipulationshort & \gtextshort / \ginstructionshort & \cite{Imagic,Forgedit,DAC} / \cite{Zone}\\
&  &  \inpaintingshort & \gtextshort \concat \gmaskshort & \cite{HD-Painter}\\

& \mycellcolor & \mycellcolor \objectmanipulationshort \concat \attributemanipulationshort & \mycellcolor \gtextshort & \mycellcolor \cite{SINE}\\
&\multirow{-2}{*}{\mycellcolor \scorebasedshort} & \mycellcolor \subjectcustomizationshort & \mycellcolor \gtextshort & \mycellcolor \cite{DCO}\\

\hdashline
& \multirow{1}{*}{\normalshort} & \objectmanipulationshort \concat \attributemanipulationshort &   \gtextshort &  \cite{EDICT,BDIA}\\

& \mycellcolor & \mycellcolor \objectmanipulationshort \concat \attributemanipulationshort & \mycellcolor  \gtextshort  & \mycellcolor \cite{NTI,NPI,ProxEdit,FPI,PnP-Inversion,DPL,Edit-Friendly,P2P,PnP,FPE,StyleDiffusion,Inversion-Free,CycleDiffusion} \\

& \mycellcolor & \mycellcolor \attributemanipulationshort & \mycellcolor  \gtextshort / \gimageshort & \mycellcolor  \cite{KV-Inversion,Localize-Object-Shape,MasaCtrl,TIC} / \cite{Cross-Image-Attention}\\

&\mycellcolor  & \mycellcolor \inpaintingshort & \mycellcolor  \gimageshort \concat \gmaskshort & \mycellcolor  \cite{TF-ICON} \\

&\multirow{-5}{*}{\mycellcolor \attentionshort} & \mycellcolor \stylechangeshort &  \mycellcolor  \gimageshort & \mycellcolor  \cite{Z-star,StyleInjection} \\

&   &  \objectmanipulationshort \concat \attributemanipulationshort &  \gtextshort  &  \cite{PTI,OIR,Diff-Edit,Pfb-diff,Uncovering-Disentanglement}\\

& &  \inpaintingshort &  \gtextshort \concat \gmaskshort / \gtextshort \concat \gimageshort \concat \gmaskshort &  \cite{Blended-Latent-Diffusion,High-Resolution-Image-Editing,Differential-Diffusion} / \cite{Tuning-Free-Image-Customization}\\

& \multirow{-3}{*}{\blendingshort} &  \spatialtransformationshort &  \guishort &  \cite{DesignEdit}\\

&  \mycellcolor  & \mycellcolor \objectmanipulationshort \concat \attributemanipulationshort & \mycellcolor  \gtextshort  & \mycellcolor \cite{pix2pix-zero,SEGA,Ledits,Ledits++,Stable-Artist,NMG,MagicRemover,AIDI}\\

& \mycellcolor \mycellcolor & \mycellcolor \spatialtransformationshort & \mycellcolor  \gtextshort~\concat~\gimageshort~\concat~\guishort  / \newline \gimageshort \concat \guishort & \mycellcolor \cite{Self-Guidance} / \newline \cite{Dragon-Diffusion}\\

&\multirow{-3}{*}{\mycellcolor \scorebasedshort}& \mycellcolor \imagetranslationshort &\mycellcolor  \gtextshort  &\mycellcolor  \cite{FreeControl}\\

&  &  \objectmanipulationshort \concat \attributemanipulationshort  &   \gtextshort / \gtextshort \concat \gmaskshort  &  \cite{DDS,CDS,Region-Aware,EBM} / \cite{Ground-A-Score}\\

\multirow{-13}{*}{\forwardspaceshort} & \multirow{-2}{*}{\optimizationshort}   & \subjectcustomizationshort &  \gtextshort & \cite{Pick-and-Draw} \\

\hdashline
\multirow{5}{*}{\tuningshort \concat \forwardspaceshort} & \multirow{1}{*}{\mycellcolor \normalshort} & \mycellcolor \stylechangeshort & \mycellcolor \gtextshort \concat \gimageshort & \mycellcolor \cite{InST} \\

& \multirow{1}{*}{\attentionshort} & \objectmanipulationshort \concat \subjectcustomizationshort &  \gtextshort \concat \gimageshort &  \cite{Custom-Edit,PhotoSwap,DreamEdit} \\

& \multirow{1}{*}{\mycellcolor \blendingshort} & \mycellcolor 
\inpaintingshort \concat \subjectcustomizationshort & \mycellcolor \gtextshort \concat \gimageshort \concat \gmaskshort & \mycellcolor \cite{DreamEdit} \\

& \multirow{1}{*}{\scorebasedshort} & 
\spatialtransformationshort & \gimageshort \concat  \guishort & \cite{DiffEditor} \\

& \multirow{1}{*}{\mycellcolor \optimizationshort} & \mycellcolor 
\spatialtransformationshort &  \mycellcolor \gmaskshort \concat \guishort & \mycellcolor \cite{Drag-Diffusion} \\

\midrule
\belowrulesepcolor{gray!30!}
\rowcolor{gray!30!}\multicolumn{5}{c}{\textbf{Training-Based Approaches (Section~\ref{sec:end-to-end image editing})}} \\ \aboverulesepcolor{gray!30!} \midrule

\belowrulesepcolor{gray!15!}
\rowcolor{gray!15!} Editing Scenario & Method & Guidance Set & Injection Scheme & Data Source\\ \aboverulesepcolor{gray!15!} \midrule
\multirow{8}{*}{\objectmanipulationshort + \attributemanipulationshort} & InstructPix2Pix ~\cite{InstructPix2Pix} & \ginstructionshort &  \imageconcatenation & \cite{LAION-5B} \\

&  MagicBrush~\cite{MagicBrush}  & \ginstructionshort &  \imageconcatenation & \cite{COCO} \\

&HIVE~\cite{HIVE} & \ginstructionshort &  \imageconcatenation & \cite{HIVE,LAION-5B,InstructPix2Pix} \\

& InstructDiffusion~\cite{InstructDiffusion} & \ginstructionshort  &  \imageconcatenation & \cite{COCO,COCO-Stuff,MPII,CrowdPose,AIC,PhraseCut,SAM,OpenImage} \\
&Emu-Edit~\cite{Emu-Edit}  & \ginstructionshort  & \imageconcatenation & \cite{Emu-Edit} \\

&  MGIE~\cite{MGIE} & \ginstructionshort &  \imageconcatenation \concat \latentspaceadapter & \cite{InstructPix2Pix} \\

&SmartEdit ~\cite{SmartEdit}  & \ginstructionshort &  \imageconcatenation \concat \texturalspaceadapter & \cite{RefCOCO,COCO-Stuff,GRES,InstructPix2Pix,MagicBrush,Lisa,LLAVA} \\

&RIE ~\cite{RIE}  & \ginstructionshort &  \texturalspaceadaptershort & \cite{GRES,RefCOCO} \\

\mycellcolor &  \mycellcolor Imagen Editor~\cite{Imagen-Editor} & \mycellcolor \gtextshort \concat \gmaskshort & \mycellcolor \imageconcatenation & \mycellcolor \cite{OpenImage,Visual-Genome}\\

\mycellcolor &\mycellcolor SmartBrush~\cite{SmartBrush} &\mycellcolor \gtextshort \concat \gmaskshort  & \mycellcolor \latentblending & \mycellcolor \cite{LAION-400M} \\

\mycellcolor &  \mycellcolor PowerPaint~\cite{PowerPaint} & \mycellcolor \gtextshort \concat \gmaskshort & \mycellcolor \imageconcatenation & \mycellcolor \cite{OpenImage,LAION-5B} \\

\mycellcolor & \mycellcolor PbE~\cite{Paint-by-Example} & \mycellcolor \gimageshort \concat \gmaskshort & \mycellcolor \imageconcatenation \concat \texturalspaceadapter & \mycellcolor \cite{OpenImage} \\

\mycellcolor & \mycellcolor ObjectStitch~\cite{ObjectStitch} & \mycellcolor \gimageshort \concat \gmaskshort & \mycellcolor \latentblending \concat \texturalspaceadapter & \mycellcolor \cite{LAION-400M} \\
\mycellcolor & \mycellcolor RIC~\cite{Reference-Based-Composition}& \mycellcolor \gimageshort \concat \gmaskshort & \mycellcolor \imageconcatenation \concat \texturalspaceadapter & \mycellcolor \cite{Danbooru} \\
\mycellcolor & \mycellcolor PhD~\cite{PHD} & \mycellcolor \gtextshort \concat \gimageshort \concat \gmaskshort& \mycellcolor \imageadapter & \mycellcolor \cite{OpenImage} \\
\multirow{-8}{*}{\mycellcolor \inpaintingshort} & \mycellcolor AnyDoor~\cite{Anydoor}& \mycellcolor \gimageshort \concat \gmaskshort  & \mycellcolor \imageadapter \concat \latentspaceadapter &\mycellcolor \cite{YouTubeVOS,YouTubeVIS,UVO,MOSE,VIPSeg,BURST,MVImgNet,VitonHD,FashionTryon,MSRA-10K,DUT,HFlickr,LVIS,SAM} \\

\multirow{8}{*}{\imagetranslationshort} &  ControlNet~\cite{ControlNet} & \gtextshort &  \imageadapter & \cite{ControlNet} \\
& T2I-Adapter~\cite{T2I-Adapter} &  \gtextshort &  \imageadapter & \cite{COCO,LAION-5B} \\
& SCEdit~\cite{SCEdit} & \gtextshort &  \imageadapter & \cite{LAION-5B} \\
& UniControl~\cite{UniControl} &  \gtextshort &  \imageadapter & \cite{LAION-5B} \\
& Cocktail ~\cite{Cocktail}  & \gtextshort &  \imageadapter & \cite{LAION-5B} \\
& Uni-ControlNet~\cite{Uni-ControlNet}  & \gtextshort &  \imageadapter & \cite{LAION-400M} \\
& CycleNet~\cite{CycleNet}  & \gtextshort &  \imageadapter & \cite{COCO,OpenImage} \\
& CycleGAN-Turbo~\cite{CycleGAN-Turbo}   & \gtextshort &  \latentblending & \cite{DENSE,Bdd100k} \\

\mycellcolor &  \mycellcolor Taming~\cite{Taming} & \mycellcolor  \gtextshort & \mycellcolor \texturalspaceadapter \concat \latentspaceadapter & \mycellcolor \cite{CelebA,LSUN} \\
\mycellcolor & \mycellcolor InstantBooth~\cite{InstantBooth} & \mycellcolor \gtextshort & \mycellcolor \texturalspaceadapter & \mycellcolor \cite{InstantBooth} \\
\mycellcolor & \mycellcolor E4T ~\cite{E4T} & \mycellcolor \gtextshort & \mycellcolor \texturalspaceadapter & \mycellcolor \cite{FFHQ,LSUN,CelebA-HQ,WikiArt} \\
\mycellcolor & \mycellcolor ProFusion~\cite{Enhance-Detail} &  \mycellcolor \gtextshort & \mycellcolor \texturalspaceadapter & \mycellcolor \cite{FFHQ} \\
\mycellcolor & \mycellcolor FastComposer~\cite{FastComposer}  & \mycellcolor \gtextshort & \mycellcolor \texturalspaceadapter & \mycellcolor \cite{FFHQ} \\
\mycellcolor & \mycellcolor PhotoMaker~\cite{PhotoMaker}  & \mycellcolor \gtextshort & \mycellcolor \texturalspaceadapter & \mycellcolor \cite{PhotoMaker} \\
\mycellcolor & \mycellcolor Photoverse~\cite{Photoverse}  & \mycellcolor \gtextshort & \mycellcolor \texturalspaceadapter \concat \latentspaceadapter & \mycellcolor \cite{FFHQ,CelebA-HQ,Fairface} \\
\mycellcolor & \mycellcolor InstantID~\cite{InstantID} & \mycellcolor \gtextshort & \mycellcolor \imageadapter \concat \texturalspaceadapter & \mycellcolor \cite{LAION-Face} \\
\mycellcolor &\mycellcolor ELITE~\cite{ELITE} & \mycellcolor \gtextshort & \mycellcolor \texturalspaceadapter \concat \latentspaceadapter & \mycellcolor \cite{OpenImage} \\
\mycellcolor & \mycellcolor BLIP-Diffusion~\cite{Blip-Diffusion}  & \mycellcolor \gtextshort & \mycellcolor \texturalspaceadapter & \mycellcolor \cite{LAION-400M,COCO,Conceptual-Captions,Visual-Genome,OpenImage} \\
\mycellcolor & \mycellcolor Domain-Agnostic~\cite{Domain-Agnostic} & \mycellcolor \gtextshort & 
 \mycellcolor \texturalspaceadapter & \mycellcolor \cite{ImageNet,OpenImage} \\

 
\mycellcolor & \mycellcolor UMM~\cite{UMM} & \mycellcolor \gtextshort & \mycellcolor \texturalspaceadapter & \mycellcolor \cite{LAION-400M} \\
\mycellcolor & \mycellcolor Subject-Diffusion~\cite{Subject-Diffusion} & \mycellcolor \gtextshort & \mycellcolor \texturalspaceadapter \concat \latentspaceadapter & \mycellcolor \cite{LAION-5B} \\
\multirow{-15}{*}{\mycellcolor \subjectcustomizationshort} & \mycellcolor Instruct-Imagen~\cite{Instruct-Imagen} & \mycellcolor \ginstructionshort & \mycellcolor \latentspaceadapter & \mycellcolor \cite{WikiArt,CelebA,CelebA-HQ,Subject-Driven-Diffusion,Sketch2Image} \\

\multirow{3}{*}{\attributecustomizationshort} & ArtAdapter~\cite{ArtAdapter} &  \gtextshort &  \texturalspaceadapter & \cite{LAION-5B,WikiArt} \\
& DreamCreature~\cite{DreamCreature} &  \gtextshort &  \texturalspaceadapter & \cite{ucsd-bird,stanford-dog} \\
& Lee~\emph{et al}~\cite{Language-Informed} & \gtextshort &  \texturalspaceadapter & \cite{Language-Informed} \\
\bottomrule
\end{tabular}
}
\vspacefigtext
\end{table*}

\subsection{Definition of Multimodal-Guided Image Editing} \label{sec:editing-definition} 
Given source image set $S_I$ and multimodal guidance set $G$, image editing aims to identify the visual elements to be preserved upon specific scenario and generate the edited image $\mathbf{z}_0^e$, which retains desired contents in $S_I$ and reflects editing targets involved in $G$. In our opinion, maintained concepts not only refer to low-level semantic, \eg, pixels of editing-unrelated region, but also some high-level semantic, \eg, identity or other attributes. \figurename~\ref{fig:editing-tasks} illustrates the examples of low-level semantics in rows 1-6, and high-level cases in rows 7-8.



\subsection{Multimodal User Guidance}
For controllable editing guided by $G$, we list some commonly used control signals of different modalities in following.

\noindent$\bullet$
\textbf{Natural Language}. Natural language is convenient and flexible for human beings to describe particular purposes, which is widely used in recent studies~\cite{PnP,NTI,P2P,DreamBooth,TI}. There are several forms of textural guidance. The common one is \textit{static~text}, which represents the target through a complete description. For example, ``a blue dog'' indicates that the user wants to change the hair of dog to blue color. For some works~\cite{NTI,Edit-Friendly}, a pair of descriptions is demanded to illustrate the difference before and after editing. In contrast, \textit{instruction} is more flexible, where a single command-style sentence is used to depict purpose, like ``turn the dog blue".

\noindent$\bullet$
\textbf{Image}. Image is an intuitive representation that conveys the semantic specified in visual content, which is hard to articulate by language. The common form is \textit{natural~image}~\cite{Cross-Image-Attention,Paint-by-Example}, which can be captured by ubiquitous devices, or synthesised by generation models~\cite{Imagen}~\cite{StableDiffusionXL}. For instance, if someone wishes to transfer his / her painting style to source image, he / she only needs to feed a certain amount of creations to editing method. Another form is \textit{mask}~\cite{Blended-Diffusion,RePaint}, which is essentially a binary image and indicates the interesting region to be modified. Mask is often used as the auxiliary condition to implement local editing.

\noindent$\bullet$
\textbf{User Interface}. User interface, such as mouse operation (like click and drag)~\cite{Drag-Diffusion,Self-Guidance}, sliding bar and input box~\cite{DesignEdit}, \emph{etc}., provides a interactive way to manipulate the image. Algorithms are responsible for translating specific user action to recognizable numerical parameters. For example, users can drag the object to target location, where editing methods transform the mouse operation to point coordinates~\cite{Drag-Diffusion}.

\noindent$\bullet$
\textbf{Combination of Control Signals}. Due to the richness of user guidance, some methods~\cite{Custom-Edit,Blended-Latent-Diffusion,Self-Guidance,Dragon-Diffusion,Instruct-Imagen} simultaneously accept multiple control signals of different modalities. For example, several works~\cite{Blended-Latent-Diffusion,SmartBrush} use textural prompt and mask for inpainting task to fill the area with semantic contents. 

\begin{figure*}[!t]
	\centering
    \includegraphics[width=1\linewidth]{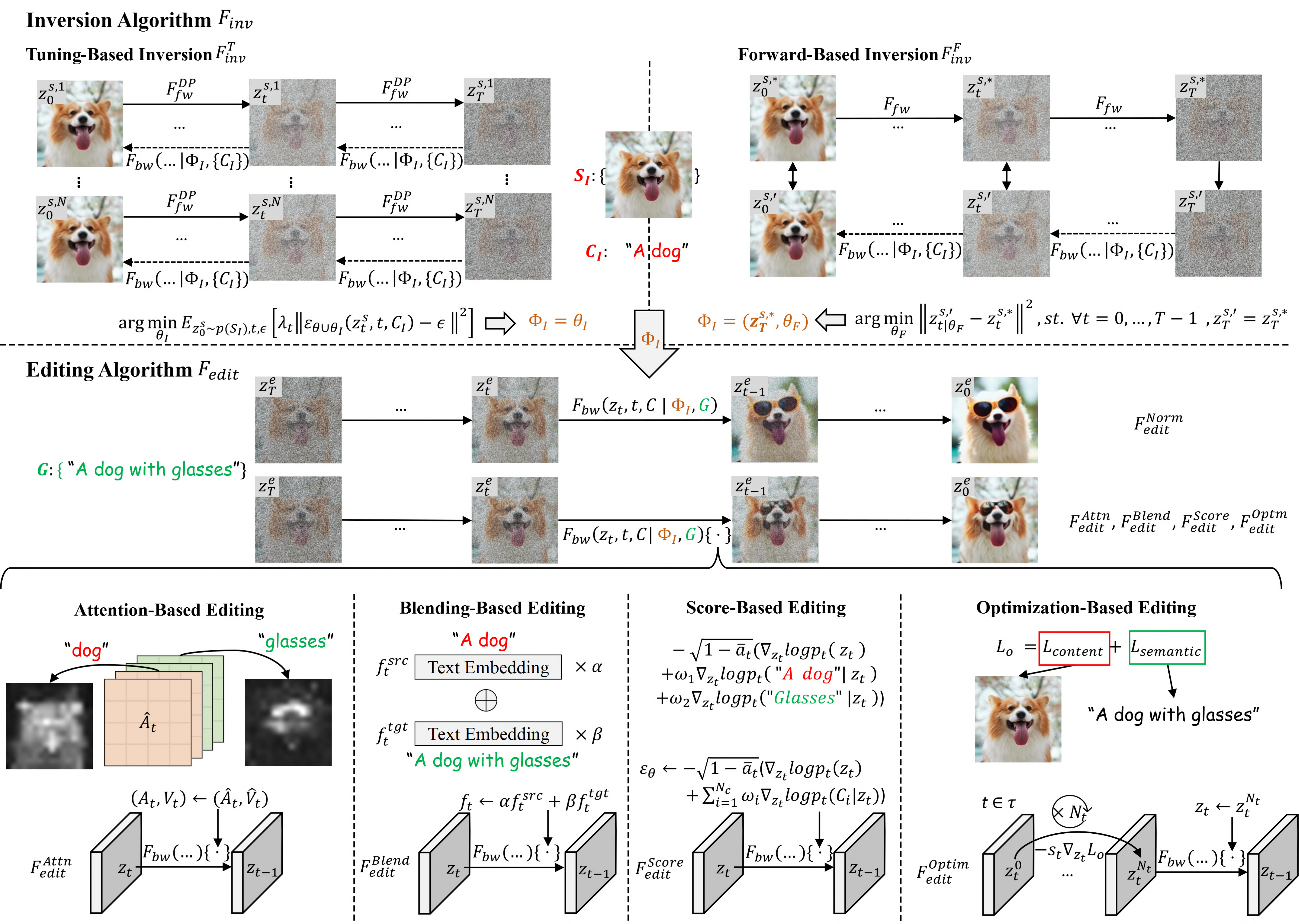}
    \caption{\textbf{Unified Framework}. We present an example of object addition to illustrate the cooperation of two algorithm families within proposed framework. Inversion algorithm $F_{inv}$ encodes source images $I_s$ into $\Phi_I$, and source prompt $\mathcal{C}_I$ identifies original contents. Editing algorithm $F_{edit}$ employs $\Phi_I$ and guidance set $G$ to infer the edited image $\mathbf{z}_0^e$.}
	\label{fig:unified-framework}
\end{figure*}

\subsection{Editing Scenario}
We enumerate most of editing scenarios / tasks involved in reviewed methods and divide them into two groups.

\noindent$\bullet$
\textbf{\contentawarebig}. Tasks belonging to this group intend to maintain low-level semantics that are unrelated with editing, while achieving targets indicated in $G$. 
\begin{enumerate}

    \item \textbf{Local Editing}. This subgroup modifies the image in local area. 1. \textit{Object Manipulation}. The task refers to addition / removal / replacement of designated object~\cite{NTI,P2P}.  2. \textit{Attribute Manipulation}. This category aims to enhance / weaken / change the intrinsic attribute of object, like color, texture, pose and action, \emph{etc}~\cite{MasaCtrl,KV-Inversion,Localize-Object-Shape,Cross-Image-Attention}.  3.\textit{Spatial Transformation}. This task changes spatial property of object\cite{Self-Guidance,Dragon-Diffusion,Drag-Diffusion}, like translation, scaling, and local distortion. 4. \textit{Inpainting}. This category fills the interesting area in source image with coherent content~\cite{Blended-Latent-Diffusion,High-Resolution-Image-Editing}.
    \item \textbf{Global Editing}. Tasks in this subgroup modify the global semantic of source image. 1. \textit{Style Change}. This task changes the style of source image to another one~\cite{Cross-Image-Attention,Style-Aligned}. 2. \textit{Image Translation}. The task transfers the image from source domain to target domain, like depth map to natural image~\cite{ControlNet,UniControl}.
\end{enumerate}

\noindent$\bullet$
\textbf{\contentfreebig}. In contrast to \contentawaresmall tasks, this group aims to preserve high-level semantics and reproduce concepts in the novel context guided by $G$. 

\begin{enumerate}
    \item \textbf{Subject-Driven Customization}: This task~\cite{InstantID,Svdiff} generates novel images of target subject, where the identity is maintained.
    \item \textbf{Attribute-Driven Customization}: Instead of learning a holistic concept, this task~\cite{Concept-Decomposition,Lego} extracts decoupled attributes, like shape, style, texture, and action, \emph{etc}., which can be assigned to other objects.

\end{enumerate}

Specifically, \contentawaresmall is the main topic discussed in current works~\cite{PnP,editing-survey,P2P}. Unlike in this common setting, our definition extends the range of discussion, where the preserved concepts involve more aspects, such as identity, style and so on. Therefore, several generation tasks meeting our definition are also included in this survey, such as customization~\cite{DreamBooth,TI,Custom-Diffusion,NeTI} and conditional generation under image guidance~\cite{ControlNet,SCEdit,ELITE}. We illustrate various editing scenarios along with different control signals in \figurename~\ref{fig:editing-tasks}. Significantly, since both textural and image conditions are difficult to depict the editing purpose in spatial transformation, most of tasks from this category receive guidance from user interface, like mouse operation.

\subsection{Evaluation of Image Editing}
Regardless of editing tasks, there are two fundamental metrics for evaluating the fidelity to source images and guidance respectively: 
\begin{enumerate}
    \item \textbf{Content Consistency}: The edited image $\mathbf{z}_0^e$ has to retain the desired visual elements of $S_I$. Several quantitative metrics measure the consistency of source and edited images, such as CLIP-I~\cite{CLIP}, LPIPS~\cite{LPIPS} and FID~\cite{FID}, where CLIP-I calculates the cosine similarity of CLIP image embeddings.
    \item \textbf{Semantic Fidelity}: The edited image $\mathbf{z}_0^e$ has to reflect editing targets involved in $G$. For quantitative evaluation, CLIP-T~\cite{CLIP} is used to measure the fidelity to textural prompt, which computes the cosine similarity of image and text embeddings from CLIP. Besides, directional CLIP similarity~\cite{Stylegan-Nada} receives image pair along with source and target descriptions, indicating the accuracy of editing direction. 

\end{enumerate}

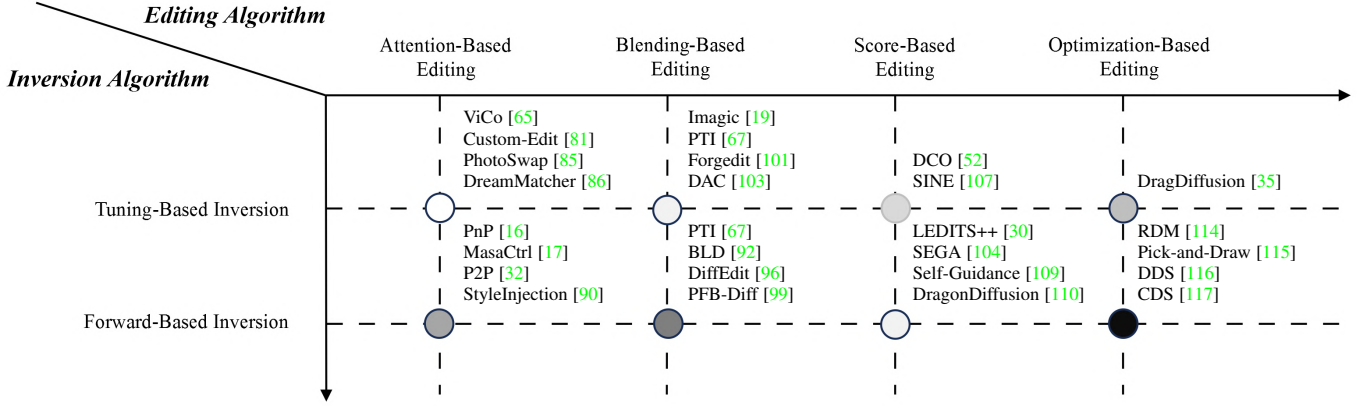
\begin{figure*}[!t]
	\centering
 \footnotesize
    \includegraphics[width=1\linewidth]{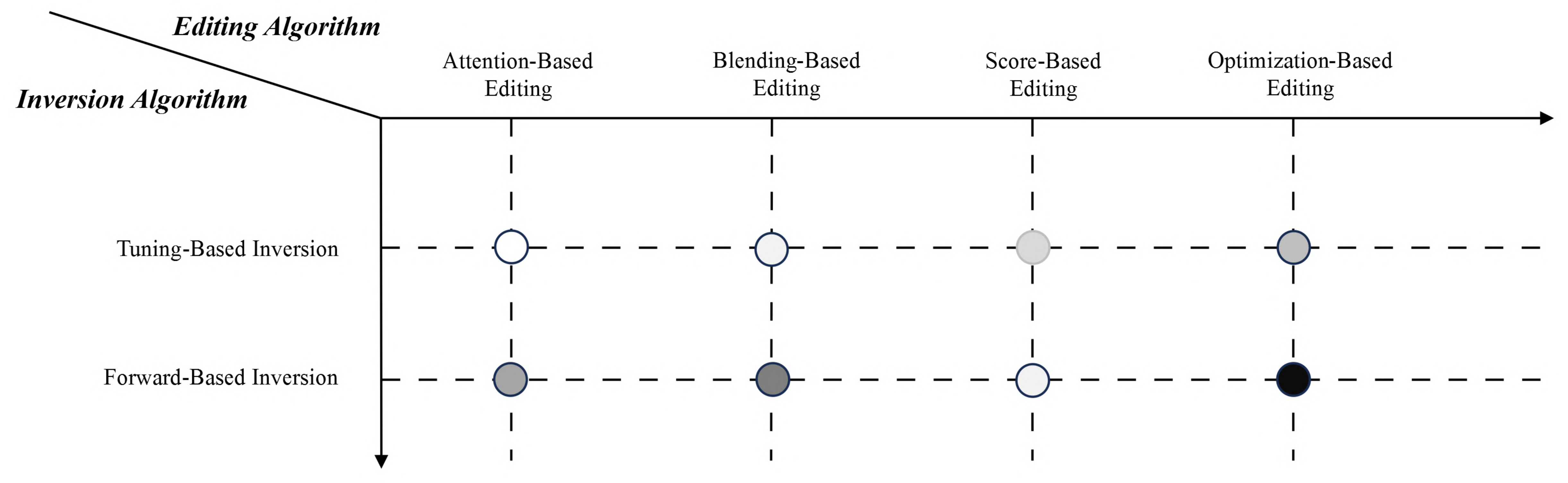}
    \begin{picture}(0,0)
        \scriptsize {
            \put(-80,119){ViCo~\cite{VICO}}
            \put(-80,111){Custom-Edit~\cite{Custom-Edit}}
            \put(-80,103){PhotoSwap~\cite{PhotoSwap}}
            \put(-80,95){DreamMatcher~\cite{DreamMatcher}}
            \put(5,119){Imagic~\cite{Imagic}}
            \put(5,111){PTI~\cite{PTI}}
            \put(5,103){Forgedit~\cite{Forgedit}}
            \put(5,95){DAC~\cite{DAC}}
            \put(90,103){DCO~\cite{DCO}}
            \put(90,95){SINE~\cite{SINE}}
            \put(175,95){DragDiffusion~\cite{Drag-Diffusion}}

            \put(-80,76){PnP~\cite{PnP}}
            \put(-80,68){MasaCtrl~\cite{MasaCtrl}}
            \put(-80,60){P2P~\cite{P2P}}
            \put(-80,52){StyleInjection~\cite{StyleInjection}}

            \put(5,76){PTI~\cite{PTI}}
            \put(5,68){BLD~\cite{Blended-Latent-Diffusion}}
            \put(5,60){DiffEdit~\cite{Diff-Edit}}
            \put(5,52){PFB-Diff~\cite{Pfb-diff}}

            \put(90,76){LEDITS++~\cite{Ledits++}}
            \put(90,68){SEGA~\cite{SEGA}}
            \put(90,60){Self-Guidance~\cite{Self-Guidance}}
            \put(90,52){DragonDiffusion~\cite{Dragon-Diffusion}}

            \put(175,76){RDM~\cite{Region-Aware}}
            \put(175,68){Pick-and-Draw~\cite{Pick-and-Draw}}
            \put(175,60){DDS~\cite{DDS}}
            \put(175,52){CDS~\cite{CDS}}

        }
    \end{picture}
    \vspace{-6mm}
    \caption{\textbf{Application of unified framework}. We represent some studies from different tasks within our framework, like object / attribute manipulation~\cite{DDS,P2P,PnP,MasaCtrl,Imagic,Forgedit,PTI,DAC,SINE,Diff-Edit,Pfb-diff,SEGA,Ledits++,Region-Aware,CDS,PTI}, spatial transformation~\cite{Drag-Diffusion,Self-Guidance,Dragon-Diffusion}, inpainting~\cite{Blended-Latent-Diffusion}, style change~\cite{StyleInjection}, and customization~\cite{DreamMatcher,Custom-Edit,PhotoSwap,VICO,DCO,Pick-and-Draw}. }
	\label{fig:representative-combination}
\end{figure*}

\subsection{Unified Framework}
Current methods can be integrated in an unified framework, where the editing process is divided into two primary algorithm families.

\noindent$\bullet$
\textbf{Inversion Algorithm}. Inversion algorithm $F_{inv}$ encodes source images into $\Phi_I$, named \textit{inversion clue}, which is used in editing stage to reconstruct desired contents. The process is expressed as:
\begin{equation} \label{eq:inversion algorithm}
    \Phi_I=F_{inv}(S_I,\mathcal{C}_I),
\end{equation}
where $\mathcal{C}_I$ is the source prompt that identifies original contents. We demonstrate different algorithms in the top of \figurename~\ref{fig:unified-framework}. Significantly, Eq.~\ref{eq:inversion algorithm} is also applicable to image condition, as users want to retain certain contents from reference image in $\mathbf{z}_0^e$.

\noindent$\bullet$
\textbf{\editingalgorithmbig Algorithm}. Through incorporating $\Phi_I$ into the base model according to different inversion methods, \editingalgorithmsmall algorithm then employs the variant version of $\mathcal{C}_I$ in $G$ to reconstruct preserved contents and achieve the purpose, like ``a dog'' and ``a dog with glasses'' in \figurename~\ref{fig:unified-framework}. Formally, the goal of \editingalgorithmsmall algorithm $F_{edit}$ is to generate the edited image:
\begin{equation} \label{eq:editing algorithm}
\begin{aligned}
\mathbf{z}_0^e&=F_{edit}(\Phi_I,G).
\end{aligned}
\end{equation}
Specifically, $F_{edit}$ intervenes the backward process as:
\begin{equation} \label{eq:editing algorithm}
\begin{aligned}
F_{bw}(\mathbf{z}_t,t,\mathcal{C} \mid \Phi_I, G)\small\{~\!\cdot~\!\small\},
\end{aligned}
\end{equation}
where $F_{bw}(\mathbf{z}_t,t,\mathcal{C} \mid \Phi_I,G)$ demonstrates that backward process is performed based on $\Phi_I$ and $G$. Meanwhile, $\small\{~\!\cdot~\!\small\}$ represents the manipulation of $F_{edit}$, which enhances both content consistency and semantic fidelity. Significantly, control signals like mask or numerical parameters from user interface are only processed in $\small\{~\!\cdot~\!\small\}$, as they cannot be recognized by T2I model. Furthermore, we also refer to the normal version of $F_{edit}$ as $F_{edit}^{Norm}$:
\begin{equation} \label{eq:noraml-based editing}
\begin{aligned}
F_{bw}(\mathbf{z}_t,t,\mathcal{C} \mid \Phi_I,G),
\end{aligned}
\end{equation}
which indicates the absence of intervention. 

\noindent $\bullet$
\textbf{Cooperation of Inversion \& \editingalgorithmbig Algorithms}. We illustrate the cooperation of inversion and \editingalgorithmsmall algorithms through a simple case in \figurename~\ref{fig:unified-framework}. As shown in \figurename~\ref{fig:unified-framework}, other \editingalgorithmsmall algorithms outperform $F_{edit}^{Norm}$ in preserving appearance of the dog. For example, blending-based algorithm fuses the text embeddings of $\mathcal{C}_I$ with target embedding and uses the blended features to guide backward process, for achieving fidelity to both source image and text guidance. Attention-based algorithm injects the attention maps from source images to retain details of the dog.

With our proposed framework, users are able to combine proper approaches to accomplish particular purposes. We illustrate the inversion and editing algorithms of reviewed methods in \tablename~\ref{tab:paper-list}. It is worth noting that many of these studies~\cite{P2P,HD-Painter,Uncovering-Disentanglement} employ multiple editing algorithms simultaneously. For simplicity, we only category them based on the primary technology they use. \figurename~\ref{fig:representative-combination} presents several representative works to illustrate the applicability of our framework in multiple tasks.

\section{Inversion Algorithm}
\label{sec:inversion algorithms}
In literature of GAN-based methods~\cite{GAN-inversion,StyleGAN-Embedding}, \textit{inversion} refers to the process that embeds natural image into latent space, and the representation is fed to GAN for reconstruction. In this section, we investigate the inversion algorithms in diffusion-based models~\cite{NTI,DreamBooth,TI,Edit-Friendly} and categorize them into \textit{tuning-based inversion} and \textit{forward-based inversion}, which 
are denoted as $F_{inv}^{T}$ and $F_{inv}^{F}$ respectively. \figurename~\ref{fig:unified-framework} illustrates the ideas of them.

\subsection{Tuning-Based Inversion} \label{sec:tuning-based inversion space}
In order to re-create source contents, a big family of works~\cite{DreamBooth,TI,InST} exploit the original training process of diffusion models to implant source images into the generative distribution. Under this paradigm, the goal of $F_{inv}^{T}$ is to solve:
\begin{equation}
\begin{aligned}\label{eq:tuning-based inverse}
\underset{\theta_I}{\arg \min } E_{\mathbf{z}_0^s \sim p(S_I),t,\epsilon} \left[\lambda_t\left\|\varepsilon_{\theta \cup \theta_I}\left(\mathbf{z}_t^s, t, \mathcal{C}_I\right)-\epsilon_t\right\|^2\right], 
\end{aligned}
\end{equation}
where $\mathbf{z}_0^s$ is the source image sampled from $S_I$. ${\theta_I}$ indicates parameters to be updated and $\cup$ is union operation. Other terms are the same with Eq.~\ref{eq:ddpm-conditional-loss}. Under this paradigm, $\Phi_I=\theta_I$. In editing time, methods load $\theta_I$ to the base model, and use variant of $\mathcal{C}_I$ to reconstruct preserved concepts.

According to Eq.~\ref{eq:tuning-based inverse} and the upper-left corner of \figurename~\ref{fig:unified-framework}, tuning-based inversion algorithm optimizes all potential denoising paths to reconstruct source images from arbitrary Gaussian noise. $F_{inv}^T$ has following characteristics. 1. Since $F_{inv}^T$ optimizes numerous denoising trajectories, it has considerable reconstruction ability. 2. Since methods based on $F_{inv}^T$ perform sampling from random noise, they have high flexibility in terms of image layout. However, it is accompanied by lengthy tuning time and may loss a certain degree of generative capability due to one-shot or few-shot tuning. Therefore, $F_{inv}^T$ is mostly used in \contentfreesmall tasks~\cite{DreamBooth,TI,CatVersion,Cones2,StyleDrop}, with several cases in \contentawareshortsmall scenarios~\cite{Imagic,Custom-Edit,Forgedit,SINE}, since the former requires higher flexibility with lower demand in maintaining image structure and low-level semantics. We mainly talk about the challenges and corresponding solutions in \contentfreesmall tasks, while organizing these methods based on their tuning spaces.

\subsubsection{Textual Space} \label{sec:textual-space}
For T2I models~\cite{StableDiffusion,Imagen,StableDiffusionXL}, text encoders~\cite{CLIP,T5XXL,CLIP-ViT-bigG} are employed to extract representative features from text prompt, facilitating the cross-modal computation to generate creative contents. Therefore, a research line intends to optimize textual embedding ($\theta_I$) to complete different editing tasks~\cite{TI,DreamArtist,UniTune,HiPer}. However, restricted by parameter size, tuning in textual space often leads to poor reconstruction performance. As indicated in literature~\cite{Prospect,Concept-Decomposition,Mix-of-Show}, textual embedding tends to capture global semantic, while overlooking finer details. Therefore, several studies~\cite{P+,NeTI} extend ordinary space to tackle the issue. In contrast, other methods~\cite{Concept-Decomposition,Reversion,Lego} take advantage of the characteristic and try to decompose entangled attributes from images.

\noindent$\bullet$
\textbf{Extension of Textual Space}. A group of methods~\cite{TI,DreamArtist,UniTune,HiPer} explore the effectiveness of textual space in distinct editing scenarios, while others~\cite{P+,NeTI} make effort to enhance the reconstruction ability through space extension. For customization, Textual-Inversion (TI)~\cite{TI} introduces a learnable word embedding to represent the subject, and assigns a rare token to it as identifier, like ``\emph{sks}''. Through constructing $\mathcal{C}_I$ with the rare token, such as ``a photo of a \textless rare\_token\textgreater'', and applying Eq.~\ref{eq:tuning-based inverse} on text-image pairs, TI preserves the subject identity. In addition, DreamArtist~\cite{DreamArtist} jointly optimizes positive and negative embeddings~\cite{classifier-free} to alleviate overfitting issue of tuning on a single image. Specifically, the negative embedding is used to correct the mistake made by positive one. Other works~\cite{UniTune,HiPer} employ the technique in \contentawaresmall tasks. UniTune~\cite{UniTune} optimizes the embedding of $\mathcal{C}_I$. In editing stage, it concatenates $\mathcal{C}_I$ with target prompt to retain basic contents. Differently, HiPer~\cite{HiPer} optimizes the tail part of text embedding for better consistency, and replaces initial pieces with target embedding. 

Limited by the number of parameters, these methods fall short in maintaining finer details. Based on StableDiffusion\cite{StableDiffusion}, XTI~\cite{P+} introduces extended textual conditioning space $P_+$ to alleviate underfitting issue. Different from TI~\cite{TI}, where a single embedding is optimized, the method learns separate vector for each U-Net layer~\cite{StableDiffusion}. Layer-wise embeddings facilitate to extract multi-level features for better reconstruction. NeTI~\cite{NeTI} further expands $P_+$ along time dimension. Specifically, it trains a projection network to map current denoising step and layer number to word embedding. Meanwhile, the method appends nested dropout layer~\cite{Nested-Dropout} to last linear layer, which balances reconstruction ability and editability through adjusting the truncation value.

\noindent$\bullet$
\textbf{Extraction of Attributes}. According to relevant studies~\cite{Prospect,Mix-of-Show}, textual space prefers to capture in-domain features. Some approaches~\cite{Prospect,MATTE,Concept-Decomposition} leverage this property to decompose entangled attributes from a holistic concept. Since each denoising stage is responsible for distinct semantics, ProSpect~\cite{Prospect} introduces a hypernetwork to infer the time-wise embedding. These features capture dissimilar attributes, like materials and shape, \emph{etc}. Furthermore, MATTE~\cite{MATTE} achieves the goal through a finer approach. Similar with NeTI~\cite{NeTI}, it extends textual space in both layer-wise and time-wise dimensions. In editing time, the method uses a subset of embeddings to assemble different attributes. Concept Decomposition~\cite{Concept-Decomposition} provides a more intuitive solution. It uses a binary tree to represent concepts in distinct semantic-levels. For each iteration, the method decomposes the concept indicated in current node, through applying Eq.~\ref{eq:tuning-based inverse} on composited child identifiers and images generated by parent embedding. 

Compared with concrete visual contents, it's more challenging to extract implicit attributes from finite images. In this situation, simply employing pixel-wise diffusion loss is insufficient. A handful of methods~\cite{Reversion,ADI} intend to learn the action from source images. Since preposition in natural language represents the correlation of different subjects, Reversion~\cite{Reversion} exploits contrastive loss to align the action embedding with sampled preposition embeddings, while distancing it from other Part-of-Speech (POE) words. From another perspective, Action-Disentangled Identifier (ADI)~\cite{ADI} seeks for the action-related channels of word embedding through calculating gradient difference between source and auxiliary images, and only applies gradient descent to these key parameters. Moreover, Lego~\cite{Lego} aims to extract more general concepts from examplar images. It first assigns distinct tokens and word embeddings for subject and the concept to be learned respectively. In tuning process, $\mathcal{C}_I$ for subject-only images only include ``\textless subject\_token\textgreater", while others add ``\textless concept\_token\textgreater'' on this basis to indicate the injection of new attributes. Meanwhile, inspired from Reversion, the method employs contrastive loss to separate these embeddings in textural space.

\subsubsection{Model Space} \label{sec:model-space}
A big family of approaches~\cite{DreamBooth,Break-A-Scene,Facechain-Sude} additionally update modules from base model ($\theta_I$) to improve reconstruction ability. For example, some works~\cite{Imagic,Forgedit,Drag-Diffusion} optimizes both textural embedding and model parameters for content consistency after non-rigid editing or local distortion. Other methods~\cite{DreamBooth,Custom-Diffusion} enhance the details in customization. However, updating numerous parameters gives rise to language drift\cite{Language-Drift1,Language-Drift2} and catastrophic forgetting problems. That is, the model forgets prior knowledge after fine-tuning. In addition, they also struggle with disentangling unrelated contents, such as background, which impairs the identity of edited subject. These overfitting phenomena undermine editability in varying degrees. Regarding these issues, several solutions are as follows.

\noindent$\bullet$
\textbf{Prior Preservation Regularization}. A group of works~\cite{DreamBooth,Data-Perspective,DCO,Facechain-Sude} regularize tuning process to maintain the prior knowledge. As a pioneer work, DreamBooth~\cite{DreamBooth} introduces class-specific regularization dataset, which contains a certain number of prior images that are generated through the original model. By jointly tuning on regularized data and source images, it mitigates overfitting problem. Work from~\cite{Data-Perspective} further enhances the regularization dataset through constructing sophisticated prompt templates to create richer data. Instead of using fixed prior images throughout tuning, DCO~\cite{DCO} proposes to reduce Kullback-Leibler Divergence (KLD) in absence of regularized dataset. Borrowing the idea from Direct Preference Optimization (DPO)~\cite{DPO-Image}, which implicitly minimizes KLD of new policy and old one, DCO fine-tunes the base model with prior preservation. In addition, to maintain intrinsic characteristics of personalized subject, FaceChain-SuDe~\cite{Facechain-Sude} regularizes to increase conditional probability $p(\mathcal{C}_{cate}|\mathbf{z}_t,t,\mathcal{C}_{I})$, where $\mathcal{C}_{cate}$ depicts the super-category of customized subject. The novel regularization term encourages the method to retain private properties from super-category.

\noindent$\bullet$
\textbf{Efficient Fine-Tuning}. Inspired from Parameter Efficient Fine-Tuning (PEFT)~\cite{Lora}, a research line~\cite{Custom-Diffusion,Svdiff,ANOVA,StyleDrop} optimizes a small set of parameters to accelerate inversion process, while alleviating the overfitting issue. Several studies~\cite{Custom-Diffusion,Cones} identify critical parameters based on established metrics. Custom Diffusion~\cite{Custom-Diffusion} verifies that the relative change of most of parameters are small after tuning, except for cross-attention modules. Therefore, the method updates projection weights of key and value, while freezing other layers. From another perspective, Cones~\cite{Cones} determines important neurons by scaling down their values and checking whether the diffusion loss is reduced. Other works~\cite{Svdiff,Lora-Image}  introduce additional parameters into particular layers for efficient tuning. SVDiff~\cite{Svdiff} exploits Singular Value Decomposition (SVD) for learning weight offsets as $\Delta W\!=\!U \Delta \Sigma V^T$, where $\Delta \Sigma$ is the change of singular value matrix. LoRA~\cite{Lora} decomposes weight offsets into the production of two low-rank matrices as $\Delta W\!\!=\!\!DU$, while seeking for optimal down matrix $D$ and up matrix $U$. Moreover, ANOVA~\cite{ANOVA} proposes a design space to inject LoRA modules for more efficiency. 

\noindent$\bullet$
\textbf{Disentanglement of Unrelated Semantic}. Customization of specific concept is often affected by irrelevant contents, like background, or other objects, which impair the editability due to concept coupling. To address the challenge, several works enhance $\mathcal{C}_I$ through crafted templates~\cite{Data-Perspective} or descriptive caption estimated from multimodal model~\cite{Forgedit,BLIP-2} to include various concepts indicated in the image. Essentially, these studies assume that the generation model is capable for learning the correlation between text token and image features while disentangling different concepts in semantic space. Other approaches~\cite{Break-A-Scene,Clic} require binary masks to separate visual elements in image space, which are given by users or predicted from segmentation networks~\cite{SAM,li2023transformer,SVCNet}. For customizing multiple subjects from a single image simultaneously, Break-a-Scene~\cite{Break-A-Scene} only applies Eq.~\ref{eq:tuning-based inverse} inside masked regions, which excludes the effect of unrelated contents. Meanwhile, the method further aligns the cross-attention map of each subject with corresponding mask by minimizing their discrepancy, which prevents attention leakage. Borrowing the idea, Clic~\cite{Clic} intends to learn local concept from a holistic one. Specifically, it further considers out-of-mask region for integration of context information, and applies diffusion loss inside soft mask. 

\subsubsection{Image Adapter Space} \label{sec:adapter-space}
Except for intrinsic parameters in base model, another family of studies~\cite{InST,Disenbooth,VICO} perform inversion through tuning additional adapter network ($\theta_I$), which incorporates features of source image into backward process. Since image and text features from CLIP~\cite{CLIP} are aligned in the joint space, some methods~\cite{InST,Disenbooth,Decoupled-Textual-Embeddings} introduce projection network to map the image embedding into textual space, and inject the features into diffusion model through cross-attention modules. For transferring the style from reference image to target one, InST~\cite{InST} proposes an auxiliary adapter based on multi-layer attention to extract global semantic of style image. Other methods~\cite{Disenbooth,Decoupled-Textual-Embeddings} employ the adapter to disentangle different visual components. DisenBooth~\cite{Disenbooth} introduces an multi layer perceptron (MLP) to extract background-related features from CLIP image embedding, while the subject identity is learned in original textual space. In addition, method from~\cite{Decoupled-Textual-Embeddings} further learns pose-related features for finer disentanglement. Instead of using CLIP encoder, ViCo~\cite{VICO} leverages native noise estimator to extract finer details from source image. Specifically, the method injects learnable attention layers into base model, where query is calculated from generating image, key and value are from source images. The pixel-level mutual computation makes ViCo outperform in identity preservation.

\subsection{Forward-Based Inversion} \label{sec:foward-based inversion space}
The forward process in diffusion models naturally inverts image into noise space, while backward process re-creates the input based on this initial point. In absence of cumbersome test-time tuning, the idea is widely studied and refined in numerous studies~\cite{NTI,NPI,ProxEdit,BDIA,DreamMatcher,Pick-and-Draw} to adapt various tasks. Specifically, these methods purpose to reverse the fixed trajectory / path generated by forward process~\cite{DDPM,DDIM} for reproducing source image. Formally, intermediate noisy latents of \textit{inversion} and \textit{reconstruction} paths in $F_{inv}^{F}$ are expressed as:
\begin{small}
\begin{equation}
\begin{aligned}\label{eq:forward-backward-trajectories}
&\left[\mathbf{z}_{t+1}^{s,*} \right]_{t=0}^{T-1}=\left[F_{fw}(\mathbf{z}^{s,*}_{t},t,\mathcal{C}^*)\right]_{t=0}^{T-1}, &&\mathrm{st}.~\mathbf{z}_0^{s,*}=\mathbf{z}_0^s, \\
&\left[ \mathbf{z}_t^{s,\prime} \right]_{t=T-1}^{0}=\left[F_{bw}(\mathbf{z}_{t+1}^{s,\prime},t+1,\mathcal{C}_I \mid \theta_F)\right]_{t=T-1}^{0}, &&\mathrm{st}.~\mathbf{z}_T^{s,\prime}=\mathbf{z}_T^{s,*}, \\
\end{aligned}
\end{equation}
\end{small}
where $\mathbf{z}_0^s$ is source image from $S_I$. The superscripts $*$ and $\prime$ indicate variables in inversion and reconstruction trajectories respectively. $\theta_F$ represents method-related parameters~\cite{NTI,NPI,Edit-Friendly} for aligning $\mathbf{z}_t^{s,*}$ and $\mathbf{z}_t^{s,\prime}$. For example, NTI~\cite{NTI} optimizes ``null-text'' embedding ($\theta_F$) to minimize $\left\| \mathbf{z}_t^{s,*} - \mathbf{z}_t^{s,\prime} \right\|^2$. The goal of $F_{inv}^{F}$ is expressed as:
\begin{equation}
\begin{aligned}\label{eq:forward-based inverse}
\underset{\theta_F}{\arg \min } \left\|\mathbf{z}_{t \mid \theta_F}^{s,\prime}-\mathbf{z}_t^{s,*} \right\|^2,~\mathrm{st}.~\forall t \in 0,...,T-1,~\mathbf{z}_T^{s,\prime}=\mathbf{z}_T^{s,*},
\end{aligned}
\end{equation}
where $\mathbf{z}_{t \mid \theta_F}^{s,\prime}$ indicates that $\mathbf{z}_t^{s,\prime}$ is related to $\theta_F$. Specifically, $F_{inv}^F$ encodes source image into $\mathbf{z}^{s,*}_T$ and $\theta_F$, \emph{ie}., $\Phi_I=\left(\mathbf{z}^{s,*}_T,\theta_F \right)$. For editing, methods use $\mathbf{z}_T^{s,*}$ as initial point, while incorporating $\theta_F$ to retain basic contents in final result.

According to Eq.~\ref{eq:forward-based inverse} and top-right part of \figurename~\ref{fig:unified-framework}, methods in this group only reconstruct a single denoising path. There are several properties of $F_{inv}^F$. 1. Since source image is encoded in initial latent, the basic contents of image is preserved during sampling. 2. In contrast to $F_{inv}^T$, $F_{inv}^F$ does not change the output distribution, thereby maintaining the generative capability of base model. Therefore, this family is commonly exploited in \contentawaresmall tasks, due to the requirement of retaining low-level semantics. In this section, we categorize methods based on their forward processes~\cite{DDPM,DDIM}.

\subsubsection{DDIM Inversion} \label{sec:ddim-foward inversion space}
Early works~\cite{P2P,DDIM,classifier-guidance} employ the deterministic sampling characteristics of DDIM~\cite{DDIM} to invert real image, where $F_{fw}$ and $F_{bw}$ in Eq.~\ref{eq:forward-based inverse} are corresponding to $F_{fw}^{DI}$ and $F_{bw}^{DI}$ respectively. As demonstrated in Eq.~\ref{eq:ddim-forward inversion}, since $\mathbf{z}_{t}^{s,*}$ is inaccessible in DDIM inversion process, a common solution is using $\varepsilon_\theta\left(\mathbf{z}_{t-1}^{s,*}, t\right)$ to approximate $\varepsilon_\theta\left(\mathbf{z}_{t}^{s,*},t \right)$, which is nearly correct under the assumption of infinitely small step. Nevertheless, according to literature~\cite{NTI,NPI,EDICT}, above approximation often causes considerable accumulated error when applying classifier-free guidance~\cite{classifier-free} along with large guidance scale $\omega$, leading to poor reconstruction. Besides, since elements of noise map are not independent from each other, which is not exactly the same with training stage of diffusion model, ordinary DDIM inversion often leads to the decline of editability. For widely used classifier-free guidance strategy, above problems are unacceptable. In this section, we focus on solutions of these issues.

\noindent$\bullet$
\textbf{Approximation of Inversion Trajectory}. To alleviate accumulated error, several methods~\cite{NTI,NPI} intend to approximate the inversion trajectory. A group of works~\cite{NTI,PTI,StyleDiffusion,KV-Inversion} introduce learnable parameters ($\theta_F$) to reduce the difference of $\mathbf{z}_t^{s,*}$ and $\mathbf{z}_t^{s,\prime}$. Null-Text Inversion (NTI)~\cite{NTI} is the first method to take the issue into account for real image editing. For addressing the degradation of editability caused by high $\omega$, the inversion trajectory is generated by setting $\omega=1$ in classifier-free guidance. For preventing the reconstruction trajectory to deviate from inversion path, the method optimizes the ``null-text'' embedding $\varnothing$ in each step by minimizing $||\mathbf{z}_{t}^{s,*}-\mathbf{z}_{t}^{s,\prime}||^2$, where $\mathbf{z}_{t}^{s,\prime}$ is obtained with large $\omega$. In contrast, Prompt Tuning Inversion (PTI)~\cite{PTI} instead optimizes text embedding, which is interpolated with target embedding in editing time. Different from these methods, StyleDiffusion~\cite{StyleDiffusion} encodes the error into value matrices of cross-attention modules for better consistency. For non-rigid editing, KV Inverson~\cite{KV-Inversion} optimizes feature offsets of keys and values in self-attention modules as well as corresponding weight factors. In editing time, keys and values of generating image are summed with learned ones, to maintain the basic content.

In addition, some approaches~\cite{NPI,ProxEdit,NMG} aim to avoid the time-consuming optimization process in above methods~\cite{NTI,PTI,KV-Inversion}, while inheriting their reconstruction ability. Negative-Prompt Inversion (NPI)~\cite{NPI} assumes that model has similar noise predictions in adjacent steps. Under this assumption, the method verifies that optimized ``null-text'' embedding in NTI~\cite{NTI} is equivalent to source embedding in each step. Benefiting from the approximation, the method circumvents cumbersome optimization and save the inference time. Based on NPI, ProxEdit~\cite{ProxEdit} further constrains the difference between conditional and unconditional terms in semantic-unrelated region for better consistency. 

Other works~\cite{FPI,AIDI} tackle the problem by solving $\mathbf{z}_{t}^{s,*}$ as the fixed point of forward function under contractive assumption: $\mathbf{z}_{t}^{s,*}=f(\mathbf{z}_{t}^{s,*})$, where $f$ is the right-hand term of Eq.~\ref{eq:ddim-forward inversion}. In each step, Fixed-point Inversion (FPI)~\cite{FPI} performs $F_{fw}^{DI}$ for $N$ times as $\mathbf{z}_{t}^{i+1}=f(\mathbf{z}_{t}^{i}),i=0,1,..,N$, where $\mathbf{z}_{t}^{0}=\mathbf{z}_{t-1}^{N}$ and superscript $s,*$ is omitted for brevity. The operation stops when the difference of $\mathbf{z}_{t}^{i+1}$ and $\mathbf{z}_{t}^{i}$ is small enough or the number of iterations reaches upper bound. Moreover, Accelerated Iterative Diffusion Iteration (AIDI)~\cite{AIDI} further accelerates the convergence through proposed variant of Anderson acceleration algorithm.

\noindent$\bullet$
\textbf{Exact DDIM Inversion}. Instead of simulating the forward trajectory, another research line~\cite{EDICT,BDIA,Inversion-Free} establishes exact DDIM inversion. Inspired from normalizing flow models~\cite{NICE,Density-Estimation}, EDICT~\cite{EDICT} reformulates DDIM processes and tracks two associated noisy variables in each step during inversion, which can be exactly derived from each other in sampling time. Bi-Directional Integration Approximation (BDIA)~\cite{BDIA} further accelerates the process by reducing number of neural function evaluations (NFE). Specifically, the method establishes the relation of current noisy latent and several preceding ones in backward step, and reverses the equation to get exact inversion process. From a more intuitive perspective, PnP Inversion~\cite{PnP-Inversion} computes and stores the difference of $\mathbf{z}_t^{s,\prime}$ and $\mathbf{z}_t^{s,*}$ for each $t$, which is added to noisy variable in editing time.

\noindent$\bullet$
\textbf{Improvement of Editability}. For ordinary DDIM inversion trajectory, noise maps estimated in middle steps do not satisfy the statistical characteristics of pure Gaussian noise, resulting in poor editability. To address the challenge, pix2pix-zero~\cite{pix2pix-zero} and it's following work~\cite{ReGeneration} introduce auto-correlation loss to guide the inversion process, which regularizes the predicted noise to prevent falling out of distribution. From another perspective, works from~\cite{StyleDiffusion,DPL} refine the correspondence of text token and image features in cross-attention map to correct inaccurate semantic alignment. Based on NTI~\cite{NTI}, DPL~\cite{DPL} introduces additional loss terms in each denoising step. Specifically, these constraints are introduced to reduce the cosine similarity of cross-attention maps from different visual components, which alleviate attention leakage and enhances the editability.

\subsubsection{DDPM Inversion} \label{sec:ddpm-foward inversion space}
Another group of methods~\cite{SDEdit,CycleDiffusion,Edit-Friendly,Ledits,Ledits++} explore the property of DDPM inversion space. These methods simply use $F_{fw}^{DP}$ to inject randomness into image without noise estimation. Since the noise map of each step complies Gaussian distribution, these approaches outperforms in editability, while saving computation. Due to the indeterministic nature of DDPM processes, some studies~\cite{CycleDiffusion,Edit-Friendly} make effort to resist the degradation of reconstruction ability caused by injected randomness. To tackle the issue, DDPM Inversion~\cite{Edit-Friendly} encodes the difference of $\mathbf{z}_t^{s,\prime}$ and $\mathbf{z}_t^{s,*}$ into randomness term from $F_{bw}^{DP}$ given in Eq.~\ref{eq:ddpm-reverse}. Since the randomness map ($\theta_F$) contains rich visual information, the method can faithfully re-create source contents during modification, while providing high editing flexibility.

\section{\editingalgorithmbig Algorithm}
\label{sec:editing-algorithm}
As the remaining part of proposed unified framework, \editingalgorithmsmall algorithm $F_{edit}$ aims to generate final edited image based on $\Phi_I$ and $G$. The normal solution illustrated in Eq.~\ref{eq:noraml-based editing} directly incorporates $\Phi_I$ to base model and performs sampling process~\cite{DDPM,DDIM} under the guidance from $G$. It is a common practice in early works~\cite{TI,SDEdit,UniTune,HiPer} and especially for \contentfreesmall tasks~\cite{Reversion,Lego}, as evident in \tablename~\ref{tab:paper-list}. However, according to \figurename~\ref{fig:unified-framework}, $F_{edit}^{Norm}$ lacks fine-grained control on final images and fall short in achieving both content consistency and semantic fidelity. Besides, some modalities like mask and numerical input from user interfaces are difficult to process by T2I models. To address these challenges, numerous studies~\cite{P2P,Imagic,Dragon-Diffusion,DDS} intervene the ordinary backward process, as formalized in Eq.~\ref{eq:editing algorithm}. We categorize current methods into four groups: \textit{Attention-Based \editingalgorithmbig}, \textit{Blending-Based \editingalgorithmbig}, \textit{Score-Based \editingalgorithmbig} and \textit{Optimization-Based \editingalgorithmbig}, and denote them as $F_{edit}^{Attn}$, $F_{edit}^{Blend}$, $F_{edit}^{Score}$ and $F_{edit}^{Optim}$ respectively. The principles of these approaches are illustrated in bottom of \figurename~\ref{fig:unified-framework}. It's worth noting that users can use these methods in combination to leverage their different characteristics, thereby gaining better performance. In this section, we primarily discuss the mechanisms and properties of these ideas, offering a reference for algorithm selection in various editing tasks.

\begin{figure}[!t]
	\centering
    \includegraphics[width=0.98\linewidth]{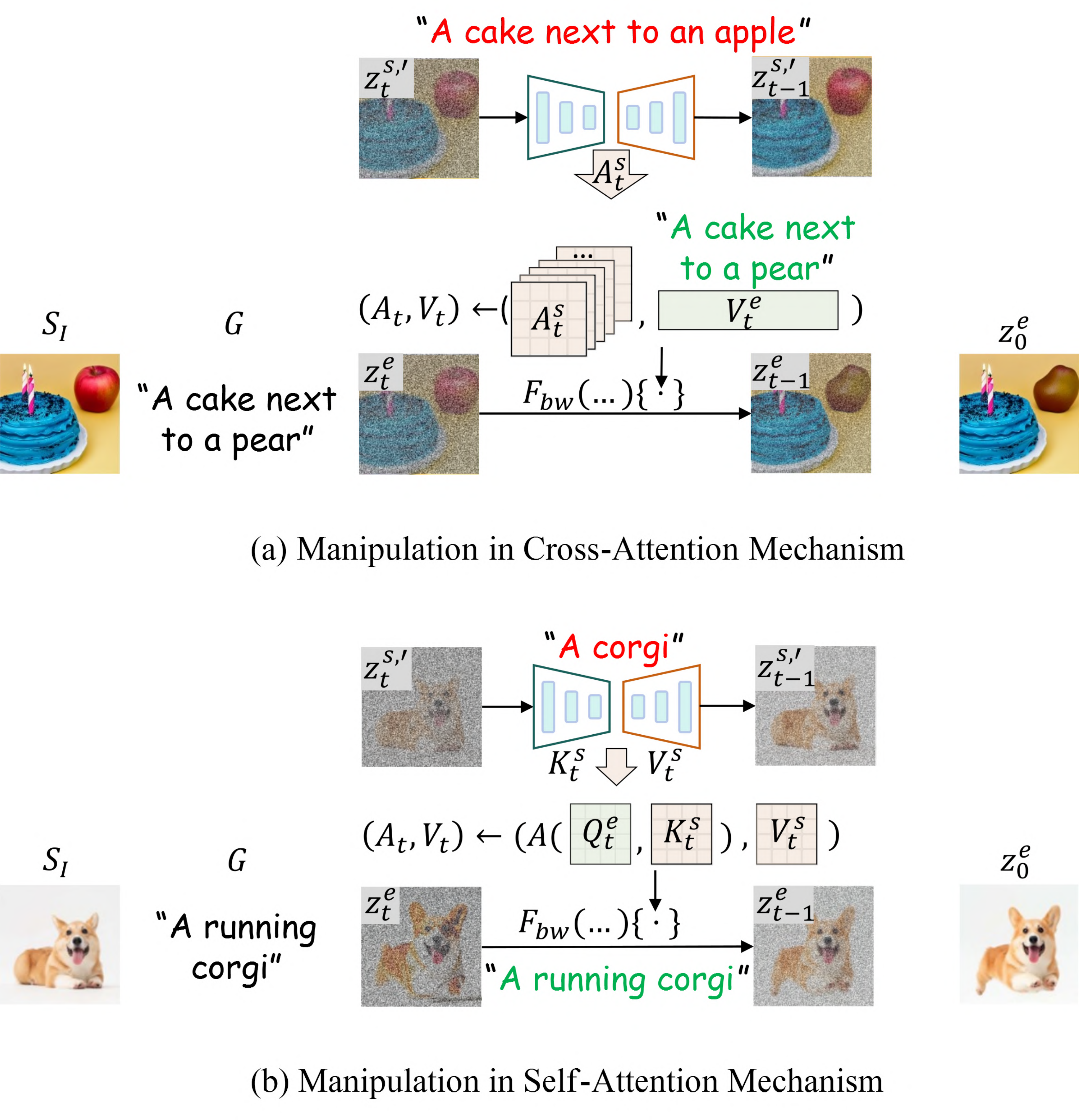}
    \caption{\textbf{Attention-Based Editing}. Illustrated methods are ~\cite{P2P,MasaCtrl}. We use red and green colors to represent source and target prompts respectively. The superscripts $s$ and $t$ denote attention features from source and editing images. $A(\cdot)$ in (b) indicates the computation of attention map.}
	\label{fig:attention-based}
\end{figure}

\subsection{Attention-Based \editingalgorithmbig} \label{sec:attention-based image editing}
A big family of methods~\cite{P2P,PnP,MasaCtrl,Cross-Image-Attention} achieve finer control over the final outcome through manipulating attention mechanisms. Formally, the process of $F_{edit}^{Attn}$ is expressed as:
\begin{equation}
\begin{aligned} \label{eq:attention-based}
~\mathbf{z_{t-1}}=F_{bw}(\mathbf{z}_t,t,\mathcal{C}\mid\Phi_I,G)\big\{ \mathcal{A}_t, \mathcal{V}_t \leftarrow \hat{\mathcal{A}}_t, \hat{\mathcal{V}}_t\big\},
\end{aligned}
\end{equation}
Where the item inside braces represents the replacement of attention map $\mathcal{A}_t$ and value $ \mathcal{V}_t$ with altered ones: $\hat{\mathcal{A}}_t$ and $\hat{\mathcal{V}}_t$. We present several representative works in \figurename~\ref{fig:attention-based} to demonstrate their manipulation approaches in different scenarios. Due to the dissimilar properties of cross-attention and self-attention, we separately discuss them in following.

\subsubsection{Manipulation in Cross-Attention Mechanism}
For T2I models~\cite{StableDiffusion,Imagen}, cross-attention enables the communication between image features and text embedding, bridging the gap between natural language and visual content. Specifically, in cross-attention, attention map indicates the correlation of text token and image region, while value identifies semantic content. Next, we introduce several manipulation schemes.

\noindent$\bullet$
\textbf{Injection of Cross-Attention Features}. A group of studies~\cite{P2P,Custom-Edit} intend to inject cross-attention map or value from other sources into backward process for fine-grained controllability in semantic layout. Prompt-to-Prompt (P2P)~\cite{P2P} employs attention maps from reconstruction path ($\hat{\mathcal{A}}_t$) to manipulate corresponding ones in editing path for various purposes. Several examples of object manipulation are illustrated in \figurename~\ref{fig:unified-framework} and \figurename~\ref{fig:attention-based}. Through injection, while keeping value from editing prompt, the method maintains the basic structure from source image. Based on P2P, Custom-Edit~\cite{Custom-Edit} combines ideas from both tuning-based~\cite{Custom-Diffusion} and forward-based inversion methods~\cite{NTI} to replace the object with reference one given in examplar image. Specifically, the method performs the same injection strategy from P2P, while using the identifier token of personalized object in target prompt to achieve object replacement. Object-Shape Variation~\cite{Localize-Object-Shape} proposes prompt-mixing to adjust the shape of object, while keeping other elements unchanged. Specifically, it injects value from another prompt ($\hat{\mathcal{V}}_t$) in middle denoising stage, which is responsible for modifying the shape-related semantic.

\noindent$\bullet$
\textbf{Adjustment of Attention Score}. In cross-attention map, activation magnitude reflects the effect extent of each text token. A group of methods~\cite{P2P,FOI} leverage the property to adjust the influence of target words in edited image. For example, P2P re-weights attention score of object token to control it's existence. Since high attention score in irrelevant region often degrades the quality, FoI~\cite{FOI} exploits the idea for preventing over-editing issue. 

\subsubsection{Manipulation in Self-Attention Mechanism}
Self-attention captures spatial correlation in attention map, and enhances communication between image tokens belonging to the same semantic, which accelerates the process of completing a holistic concept. Similar with injection scheme of cross-attention, some methods~\cite{PnP,FPE,PhotoSwap,DreamMatcher} leverage attention map or value from other trajectories to edit in a finer approach. For object replacement, Plug-and-Play (PnP)~\cite{PnP} replaces self-attention maps with ones from reconstruction path ($\hat{\mathcal{A}}_t$) to keep image layout. Free-Prompt Editing (FPE)~\cite{FPE} investigates the effectiveness of self-attention in text-guided image editing tasks, while boosting the performance in several scenarios through attention injection. For retaining details of customized subject, DreamMatcher~\cite{DreamMatcher} uses warped value from reference path ($\hat{\mathcal{V}}_t$) to spatially align self-attention map and value from different sources. 

In addition, other methods~\cite{MasaCtrl,TF-ICON,HD-Painter,DesignEdit} manipulate self-attention map locally to refine the spatial correspondence. Among them, TF-ICON~\cite{TF-ICON} merges self-attention maps from input images, to achieve harmonious image composition. HD-Painter~\cite{HD-Painter} adjusts attention score to prevent the influence from irrelevant pixels over inpainted region. Similarly, for layer-wise editing, DesignEdit~\cite{DesignEdit} uses the idea to get the clean background through constraining the activation values in object area. 

Another line of research~\cite{MasaCtrl,TIC,Cross-Image-Attention,FEC} queries to other contexts for controlling the appearance. As demonstrated in \figurename~\ref{fig:attention-based}, MasaCtrl~\cite {MasaCtrl} introduces mutual self-attention mechanism to achieve non-rigid editing, where key and value are sourced from reconstruction path. By querying to reference context, MasaCtrl maintains object appearance while complying non-rigid guidance, such as change of pose or action. Furthermore, due to the imperfect reconstruction of ordinary DDIM inversion space~\cite{NTI,NPI}, TIC~\cite{TIC} boosts the performance by directly utilizing the features from inversion trajectory. Inspired from mutual computation, work from~\cite{Cross-Image-Attention} proposes cross-image attention for appearance transfer between subjects with similar semantic (like zebra and horse), where key and value are calculated from appearance image. Similarly, other works~\cite{Z-star,StyleInjection} leverage the idea for style change, and obtain key and value from style image.

\subsection{Blending-Based \editingalgorithmbig} \label{sec:interpolation-based image editing}
A number of methods~\cite{Imagic,TF-ICON,DAC,Blended-Latent-Diffusion,Zone} represent source image and editing target in the same space, while seeking for optimal intermediate state that retains desired contents and reflects purpose simultaneously. $F_{edit}^{Blend}$ is formalized as:
\begin{equation}
\begin{aligned} \label{eq:interpolation-based}
\mathbf{z_{t-1}}=F_{bw}(\mathbf{z}_t,t,\mathcal{C}\mid\Phi_I,G)\big\{ f_t \leftarrow \alpha f_t^{src}+ \beta f_t^{tgt}\big\},
\end{aligned}
\end{equation}
where the item inside braces indicates the fusion of $f_t^{src}$ and $f_t^{tgt}$ using respective blending factors $\alpha$ and $\beta$. $f_t$ is blended features~\cite{Blended-Latent-Diffusion,Imagic} or parameters~\cite{DAC} specified in methods. In this section, we organize methods based on their blending spaces, and illustrate several representative works~\cite{Forgedit,Blended-Latent-Diffusion} in \figurename~\ref{fig:interpolation-based}.

\subsubsection{Blending in Spatial Space}
Blending in spatial space is a common practice for compositing visual elements into a single image, which assembles the contents from source image and generating one for local editing. Based on multi-step backward process, a group of works~\cite{High-Resolution-Image-Editing,Pfb-diff,Tuning-Free-Image-Customization} explore to blend diffusion features for harmonious results.
\begin{figure}[!t]
	\centering
    \includegraphics[width=1.0\linewidth]{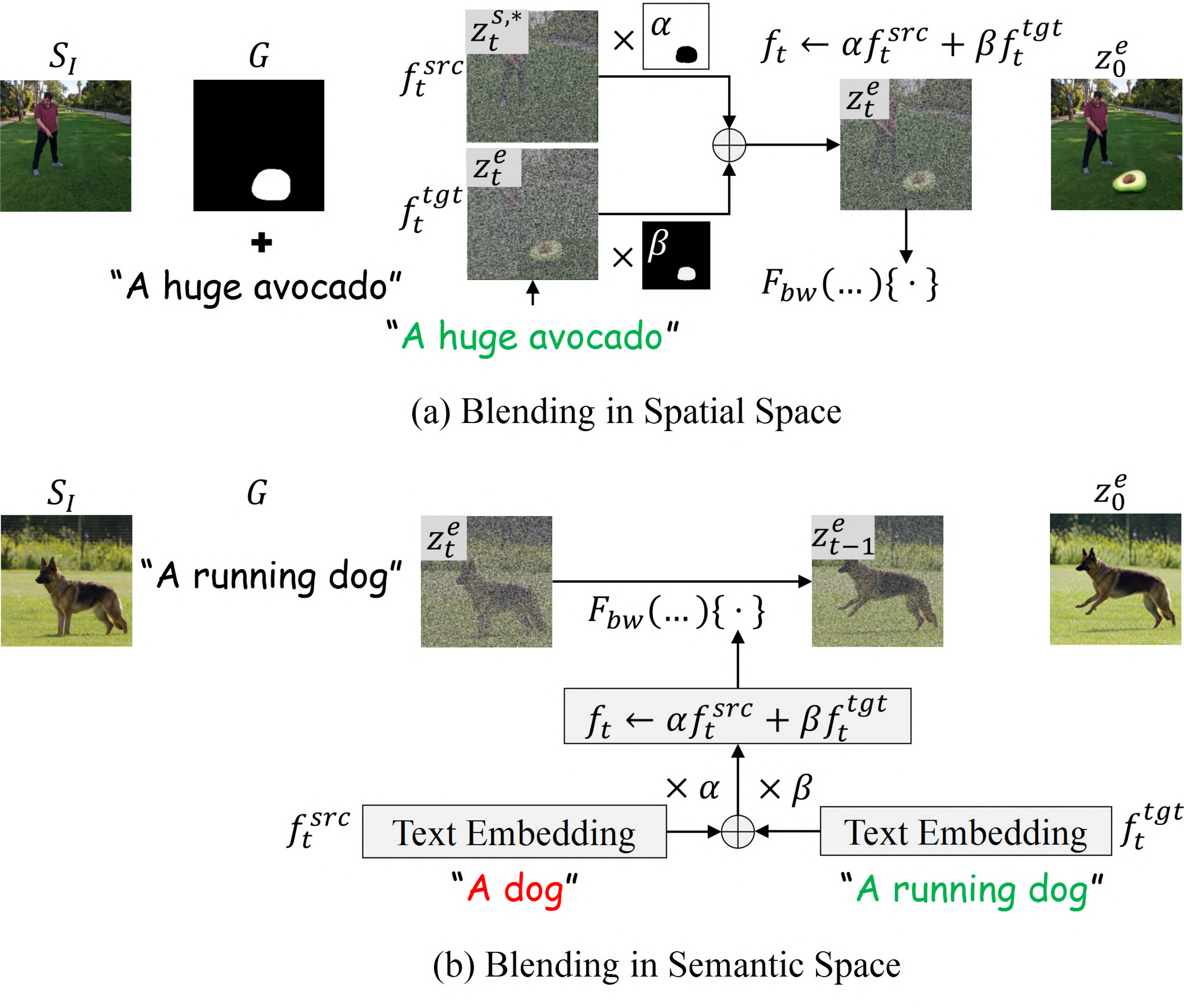}
    \caption{\textbf{Blending-Based Editing}. Illustrated methods are~\cite{Blended-Latent-Diffusion,Forgedit}. We use red and green colors to represent source prompt and editing target respectively.}
	\label{fig:interpolation-based}
\end{figure}

\noindent$\bullet$
\textbf{Blending in Noisy Latent Space}. According to literature~\cite{PnP,MasaCtrl}, noisy latent encodes spatial characteristics of the generating image, like appearance and structure. In this situation, $f_t^{src}$ and $f_t^{tgt}$ represent for intermediate noisy variables from source and generating images respectively. Some methods~\cite{TF-ICON,Blended-Latent-Diffusion,Diff-Edit} use either the user-provided mask~\cite{Blended-Latent-Diffusion,High-Resolution-Image-Editing} or estimated one~\cite{P2P,Localize-Object-Shape,Diff-Edit,Zone,OIR} to employ blending. For inpainting task, users require to provide mask condition to indicate the interesting region. Borrowing the idea from previous work~\cite{Blended-Diffusion}, Blended Latent Diffusion (BLD)~\cite{Blended-Latent-Diffusion} blends noisy latents from editing path and DDPM forward trajectory of source image, as depicted in top of \figurename~\ref{fig:interpolation-based}. Moreover, High-Resolution Blended Diffusion~\cite{High-Resolution-Image-Editing} further improves the quality though super-resolution model. Meanwhile, similar with RePaint~\cite{RePaint}, it applies blending process and Eq.~\ref{eq:ddpm-forward-step} iteratively in each denoising step, resulting in more harmonious and consistent result. In addition, Differential Diffusion~\cite{Differential-Diffusion} allows users to provide change map, which presents the editing strength in each location for smooth inpainting. Specifically, instead of using fixed mask, the method calculates dynamic blending factors according to current step and control map, adjusting the fusion smoothness for higher coherency. DreamEdit~\cite{DreamEdit} exploits the idea to inpaint customized subject into designated area. 

Recently, some methods~\cite{DesignEdit,OIR} blend the noisy latents from multiple trajectories, to assemble different contents. Since the optimal inversion step of each target is dissimilar, OIR~\cite{OIR} constructs multiple editing paths accordingly and fuse them to integrate their effects in final image.

Other methods~\cite{P2P,MasaCtrl,Localize-Object-Shape,Diff-Edit,Zone} leverage the idea to prevent methods modifying semantic-unrelated content. Some of these works~\cite{P2P,MasaCtrl} use cross-attention map to estimate editing region. Another line of research~\cite{Diff-Edit,Watch-Your-Step,ProxEdit} predicts based on the difference of noise maps, where elements below a certain threshold stand for the region to be preserved. To avoid over-editing issue in object replacement, DiffEdit~\cite{Diff-Edit} computes the discrepancy of estimated noises that are guided by source and target prompts respectively to infer the blending map. Since background pixels make little change in terms of noise value, the map coarsely indicates editing area. Differently, other methods~\cite{ProxEdit,Watch-Your-Step} compute the divergence of conditional and unconditional terms to identify fusion factors.

\noindent$\bullet$
\textbf{Blending in Layer-Level Feature Space}. Instead of blending noisy variables, another group of methods~\cite{MasaCtrl,Localize-Object-Shape,TF-ICON,DreamMatcher,Pfb-diff,Tuning-Free-Image-Customization} operate in layer-level feature space for more seamless and coherent result. PFB-Diff~\cite{Pfb-diff} introduces blending modules to fuse the features from certain transformer blocks for object replacement. Some of other methods~\cite{MasaCtrl,Localize-Object-Shape} fuses self-attention features for local modification. For example, MasaCtrl~\cite{MasaCtrl} blends the features from editing and reconstruction paths to preserve the background content. For zero-shot customization, Tuning-Free Image Customization~\cite{Tuning-Free-Image-Customization} blends output of self-attention modules from different trajectories with time-variant blending factors to combine scene image and personalized subject.

\subsubsection{Blending in Semantic Space} 
Another study line~\cite{Forgedit,Imagic,PTI,DAC} represents $f_{src}^t$ and $f_{tgt}^t$ in the semantic space for highly flexible editing. 

\noindent$\bullet$
\textbf{Blending in Textual Space}. Since textual embedding encompasses rich information, some approaches~\cite{Imagic,PTI,Forgedit} encode the source image into textual space, and blend it with the target embedding. For non-rigid editing, Imagic~\cite{Imagic} first inverts the input through tuning the embedding of target prompt with finite steps. This prevents the target embedding deviating far from it's original value, thereby keeping linear property. The method then optimizes parameters from backbone network to improve consistency, which is not satisfied in first step. In editing time, Imagic interpolates updated target embedding and it's original value to get final result. In contrast, Forgedit~\cite{Forgedit} jointly optimizes the source embedding and base model for training efficiency. As demonstrated in \figurename~\ref{fig:interpolation-based}, through merging the source and target embeddings, Forgedit exhibits a good balance in content consistency and semantic fidelity. Borrowing the idea from NTI~\cite{NTI}, PTI~\cite{PTI} optimizes the source embedding to correct imperfect DDIM inversion space~\cite{NTI}, which is interpolated with target embedding in editing time. From another perspective, since denoising stages are responsible for distinct visual contents~\cite{P2P,Localize-Object-Shape,Prospect}, work from~\cite{Uncovering-Disentanglement} seeks for optimal blending factors in each step through applying semantic and content losses~\cite{Stylegan-Nada,Perceptual-Loss} on editing image, balancing the ratio of preserved and modified contents. 

\noindent$\bullet$
\textbf{Blending in Parameter Space}. A handful of studies~\cite{Forgedit,DAC} operate in parameter space. For mitigating the overfitting issue caused by tuning on a single image, Forgedit replaces a part of modules with corresponding ones from original model to preserve generative capability. Inspired from counterfactual inference framework, Doubly Abductive Counterfactual (DAC)~\cite{DAC} optimizes LoRA~\cite{Lora-Image} modules to encode source image and editing target. By adjusting the weight factor, DAC makes a equilibrium between modification and preservation of source contents.

\subsection{Score-Based \editingalgorithmbig}\label{sec:score-based image editing}
\begin{figure}[!t]
	\centering
    \includegraphics[width=0.9\linewidth]{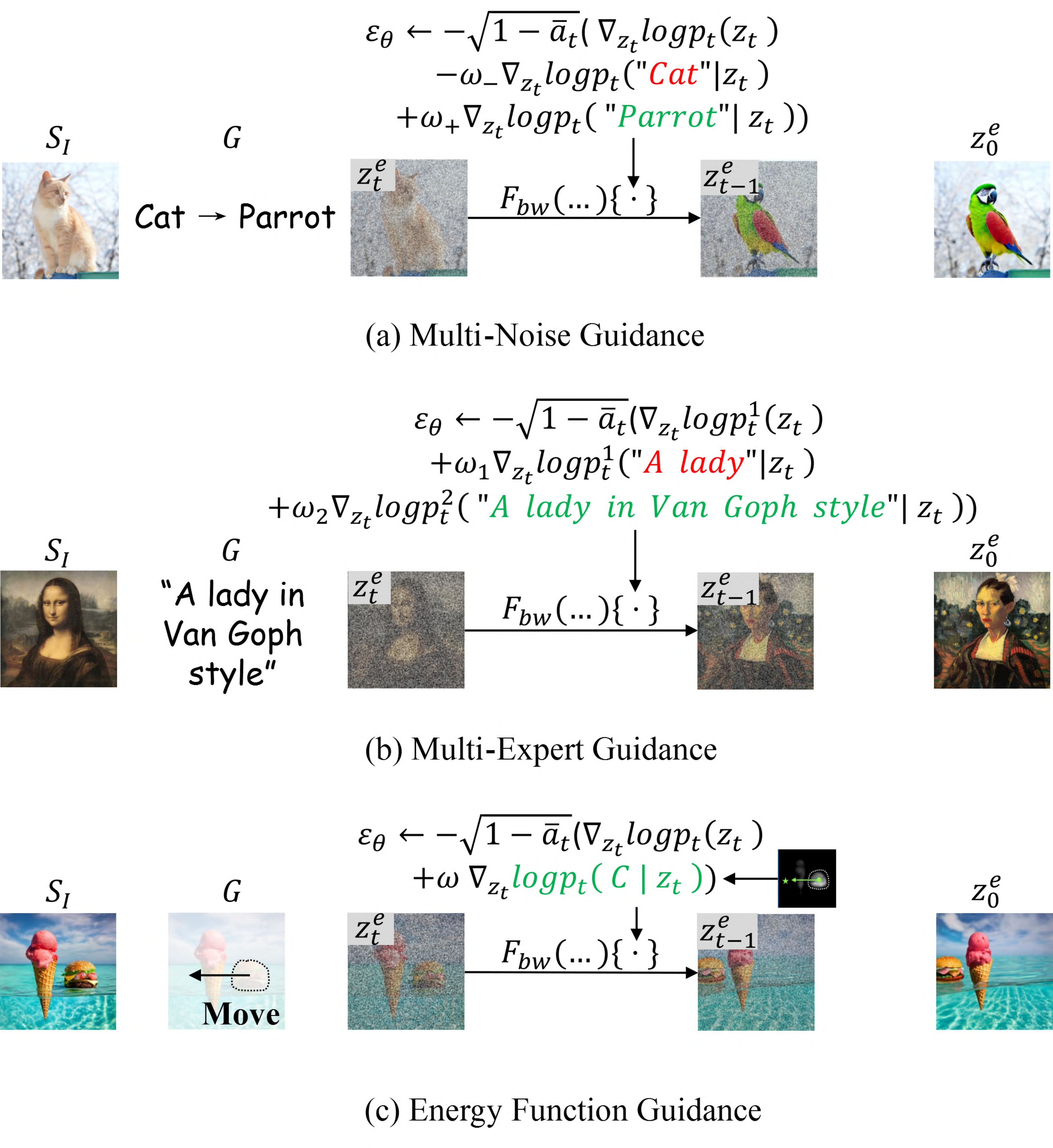}
    \caption{\textbf{Score-Based Editing}. Illustrated methods are ~\cite{SEGA,SINE,Self-Guidance}. We use red and green colors to represent source prompt and editing target respectively. The superscripts $1,2$ of $logp_t(\mathcal{C}_i \mid \mathbf{z}_t)$ in (b) indicate that they are predicted by different networks. In (c), the small image next to $logp_t(\mathcal{C} \mid \mathbf{z}_t)$ represents the energy function of object translation. }
	\label{fig:score-based}
\end{figure}

For image generation, conditional input endows model with the controllability of created contents~\cite{classifier-guidance,classifier-guidance}. From score-based perspective, conditional score function can be extended to following equation under the assumption of independence. 
\begin{equation}
\begin{aligned} \label{eq:multi-score-function}
\nabla_{\mathbf{z}_t} \log p_t\left(\mathbf{z}_t\mid  \mathcal{C}_1, \mathcal{C}_2,..., \mathcal{C}_{N_c} \right) \propto& \nabla_{\mathbf{z}_t} \log p_t \left( \mathbf{z}_t \right)\\
&+ \Sigma_{i=1}^{N_c} \nabla_{\mathbf{z}_t} \log p_t \left(\mathcal{C}_i \mid  \mathbf{z}_t \right),
\end{aligned}
\end{equation}
where $N_c$ is the number of conditions, and $\log p_t\left( \mathcal{C}_i \mid \mathbf{z}_t \right)$ could be modeled by classifier-guidance~\cite{classifier-guidance} or classifier-free guidance~\cite{classifier-free}. The score-based perspective inspires many studies~\cite{SEGA,SINE,Dragon-Diffusion} to gather information from auxiliary distributions, providing directional guidance for distinct editing targets. $F_{edit}^{Score}$ is formalized as:
\begin{equation*}
\begin{aligned}
\mathbf{\mathbf{z}_{t-1}}=F_{bw}(\mathbf{z}_t,t,\mathcal{C} \mid \Phi_I,G)&\big\{ \varepsilon_{\theta} \leftarrow -\sqrt{1-\bar{\alpha}_t}( \nabla_{\mathbf{z}_t} \log p_t \left( \mathbf{z}_t \right) \big.\\
&\big. +\Sigma_{i=1}^{N_c} \omega_i \nabla_{\mathbf{z}_t} \log p_t\left(\mathcal{C}_i\mid  \mathbf{z}_t \right) )\big\},
\end{aligned}
\end{equation*}
where the item inside braces represents the noise map is calculated by aggregating effects from multiple conditions and transforming the score to noise through Eq.~\ref{eq:noise-score-version}. $\omega_i$ is guidance scale. We categorize the methods based on the origin of $\nabla_{\mathbf{z}_t} \log p_t\left( \mathcal{C}_i \mid \mathbf{z}_t \right)$, and present several representative works in \figurename~\ref{fig:score-based}.

\noindent$\bullet$
\textbf{Multi-Noise Guidance}. A group of works~\cite{SEGA,Stable-Artist,Ledits,Ledits++} aggregate multiple noise estimates from distinct text prompts to achieve multi-target editing, where $\nabla_{\mathbf{z}_t} \log p_t\left( \mathcal{C}_i \mid \mathbf{z}_t \right)$ is calculated through $\varepsilon_\theta(\mathbf{z}_t,t,\mathcal{C}_i)-\varepsilon_\theta(\mathbf{z}_t,t,\varnothing)$ based on Eq.~\ref{eq:noise-score-version} and Bayes rule. For each editing target,. SEGA~\cite{SEGA} and it's following works~\cite{Ledits,Ledits++} calculate conditional terms to inject or remove the designated concept in source image. The process is demonstrated in the top of \figurename~\ref{fig:score-based}. Furthermore, since image patches have different responses to each editing target, these methods compute patch-level guidance scale accordingly. Through multi-noise guidance, they accomplish several goals in a single backward process.

\noindent$\bullet$
\textbf{Multi-Expert Guidance}. Since diffusion models fine-tuned on different datasets have dissimilar domain knowledge, several methods~\cite{DCO,SINE} aim to inherit generative capacities from multiple experts. In order to avoid overfitting issue of tuning on a single image, SINE~\cite{SINE} employs two generation models. As illustrated in the middle of \figurename~\ref{fig:score-based}, it first performs tuning-based algorithm~\cite{DreamBooth} to invert source image, and then integrates conditional terms estimated from fine-tuned model and original one to simultaneously achieve consistency and semantic fidelity. Similarly, for alleviating language drift and catastrophic neglecting issues in customization, DCO~\cite{DCO} introduces reward guidance to combine noise maps from base model and customization expert.

\noindent$\bullet$
\textbf{Energy Function Guidance}. Borrowing the idea from energy-based models~\cite{EBM}, which assign low energy to in-distribution data, another research line~\cite{NMG,Self-Guidance,MagicRemover,FreeControl,DiffEditor,HD-Painter} simulates $\log p_t\left(\mathcal{C}_i\mid  \mathbf{z}_t \right)$ through established energy function and steers the editing direction to reduce system energy. For example, NMG~\cite{NMG} designs energy function as the difference of noisy latents from inversion and editing trajectories to prevent over-editing issue. Benefiting from the encoded structure and appearance information in diffusion features, several works~\cite{Self-Guidance,Dragon-Diffusion,MagicRemover,FreeControl} intend to provide spatial guidance. For spatial transformation, Self-Guidance~\cite{Self-Guidance} elaborately devises energy function for each scenario. An example for translation is depicted in bottom of \figurename~\ref{fig:score-based}. Specifically, the method minimizes the energy through aligning the object characteristics before and after transformation based on cross-attention maps and local features, which are responsible for location and appearance respectively. Similarly, DragonDiffusion~\cite{Dragon-Diffusion} maximizes cosine similarity of image features inside transformation-related regions. In addition, MagicRemover~\cite{MagicRemover} refines the idea in object removal task. By constraining the activation of cross-attention map, it erases the target from source image. FreeControl~\cite{FreeControl} leverages the idea in image translation. It constructs energy functions for both structure and appearance guidance to maintain the layout from source image while generating finer details.

\subsection{Optimization-Based \editingalgorithmbig} \label{sec:optimization-based image editing}
\begin{figure}[!t]
	\centering
    \includegraphics[width=0.98\linewidth]{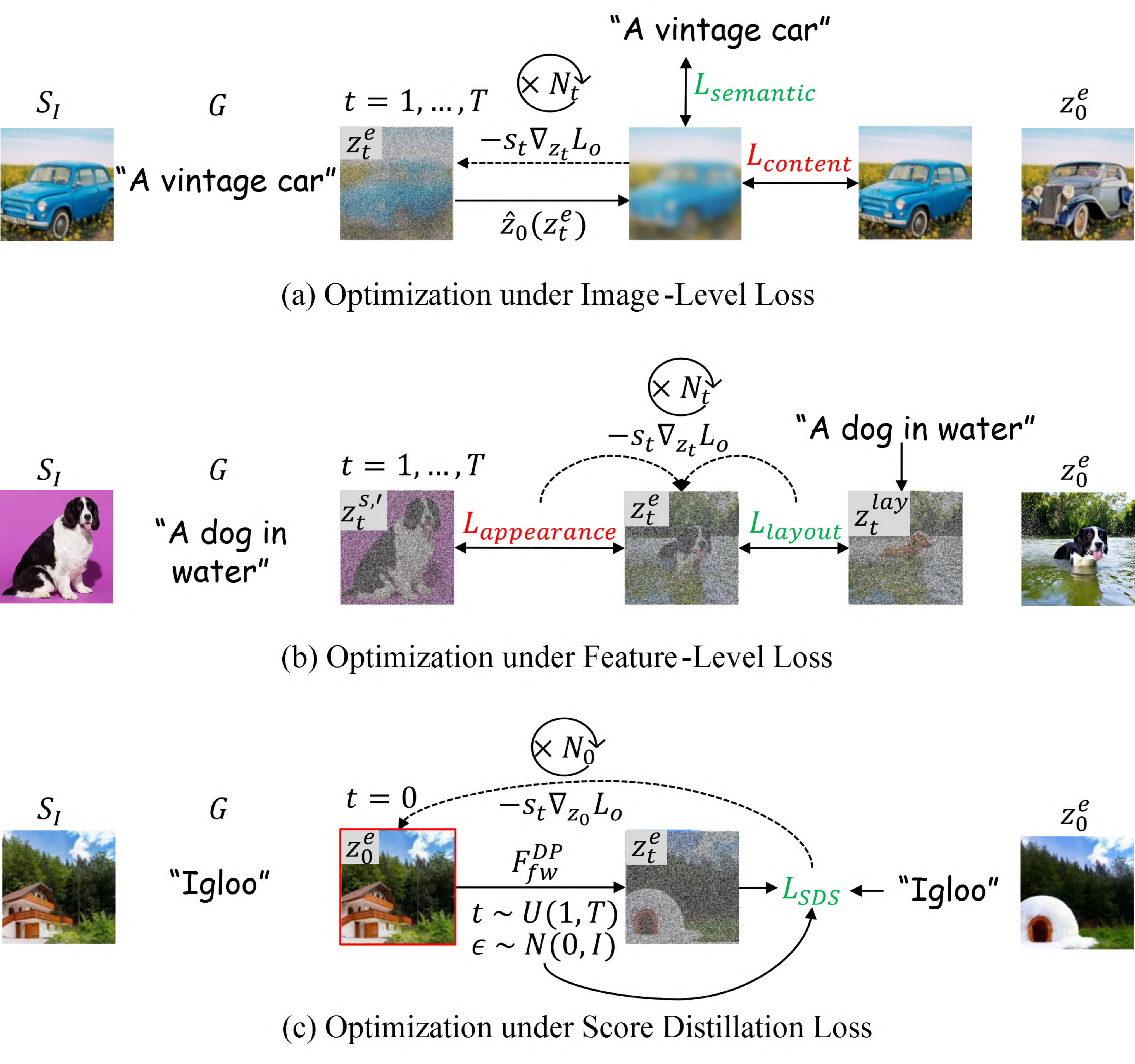}
    \caption{\textbf{Optimization-Based Editing}. Illustrated methods in (a) and (b) are~\cite{Region-Aware,Pick-and-Draw}, and (c) demonstrates the underlying logic of editing based on score distillation loss~\cite{DreamFusion}. We use red and green colors to represent the constraints from source image and editing target respectively. $\hat{\mathbf{z}}_0(\mathbf{z}_t^e)$ in (a) denotes the predicted clean image based on $\mathbf{z}_t^e$. $\mathbf{z}_t^{lay}$ in (b) originates from template trajectory that is generated with random noise and target prompt. The red border in (c) indicates $\mathbf{z}_0^e$ is initialized by source image.}
	\label{fig:optimize-based}
\end{figure}

Similar with supervision learning, a family of works~\cite{Drag-Diffusion,Region-Aware,DDS} construct proper objectives to guide editing process. Significantly, the optimized parameters could be any variables that influence noise estimates, such as network parameters~\cite{Diffusion-Clip}, text embedding~\cite{Uncovering-Disentanglement}, noisy latent~\cite{Disentangled-Style-Content}, and so on~\cite{drag-your-noise,EBCA}. Among them, we focus on studies that update $\mathbf{z}_t$ due to the wide applicability. In our discussion, $F_{edit}^{Optim}$ is expressed as:
\begin{equation}
\begin{aligned} \label{eq:score-based}
\mathbf{\mathbf{z}_{t-1}}=F_{bw}(\mathbf{z}_t,t,\mathcal{C}\mid& \Phi_I,G)\\
&\big\{ \mathbf{z}_t \leftarrow \left[\mathbf{z}_t-s_t\nabla_{\mathbf{z}_t} \mathcal{L}_o\right]_{i=1}^{N_t} \big\},~\mathrm{st}.~t \in \tau,
\end{aligned}
\end{equation}
where the item inside braces indicates $\mathbf{z}_t$ is replaced with updated one that is optimized with $\mathcal{L}_o$ for $N_t$ iterations. $s_t$ is gradient scale. $\tau=\left[\tau_1,\tau_2...,\tau_{dim\left(\tau \right)} \right]$ is the subset of time series, which is specified in methods. Since both $F_{edit}^{Optim}$ and methods based on energy function~\cite{Self-Guidance,DiffEditor,Dragon-Diffusion} aim to minimize the loss or energy, they exhibit similar characteristics. Some representative approaches~\cite{Pick-and-Draw,DDS,Disentangled-Style-Content} are shown in \figurename~\ref{fig:optimize-based}, and we organize these methods based on their optimization objectives.

\noindent$\bullet$
\textbf{Optimization under Image-Level Loss}. Several methods~\cite{Uncovering-Disentanglement,Region-Aware,Disentangled-Style-Content} use training objectives~\cite{CLIP,LPIPS,Stylegan-Nada,Perceptual-Loss} that are commonly used in image space to preserve maximal amount of details while complying with the guidance. Early works~\cite{Uncovering-Disentanglement,Diffusion-Clip} necessitate applying a intact backward process to get edited image. Different from these works, as demonstrated in top of \figurename~\ref{fig:optimize-based}, some studies~\cite{Region-Aware,Disentangled-Style-Content} perform image-level loss on predicted clean image, due to it reflects coarse visual content. For object replacement, Region-Aware Diffusion Model (RDM)~\cite{Region-Aware} computes CLIP loss~\cite{CLIP} between predicted image and target text, while confining the change of semantic-unrelated region through content loss~\cite{LPIPS} to optimize current latent.

\noindent$\bullet$
\textbf{Optimization under Feature-Level Loss}. Another line of method~\cite{Drag-Diffusion,Pick-and-Draw,drag-your-noise,FreeDrag} establishes feature-level constraints for special editing purposes. To achieve local deformation with the guidance of point dragging~\cite{Drag-GAN}, DragDiffusion~\cite{Drag-Diffusion} builds spatial objectives to optimize initial noisy latent. Like in GAN-based counterpart~\cite{Drag-GAN}, DragDiffusion alternately performs motion supervision and point tracking in each iteration. Specifically, motion supervision constrains the difference between features of original point and the next moving point, where the gradient encourages the correct deformation in noisy variable. After moving all points to corresponding destinations in initial latent, the sampling process can generate deformed image. For zero-shot customization, Pick-and-Draw~\cite{Pick-and-Draw} proposes to minimize the optimal transport cost of features from generating and reference images in each denoising step, ensuring appearance consistency with personalized subject. Besides, the method further introduces the attention loss~\cite{Break-A-Scene,DPL} to align with cross-attention maps from template image that is generated under text guidance, providing high flexibility in controlling image layout. The process of the method is illustrated in middle of \figurename~\ref{fig:optimize-based}.

\noindent$\bullet$
\textbf{Optimization under Score Distillation Loss}. Borrowing the idea from Score Distillation Loss (SDS) in 3D realm~\cite{DreamFusion}, a group of methods~\cite{DDS,CDS,Ground-A-Score} directly perform optimization on editing image. Specifically, ~\cite{DreamFusion} calculates SDS as conditional diffusion loss using the pseudo image created from generator (NeRF\cite{NeRF}) and updates generator's parameters through gradient descent. Essentially, this loss aligns the distributions of base model~\cite{StableDiffusion} and image generator under specific text condition. For image editing~\cite{CDS,Ground-A-Score}, as demonstrated in bottom of \figurename~\ref{fig:optimize-based}, the generator is treated as editing image, which is initialized with source image for consistency. Since directly applying SDS is prone to cause blurry results with loss of details, Delta Denoising Score (DDS)~\cite{DDS} decomposes SDS gradient to desired and undesired components separately, where the latter causes unwanted artifacts. Meanwhile, it argues that the undesired term can be modeled as the SDS gradient from source image. Therefore, the method substracts the undesired gradient from total one for high-quality results. Based on DDS, CDS~\cite{CDS} further introduces Contrastive Unpaired Translation (CUT) loss for structure consistency. In addition, Ground-a-Score~\cite{Ground-A-Score} enhances the idea to support multi-region editing. It applies DDS loss on designated image areas to only optimize local pixels and keep others unchanged.

\section{Design Space within Unified Framework} \label{sec:design-space}
In this section, we summarize the properties of each \editingalgorithmsmall algorithm and provide a design space within the proposed framework. To this end, we conduct two experiments. The first one illustrates the applications of different combinations of inversion and editing algorithms in multimodal-guided editing tasks. The second experiment compares state-of-the-art methods to demonstrate the strengths and weaknesses of distinct combinations in various text-driven editing scenarios. Our experiments are conducted on NVIDIA A100 GPUs, and use the pre-trained weights from StableDiffusion V1.5~\cite{StableDiffusion}.

\subsection{Applications in Multimodal-Guided Editing}
\begin{figure}[!t]
	\centering
    \includegraphics[width=1\linewidth]{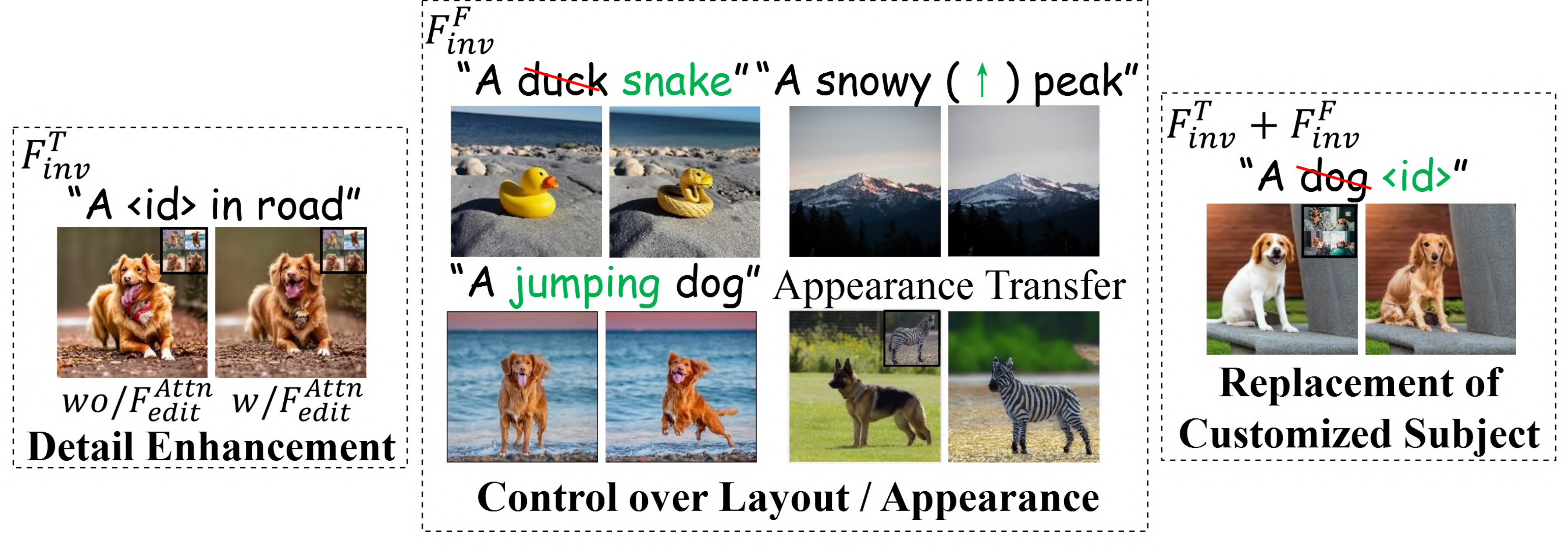}
    \caption{\textbf{Combination with attention-based editing}. Evaluated methods are \scalebox{0.8}{$F_{inv}^T$}:~\cite{DreamMatcher}, \scalebox{0.8}{$F_{inv}^F$}:~\cite{NTI,MasaCtrl,Cross-Image-Attention} and \scalebox{0.8}{$F_{inv}^T+F_{inv}^F$}:~\cite{PhotoSwap}.}
	\label{fig:exp1-1}
\end{figure}

\begin{figure}[!t]
	\centering
    \includegraphics[width=0.8\linewidth]{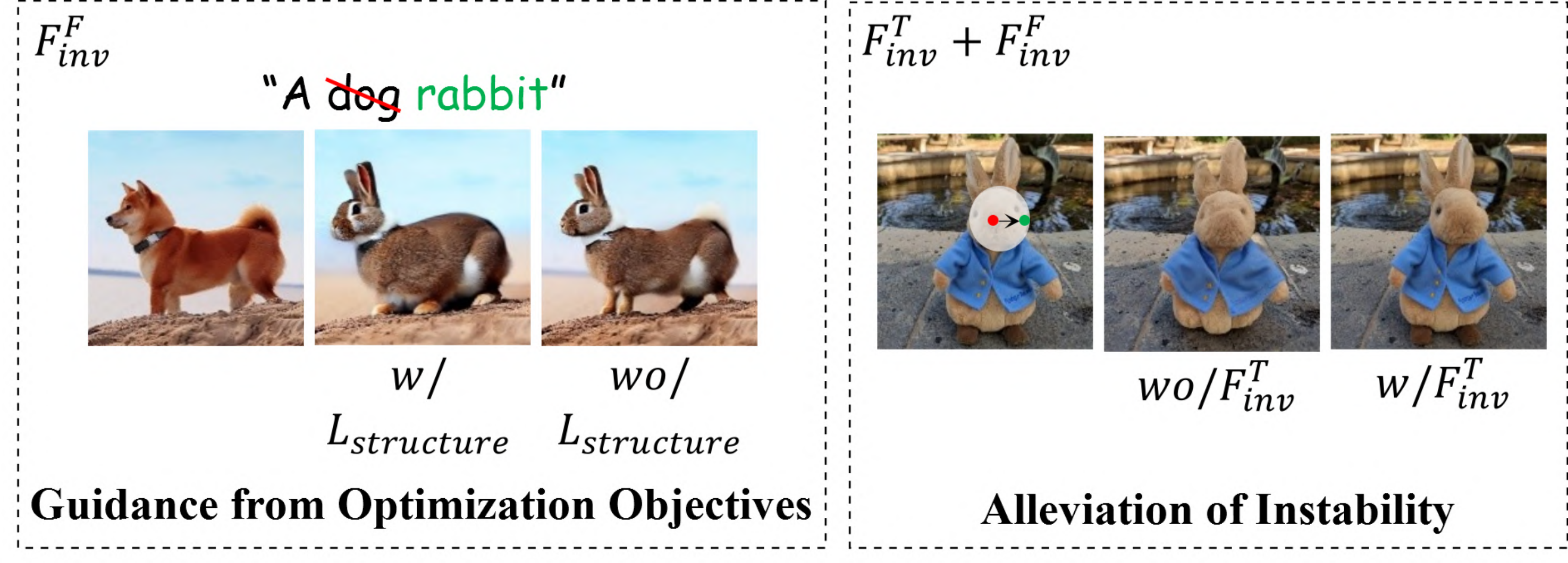}
    \caption{\textbf{Combination with optimization-based editing}. Evaluated methods are \scalebox{0.8}{$F_{inv}^F$}:~\cite{CDS} and \scalebox{0.8}{$F_{inv}^T+F_{inv}^F$}:~\cite{Drag-Diffusion}.}
	\label{fig:exp1-4}
\end{figure}

In following, we summarize the characteristics of each \editingalgorithmsmall algorithm and present evaluated results in \figurename~\ref{fig:exp1-1} - \figurename~\ref{fig:exp1-4} to illustrate the applicable scenarios of each combination. The data for evaluation are sourced from ~\cite{DreamBooth,FFHQ,Imagic,Custom-Diffusion,Cross-Image-Attention}.

\begin{figure}[!t]
	\centering
    \includegraphics[width=1\linewidth]{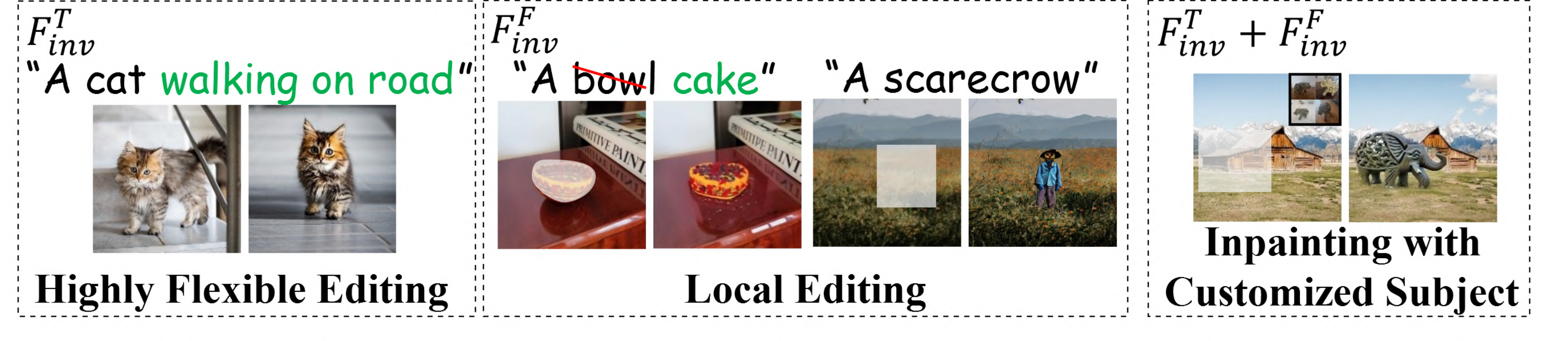}
    \caption{\textbf{Combination with blending-based editing}. Evaluated methods are \scalebox{0.8}{$F_{inv}^T$}:~\cite{Forgedit}, \scalebox{0.8}{$F_{inv}^F$}:~\cite{NTI,Blended-Latent-Diffusion} and \scalebox{0.8}{$F_{inv}^T+F_{inv}^F$}:~\cite{DreamEdit}.}
	\label{fig:exp1-2}
\end{figure}
\begin{figure}[!t]
	\centering
    \includegraphics[width=0.93\linewidth]{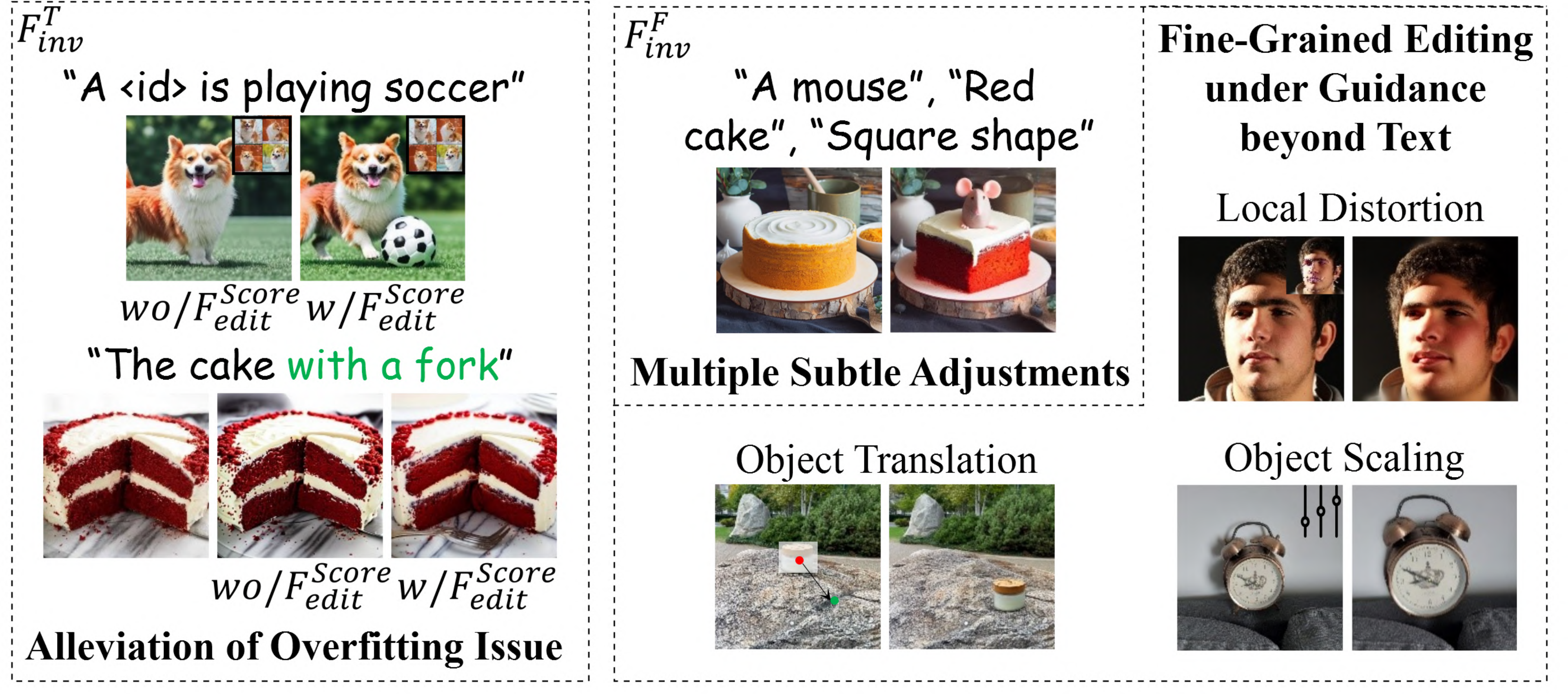}
    \caption{\textbf{Combination with score-based editing}. Evaluated methods are \scalebox{0.8}{$F_{inv}^T$}:~\cite{SINE}, \scalebox{0.8}{$F_{inv}^F$}:~\cite{Ledits++,Dragon-Diffusion}.}
	\label{fig:exp1-3}
\end{figure}

\noindent$\bullet$
\textbf{Attention-Based Editing}. Methods based on $F_{edit}^{Attn}$ manipulate attention maps or value in editing path, offering flexible control over the image layout and object appearance. (\romannumeral 2). $F_{inv}^{T}+F_{edit}^{Attn}$. For customization tasks, ~\cite{VICO,DreamMatcher} employ mutual self-attention for enhancement of subject details. (\romannumeral 2). $F_{inv}^{F}+F_{edit}^{Attn}$. Numerous methods~\cite{P2P,PnP,NTI,ProxEdit} inject attention maps from source image to control the semantic layout of the edited image. Other works~\cite{MasaCtrl,TIC,Cross-Image-Attention,Z-star} use mutual self-attention to control the object appearance in tasks like non-rigid editing and appearance transfer. (\romannumeral 3). $F_{inv}^{T}+F_{inv}^{F}+F_{edit}^{Attn}$. Some studies ~\cite{Custom-Edit,PhotoSwap} simultaneously employ tuning-based and forward-based inversion approaches to inject image conditions into the source image, such as replacing the object in source image with customized subject. We select ~\cite{DreamMatcher,NTI,MasaCtrl,Cross-Image-Attention,PhotoSwap} as the evaluated methods of above combinations and present the results in \figurename~\ref{fig:exp1-1}.

\noindent$\bullet$
\textbf{Blending-Based Editing}. Methods based on $F_{edit}^{Blend}$ represent source image and editing targets in the same space for blending, achieving both content consistency and semantic fidelity. Specifically, spatial space blending assembles the contents from different sources into a single image harmoniously, while semantic space blending enables highly flexible editing to address various tasks under the same paradigm. (\romannumeral 1). $F_{inv}^{T}+F_{edit}^{Blend}$. For content-aware editing tasks, works from~\cite{Imagic,Forgedit} use inversion-based algorithms to embed source image into textual space, where the textural embedding is used to blend with target embedding. (\romannumeral 2). $F_{inv}^{F}+F_{edit}^{Blend}$. Some studies~\cite{Blended-Latent-Diffusion,High-Resolution-Image-Editing,NTI,MasaCtrl} blend diffusion features to achieve local editing, such as avoidance of over-editing or inpainting. (\romannumeral 3). $F_{inv}^{T}+F_{inv}^{F}+F_{edit}^{Blend}$. Work from ~\cite{DreamEdit} combines tuning-based and forward-based approaches to insert customized object into the specified location of source image. We select ~\cite{Forgedit,NTI,Blended-Latent-Diffusion,DreamEdit} as the evaluated methods of above combinations, and present the results in \figurename~\ref{fig:exp1-2}.

\noindent$\bullet$
\textbf{Score-Based Editing}. Methods based on $F_{edit}^{Score}$ guide the editing process by maximizing the posterior probability of each editing target. The posterior distributions could be modeled using noise estimates from distinct conditions or experts, or through manually constructed energy functions. (\romannumeral 1). $F_{inv}^{T}+F_{edit}^{Score}$. Some studies~\cite{SINE,DCO} combine the predicted noise from updated and pre-trained models for preserving details in source image and maintaining generative capability. (\romannumeral 2). $F_{inv}^{F}+F_{edit}^{Score}$. Works from~\cite{SEGA,Ledits,Ledits++} combine the predicted noises from different text prompts to achieve multiple subtle adjustments in a single backward process. In addition, some methods~\cite{Self-Guidance,Dragon-Diffusion} construct energy functions to handle various control signals, and achieve fine-grained modification of image layout, object shape, or appearance. We select ~\cite{SINE,Ledits++,Dragon-Diffusion} as the evaluated methods of above combinations, and present the results in \figurename~\ref{fig:exp1-3}.

\noindent$\bullet$
\textbf{Optimization-Based Editing}. Similar to approaches using energy functions, methods based on $F_{edit}^{Optim}$ guide the editing direction by minimizing optimization objectives. (\romannumeral 1). $F_{inv}^{F}+F_{edit}^{Optim}$. In content-aware tasks, works from~\cite{CDS,Region-Aware,Pick-and-Draw} adjust image content through introducing several loss terms for achieving fidelity to both source image and editing targets. (\romannumeral 2). $F_{inv}^{T}+F_{inv}^{F}+F_{edit}^{Optim}$. To preserve details of the source content after local distortion, ~\cite{Drag-Diffusion} employs test-time tuning to avoid instability. We select ~\cite{CDS,Drag-Diffusion} as the evaluated methods of above combinations, and present the results in \figurename~\ref{fig:exp1-4}.

\begin{figure*}[!t]
	\centering
    \includegraphics[width=0.90\linewidth]{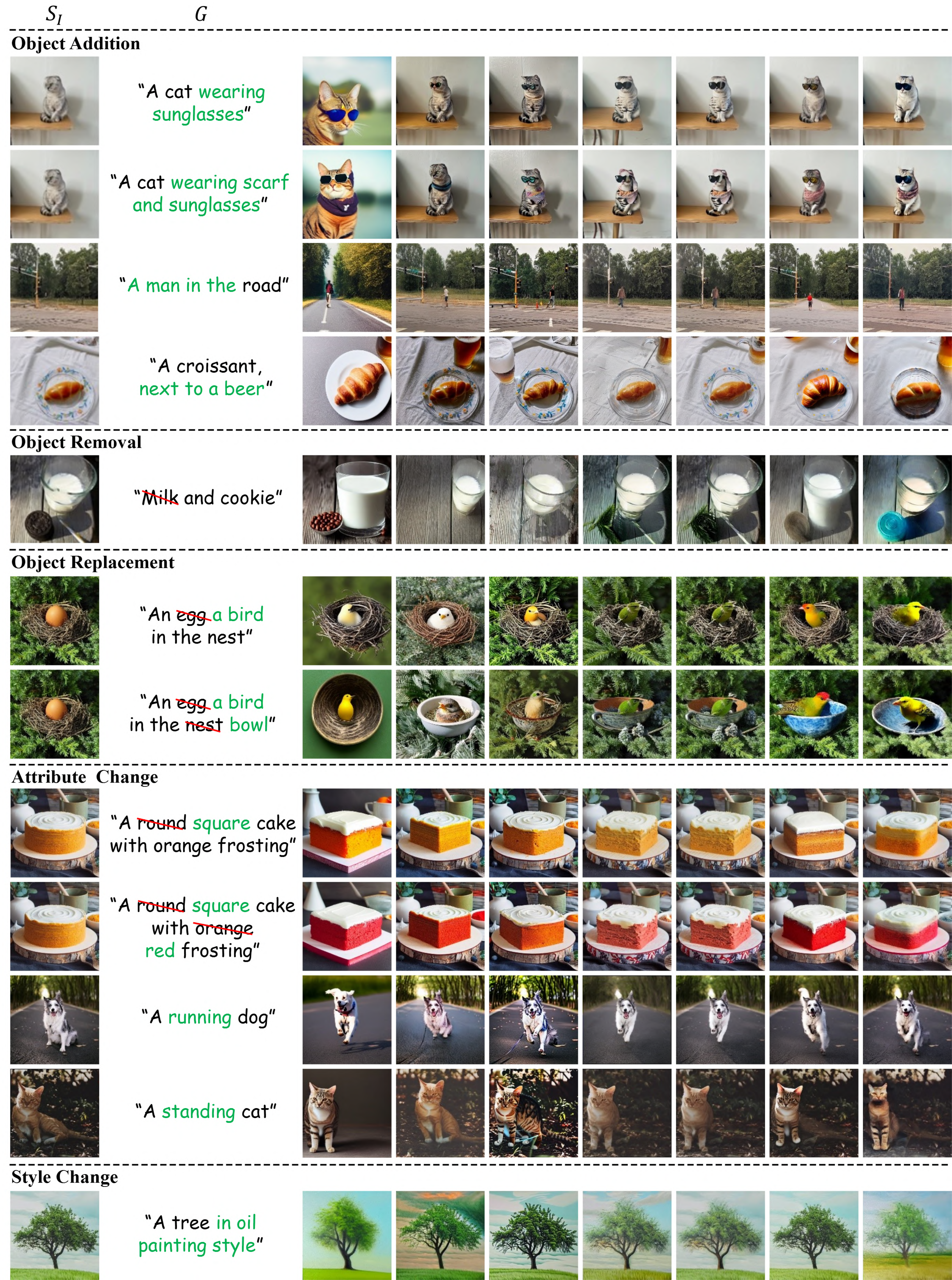}
    \begin{picture}(0,0)
        \scriptsize {
            \put(-306,617){\scalebox{1.5}{\protect \circled{1}}}
            \put(-260,617){\scalebox{1.5}{\protect \circled{2}}}
            \put(-214,617){\scalebox{1.5}{\protect \circled{3}}}
            \put(-168,617){\scalebox{1.5}{\protect \circled{4}}}
            \put(-122,617){\scalebox{1.5}{\protect \circled{5}}}
            \put(-76,617){\scalebox{1.5}{\protect \circled{6}}}
            \put(-30,617){\scalebox{1.5}{\protect \circled{7}}}
        }
    \end{picture}
    
    \caption{\textbf{Evaluation in Content-Aware Editing}. \scalebox{0.85}{\protect \circled{1}~-~\protect \circled{7}} represent \scalebox{0.8}{$F_{inv}^F+F_{edit}^{Norm}$}, \scalebox{0.8}{$F_{inv}^T+F_{edit}^{Blend,se}$}~\cite{Forgedit}, \scalebox{0.8}{$F_{inv}^T+F_{edit}^{Score}$}~\cite{SINE}, \scalebox{0.8}{$F_{inv}^F+F_{edit}^{Attn}$}~\cite{NTI,MasaCtrl}, \scalebox{0.8}{$F_{inv}^F+F_{edit}^{Attn}+F_{edit}^{Blend,sp}$}~\cite{NTI,MasaCtrl}, \scalebox{0.8}{$F_{inv}^F+F_{edit}^{Score}$}~\cite{Ledits++} and \scalebox{0.8}{$F_{inv}^F+F_{edit}^{Optim}$}~\cite{DDS} respectively.}
	\label{fig:exp2}
\end{figure*}
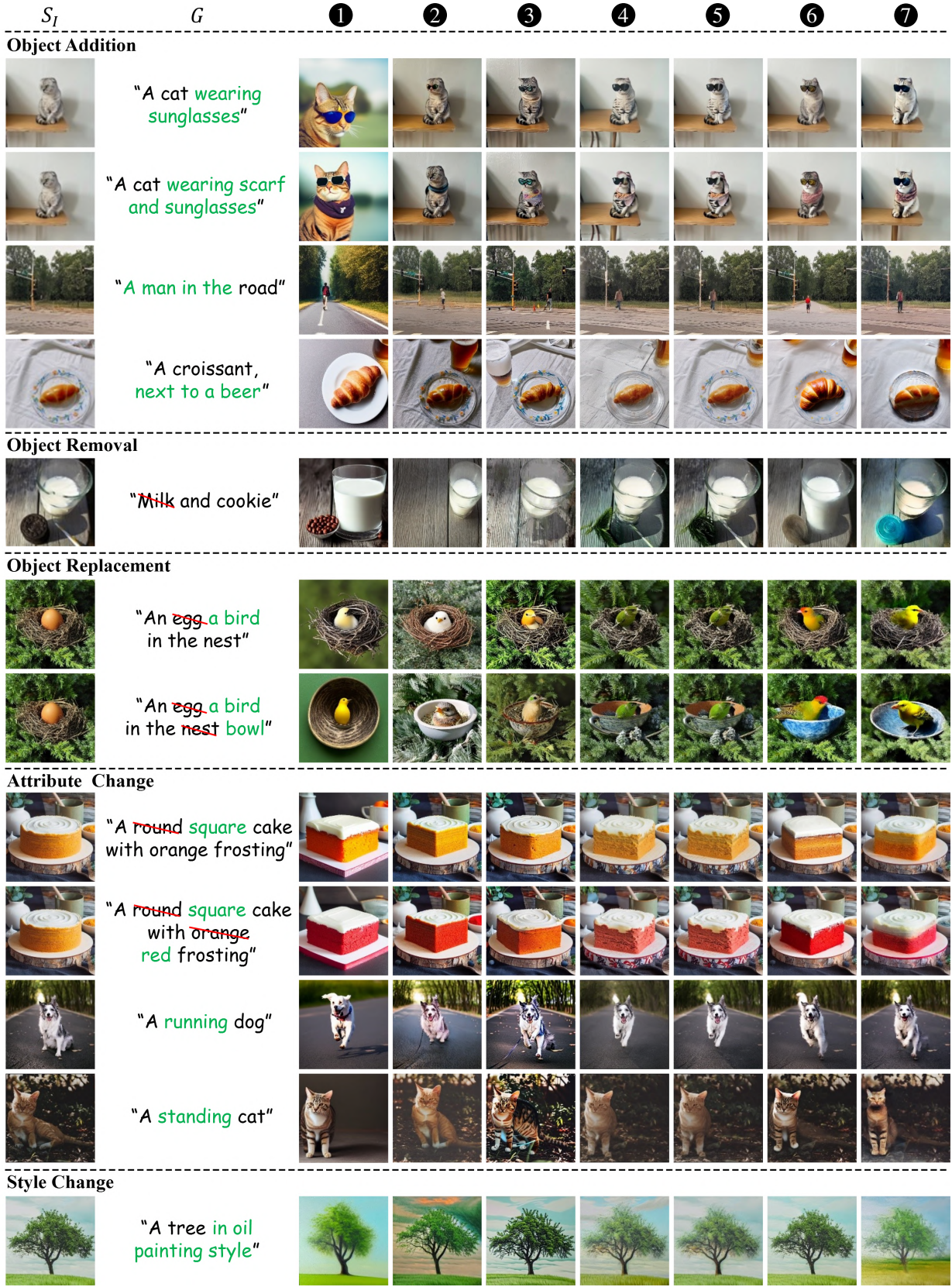

\subsection{Comparison in Text-Driven Editing}
\begin{figure}[!t]
	\centering
    \includegraphics[width=1\linewidth]{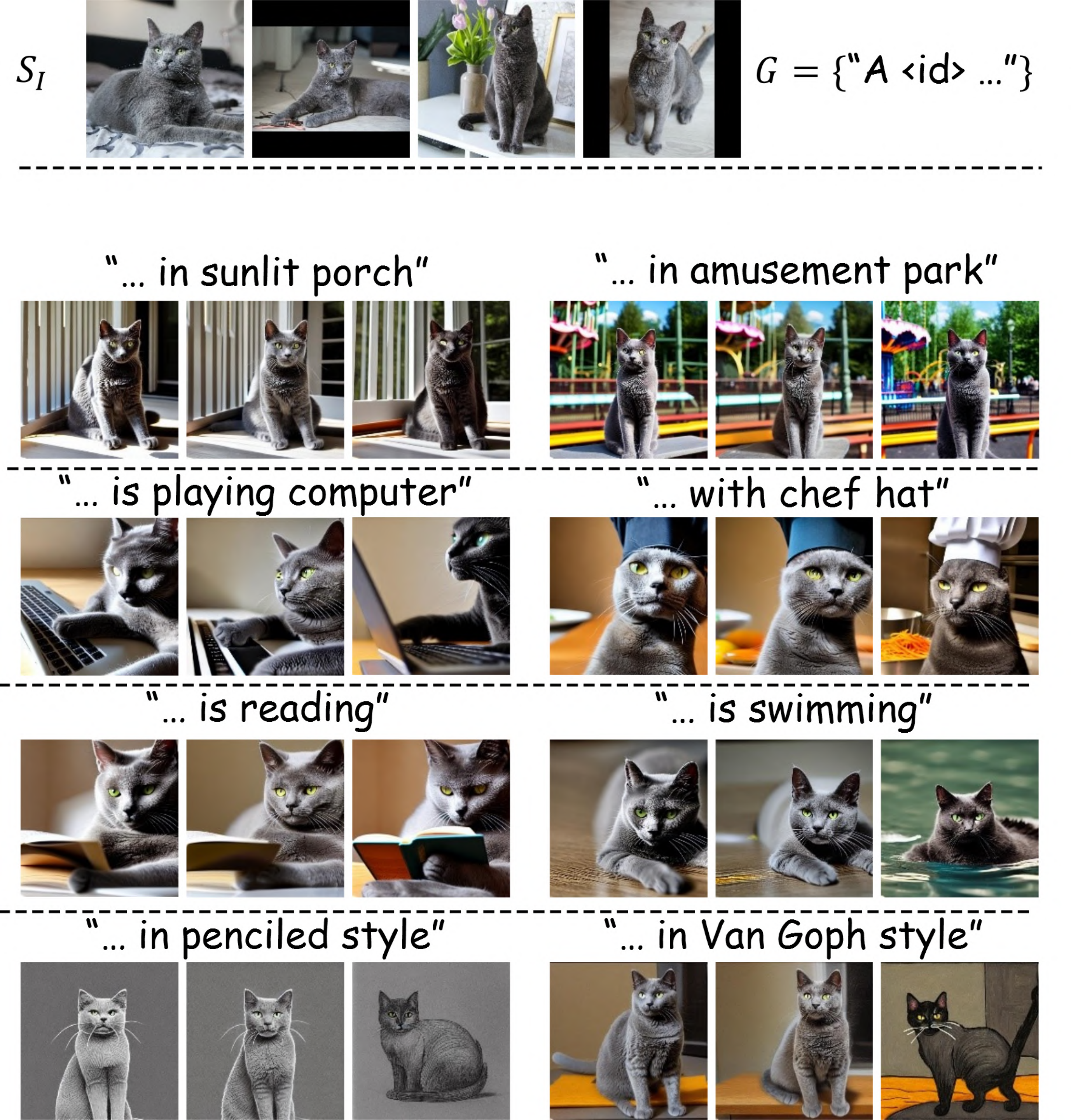}
        \begin{picture}(0,0)
        \scriptsize {
            \put(-106,222){\scalebox{1.2}{\protect \circled{1}}}
            \put(-67,222){\scalebox{1.2}{\protect \circled{2}}}
            \put(-28,222){\scalebox{1.2}{\protect \circled{3}}}

            \put(18,222){\scalebox{1.2}{\protect \circled{1}}}
            \put(57,222){\scalebox{1.2}{\protect \circled{2}}}
            \put(96,222){\scalebox{1.2}{\protect \circled{3}}}

        }
    \end{picture}
    \vspace{-6mm}
    \caption{\textbf{Evaluation in \contentfreebig}. \scalebox{0.85}{\protect \circled{1}~-~\protect \circled{3}} represent \scalebox{0.8}{$F_{inv}^T+F_{edit}^{Norm}$}, \scalebox{0.8}{$F_{inv}^T+F_{edit}^{Attn}$}~\cite{DreamMatcher} and \scalebox{0.8}{$F_{inv}^T+F_{edit}^{Score}$} (our implementation) respectively.}
	\label{fig:exp3}
\end{figure}

Next, we select several text-driven editing scenarios from content-aware and content-free tasks to compare the performance of advanced methods in different combinations. The evaluation data are sourced from ~\cite{DreamBooth,Dragon-Diffusion,Imagic,P2P,Custom-Diffusion}. Specifically, we adjust the pivotal parameters for each method and present their best results in \figurename~\ref{fig:exp2} and \figurename~\ref{fig:exp3}.

\noindent$\bullet$
\textbf{Evaluation in Content-Aware Editing}. For content-aware editing tasks, we mainly consider following scenarios: object manipulation, attribute change, and style change. The evaluation results are shown in \figurename~\ref{fig:exp2}. Specifically, we include some challenging cases, such as multi-target editing (rows 2, 7, 9 in \figurename~\ref{fig:exp2}) and instances with significant modification in image structure (rows 4, 5 in \figurename~\ref{fig:exp2}) to assess the effectiveness of each method. In following, we annotate $F_{edit}^{blend}$ with different superscripts to indicate specific blending space ($sp$, $se$ for spatial and semantic spaces respectively). The chosen combinations and corresponding methods in this experiment are listed as below. (\romannumeral 1). $F_{inv}^T+F_{edit}^{Blend,se}$. Forgedit~\cite{Forgedit}.  (\romannumeral 2). $F_{inv}^T+F_{edit}^{Score}$. SINE~\cite{SINE}. (\romannumeral 3). $F_{inv}^F+F_{edit}^{Attn}$. NTI~\cite{NTI} and MasaCtrl~\cite{MasaCtrl}. Specifically, MasaCtrl is used for evaluating non-rigid editing, as demonstrated in rows 10 and 11 of \figurename~\ref{fig:exp2}. In this setting, we comment out the source code related to spatial space blending. (\romannumeral 4). $F_{inv}^F+F_{edit}^{Attn}+F_{edit}^{Blend,sp}$. NTI and MasaCtrl. In this setting, we maintain the original code of each method. (\romannumeral 5). $F_{inv}^F+F_{edit}^{Score}$. LEDITS++~\cite{Ledits++}. (\romannumeral 6). $F_{inv}^F+F_{edit}^{Optim}$. DDS~\cite{DDS}. The results of $F_{inv}^F+F_{edit}^{Norm}$ are also presented in the first column as the baseline. Next, we elaborate our analysis of the experiment results.
\begin{enumerate}
    \item For content consistency, methods using tuning-based inversion~\cite{SINE,Forgedit} or mutual self-attention~\cite{MasaCtrl} outperform in retaining details of edited object, such as the cat in rows 1 and 2 (column 2, 3), and non-rigid editing case in row 10 (column 2, 4, 5). Among them, Forgedit~\cite{Forgedit} and SINE~\cite{SINE} exhibit a potential risk of overfitting due to the one-shot tuning process (partial examples in rows 10, 11). Moreover, most methods change the visual elements that are not related with editing, like image layout or background. In contrast, NTI~\cite{NTI} outperforms in structure maintenance, while the additional spatial space blending retains background content, as demonstrated in some object addition scenarios (rows 3, 4) and local style change (row 12).
    \item From the perspective of semantic fidelity, the selected methods perform well in most examples. For multi-target editing, LEDITS++~\cite{Ledits++} achieves the goal in a convenient and efficient fashion with the aid of multi-noise guidance. However, Forgedit~\cite{Forgedit} based on semantic space blending exhibits insufficient generation in some cases (row 2). In scenarios with significant layout modification, such as adding or removing the large objects in source image (rows 4, 5), methods using tuning-based inversion approaches~\cite{SINE,Forgedit} demonstrate better flexibility. 
    \item For the adjustment of method-related parameters, DDS~\cite{DDS} outperforms other methods in terms of user-friendliness. Other methods require to seek for optimal parameters~\cite{SINE,Forgedit,Ledits++,NTI,MasaCtrl} or rely on the choice of random seed~\cite{SINE,Forgedit} in some challenging scenarios.
\end{enumerate}

We conduct comprehensive evaluation on advanced methods using our own dataset to quantitatively access their performance across various scenarios. Unlike publicly available datasets~\cite{Imagic,editing-survey,PnP-Inversion}, our dataset includes numerous challenging scenarios, such as multi-target editing and situations with substantial influence on semantic layout. The details of dataset and the quantitative results are presented in supplementary materials.

\noindent$\bullet$
\textbf{Evaluation in \contentfreebig}. For \contentfreesmall tasks, we mainly consider the subject-driven customization. \figurename~\ref{fig:exp3} enumerates multiple scenarios for evaluation, such as change of background, interaction with objects, pose variation and style change. The chosen combinations and corresponding methods in this experiment are listed as following. (\romannumeral 1). $F_{inv}^T+F_{edit}^{Attn}$. DreamMatcher~\cite{DreamMatcher}. (\romannumeral 2).$F_{inv}^T+F_{edit}^{Score}$. Our implementation. Since DCO~\cite{DCO} belonging to this category is based on StableDiffusionXL~\cite{StableDiffusionXL}, we provide a similar solution using StableDiffusion~\cite{StableDiffusion}. Specifically, we employ the pre-trained models fine-tuned with class-prior preservation loss~\cite{DreamBooth} and combine the estimated noises of customization expert and original model like in DCO. We also present the results of $F_{inv}^T+F_{edit}^{Norm}$ in the first and fourth columns as the baseline. We present our analysis of the experiment results as below.
\begin{enumerate}
    \item For content consistency, as shown in \figurename~\ref{fig:exp3}, DreamMatcher~\cite{DreamMatcher} based on mutual self-attention outperforms in preservation of subject details in most cases.
    \item For semantic fidelity, our implementation using multi-expert guidance effectively avoids overfitting in some examples, maintaining the generative capacity of pre-trained T2I model. 
\end{enumerate}

In addityion, we carry out the comprehensive quantitative experiment on above methods with constructed collection of prompt templates, which encompasses rich and diverse scenarios for customization. The experiment details and results are presented in the supplementary materials.

Moreover, we attempt to adapt approaches~\cite{Forgedit,MasaCtrl,Self-Guidance} in content-aware tasks to customization for investigating their potential capabilities of zero-shot customization. However, these methods either fail to achieve significant performance improvement~\cite{Forgedit} or struggle with maintaining subject identity~\cite{MasaCtrl,Self-Guidance}. We encourage researchers to explore this unsolved challenge and improve the performance of existing methods.

\section{Training-Based Image Editing} \label{sec:end-to-end image editing}
Aforementioned works~\cite{PnP,P2P,Cross-Image-Attention,Drag-Diffusion} solve editing tasks in zero-shot or few-shot fashion. Due to the plug-and-play characteristics, these methods are widely studied and applied in many scenarios. Nevertheless, some of these advanced approaches necessitate test-time optimization~\cite{NTI,DreamBooth,TI,DDS} or require users to adjust pivotal parameters~\cite{P2P,MasaCtrl,Edit-Friendly,Imagic}, which impede their applicability. To tackle these challenges, a big family of studies~\cite{IP-Adapter,InstructPix2Pix} intend to train a task-specific model with amount of data, to directly transform the source image to target one under user guidance. Specifically, they can be regarded as the extension of tuning-based approaches~\cite{TI,DreamBooth}, where source images are encoded by newly introduced parameters. Several researchers~\cite{Zone,HD-Painter,FOI,DiffEditor} also combine the pre-trained editing models~\cite{InstructPix2Pix,StableDiffusion} with $F_{edit}$ for better outcome.

\begin{figure}
	\centering
    \includegraphics[width=0.98\linewidth]{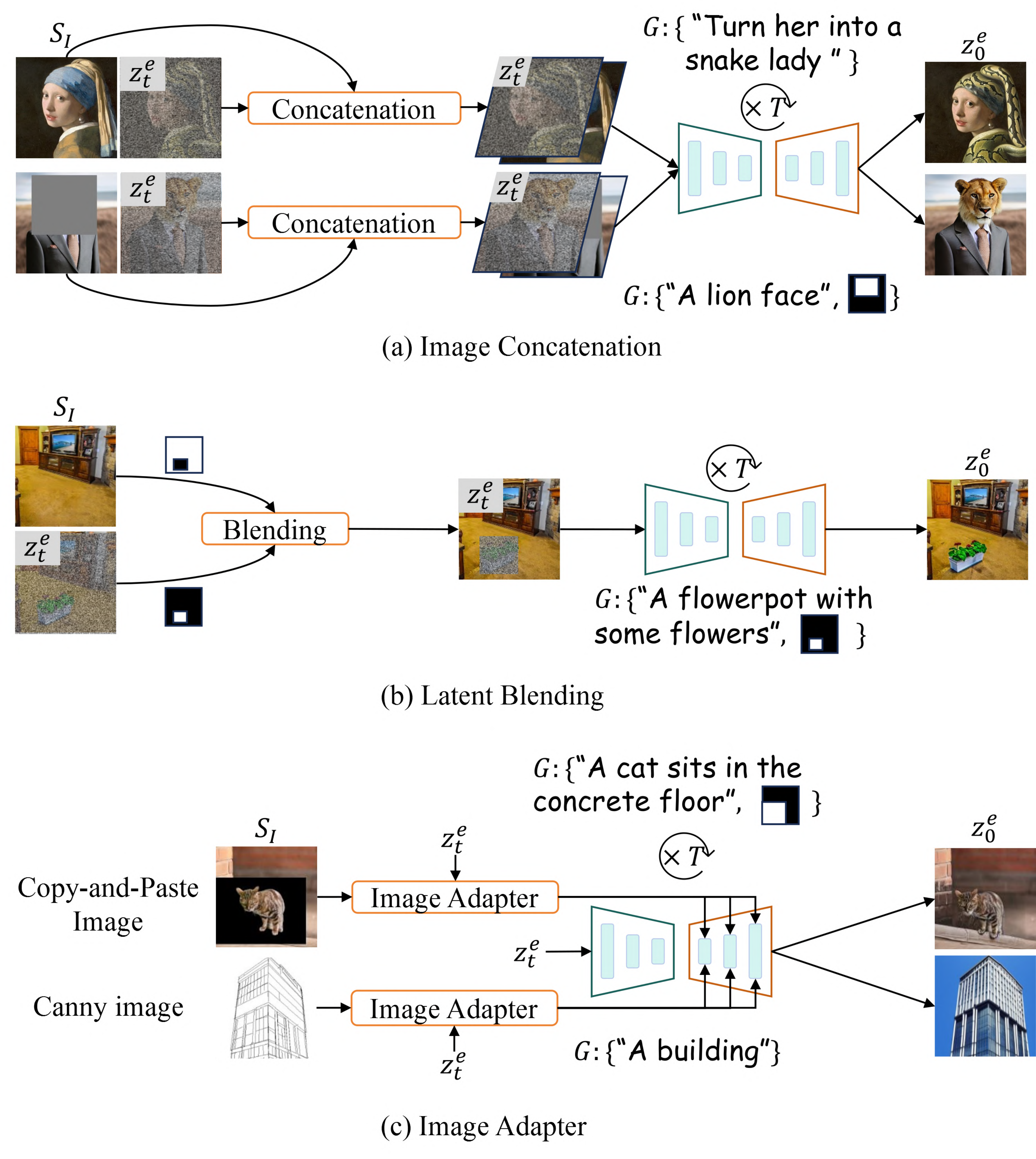}
    \caption{\textbf{\contentawareuppercase injection schemes}. The illustrated methods are~\cite{InstructPix2Pix,Imagen-Editor,ObjectStitch,PHD,ControlNet}. The orange block indicates operation of each scheme. Copy-and-paste image in (c) is obtained by filling the segmented foreground object into the interesting area of background image. The output of image adapter in (c) is used to modulate the features from backbone.}
	\label{fig:content-aware-injection}
\end{figure}

Without the aid of large-scale T2I models~\cite{StableDiffusion,Imagen}, most of early works~\cite{Diffusion-Clip,Diffusion-Autoencoder,Diffstyler,Egsde} are restricted in terms of guidance diversity and generalization. Therefore, we mainly introduce studies building upon pre-trained models. To incorporate source images into base model, these methods have to design the proper injection schemes for adapting different tasks. For existing training-based editing methods~\cite{ControlNet,InstructPix2Pix,Paint-by-Example,ELITE}, we enumerate several commonly used strategies in following and categorize them based on editing tasks. 
\begin{enumerate}
    \item \textbf{\contentawareinjectionbig~Scheme}. There are several injection schemes for \contentawaresmall tasks, where methods require to retain low-level semantic contents. 1. \textbf{Image Concatenation}. Some works~\cite{InstructDiffusion,InstructPix2Pix} concatenate the noisy latent with image condition in each denoising stage, while modifying the weights of input layer to accept additional channels. 2. \textbf{Latent Blending}. Several approaches~\cite{SmartBrush,ObjectStitch,CycleGAN-Turbo} blend source image with generating one in certain denoising steps. 3. \textbf{Image Adapter}. Another group of studies~\cite{ControlNet,UniControl,PHD} introduce auxiliary branched network to process the source image, while modulating features in the main branch. 
    \item \textbf{\contentfreeinjectionbig~Scheme}. For \contentfreesmall tasks, methods require to extract high-level semantics from source images, while overlooking irrelevant contents. Most of these works introduce additional image adapter to process image features, and subsequently integrate the processed features into original model. There are two kinds of adapters in this group. 1. \textbf{\texturalspaceadapterbig}. Some methods~\cite{Paint-by-Example,FastComposer,Blip-Diffusion} project image features to textural space, facilitating the text-driven generation and retaining global semantic.  2. \textbf{\imagespaceadapterbig}. Other works~\cite{ELITE,IP-Adapter} enable the communication between features of source image and generating one through additional modules, which outperform in capturing finer details.
\end{enumerate}
We demonstrate these strategies in \figurename~\ref{fig:content-aware-injection} and \figurename~\ref{fig:content-free-injection} respectively, which showcase the utilization in several representative studies~\cite{InstructPix2Pix,IP-Adapter,ControlNet,Imagen-Editor,ObjectStitch,PHD,Blip-Diffusion}. Significantly, methods are able to exploit different injection schemes in conjunction to adapt challenging tasks~\cite{Anydoor,MGIE,Paint-by-Example,Reference-Based-Composition}. Moreover, due to the wide variety of editing scenarios, these methods have to build task-specific dataset accordingly. We present the injection scheme and data source of reviewed methods in \tablename~\ref{tab:paper-list}, and organize these works based on editing tasks.

\subsection{\contentawarebig} \label{sec:content-aware-editing}
In this section, we introduce several widely studied scenarios in \contentawaresmall tasks.
\subsubsection{Instruction-Based Image Editing} 
Instruction-based editing~\cite{InstructPix2Pix,MagicBrush,SmartBrush} provides a intuitive way for human beings to manipulate images, where users input a command-style text instead of an exhaustive description. Since original T2I models~\cite{StableDiffusion,Imagen,GLIDE} are trained with static texts, they exhibit limited capacity in understanding the instruction. Therefore, it's necessary for researchers~\cite{InstructPix2Pix,MagicBrush,HIVE} to collect sufficient training triplets, which contain instruction, source image, and corresponding edited image. To address the challenge, InstructPix2Pix~\cite{InstructPix2Pix} first optimizes GPT-3~\cite{GPT-3} to infer both instruction and target description from image caption. This step enables the method to leverage zero-shot editing technique~\cite{P2P} to translate the image into edited one based on source and target descriptions. Through employing concatenation scheme illustrated in top of \figurename~\ref{fig:content-aware-injection}, InstructPix2Pix is trained to map source image to target one under instruction guidance. However, the synthesized dataset still contains a certain amount of low-quality samples. In order to tackle the issue, MagicBrush~\cite{MagicBrush} hires technical workers to generate high-quality images through professional platform~\cite{DALLE-2}. Moreover, the method simultaneously considers single-turn and multi-turn editing scenarios, where the later needs workers to iteratively manipulate the edited image from last turn, offering a more challenging setting. Borrowing the idea from RLHF (Reinforcement Learning with Human Feedback)~\cite{InstructGPT}, HIVE~\cite{HIVE} ranks edited images to build auxiliary reward dataset, which is used to train an in-domain reward model. The method then fine-tunes the instruction-based model~\cite{InstructPix2Pix} to maximize reward expectations for better performance.

In addition, some studies~\cite{InstructDiffusion,Emu-Edit} incorporate traditional vision tasks into instruction-based editing framework. InstructDiffusion~\cite{InstructDiffusion} proposes IEIW (Image Editing in the Wile) dataset to unify a variety of tasks, such as segmentation~\cite{VLT,openvocabularysurvey,VLTPAMI}, key point detection, and so on. For example, object detection is drawing a bounding box in proper region. Through constructing instructions and image pairs with off-the-shelf tools~\cite{BLIP-2,Paint-by-Example}, InstructDiffusion is applicable for versatile purposes. Furthermore, to avoid confusion in various editing scenes, Emu-Edit~\cite{Emu-Edit} introduces task embedding as an additional condition. The embedding prevents the method executing incorrect action, such as applying segmentation in object manipulation task. 

Moreover, since CLIP is trained with static texts, it's difficult to precisely interpret command-style sentence. Several methods~\cite{MGIE,SmartEdit} exploit Multimodal Large language model (MLLM) to tackle the issue. MGIE employs LLaVA~\cite{LLAVA} to extract more expressive features from instruction-image pair, which are incorporated into base model through injected attention layers. In addition, SmartEdit~\cite{SmartEdit} considers more challenging scenarios, where the method requires to comprehend complicated instruction and carry out further reasoning. Specifically, it proposes a Bidirectional Interaction Module (BIM) to process image features extracted from LLaVA's visual encoder, which encompasses rich information to infer the precise visual transformation.

\subsubsection{Image Inpainting}
Inpainting is designed to fill the interesting area with meaningful contents, where the modification region is identified by a binary mask. Building upon T2I models, a group of methods~\cite{SmartEdit,Imagen-Editor} complete the missing part under text guidance. Several pre-trained models~\cite{GLIDE,StableDiffusion} naturally support text-driven inpainting. In training time, they randomly erase part of image and predict pixels with global description. However, this training strategy does not closely align the semantics of masked area and text prompt, making it difficult to perform fine-grained and faithful manipulation. To alleviate the problem, Imagen Editor~\cite{Imagen-Editor} proposes a sophisticated dataset, named EditBench. It handcrafts mask conditions in varying coarse-levels as well as corresponding descriptions. Furthermore, SmartBrush~\cite{SmartBrush} employs the precision factor of mask as additional condition, which reflects the degree of alignment between the shapes of mask and inpainted object. Meanwhile, the method simultaneously estimates the mask for preserving pixels in out-of-mask region. In addition, PowerPaint~\cite{PowerPaint} differentiates distinct inpainting scenes and learns the task-related embedding to adapt various scenarios.

Integrating the object from reference image into source image is also a challenging task, which entails methods coherently compositing portions from different inputs. To prevent inpainting in a copy-and-paste manner, Paint-by-Example (PbE)~\cite{Paint-by-Example} extracts global semantic of reference image through CLIP encoder~\cite{CLIP}, and incorporates into the base model through cross-attention layers. ObjectStitch~\cite{ObjectStitch} introduces content adapter to process reference image. Meanwhile, as illustrated in middle of \figurename~\ref{fig:content-aware-injection}, it blends the source image and generating one in each denoising step to maintain the background contents. Moreover, Reference-Based Image Composition (RIC)~\cite{Reference-Based-Composition} further employs sketch information for structural control inside masked area. Instead of using CLIP image encoder, PhD~\cite{PHD} adopts a branched network~\cite{ControlNet} to process copy-and-paste image. According to the bottom part of \figurename~\ref{fig:content-aware-injection}, the network predicts offsets to modulate features in main branch. AnyDoor~\cite{Anydoor} employs self-supervised model~\cite{DINO-V2} to extract expressive features of reference object, while leveraging auxiliary detail extractor~\cite{ControlNet} to retain details.

\subsubsection{Image Translation}
Image translation~\cite{Image-to-Image,CycleGan} purposes to transfer the source image to target domain, like night to daytime, sketch to natural image, \emph{etc}. Several studies~\cite{ControlNet,SCEdit,UniControl,Uni-ControlNet} train with paired images from source and target domains. To incorporate image condition into diffusion model, like edge~\cite{HED}, skeleton~\cite{OpenPose}, depth~\cite{Midas}, and so on, ControlNet~\cite{ControlNet} and it's concurrent work~\cite{T2I-Adapter} introduces additional branched network to modulate the features in backbone model through predicted offsets. Specifically, ControlNet adopts the architecture and pre-trained parameters from native noise estimator~\cite{U-Net} with additional zero-initialized convolution layers, which are used to process source image and adjust the output of branched network. Similarly, SCEdit~\cite{SCEdit} introduces lightweight SC-Tuner for efficient training. Different from these studies, where an unique network is learned for different domains, other works~\cite{UniControl,Cocktail,Uni-ControlNet} handle multiple modalities in a single network. UniControl~\cite{UniControl} employs Mixture-of-Experts (MOEs)~\cite{MOE} architecture to aggregate features from distinct conditions, reflecting properties of each modality in final image. Instead of estimating features offsets, Cocktail~\cite{Cocktail} and Uni-ControlNet~\cite{Uni-ControlNet} modulate the normalized features by adjusting mean and standard deviation.

Above methods require paired data to learn the translation model. Another line of research~\cite{CycleNet,CycleGAN-Turbo} inherits the idea of cycle consistency loss from GAN-based counterpart~\cite{CycleGan} and trains with unpaired images. CycleNet~\cite{CycleNet} introduces cycle consistency regularization to preserve translation-unrelated content, while proposed self regularization ensures the alignment with target distribution. CycleGAN-Turbo~\cite{CycleGAN-Turbo} employs Latent Consistency Model (LDM)~\cite{LCM} to accelerate sampling process and adjusts the injection scheme accordingly, where the noisy image is directly fed into base model for structure consistency. In addition to cycle loss, the method exploits adversarial loss~\cite{GAN} to generate the image belonging to target domain.

\subsection{\contentfreebig} \label{sec:content-free-editing}
\begin{figure}
	\centering
    \includegraphics[width=1.0\linewidth]{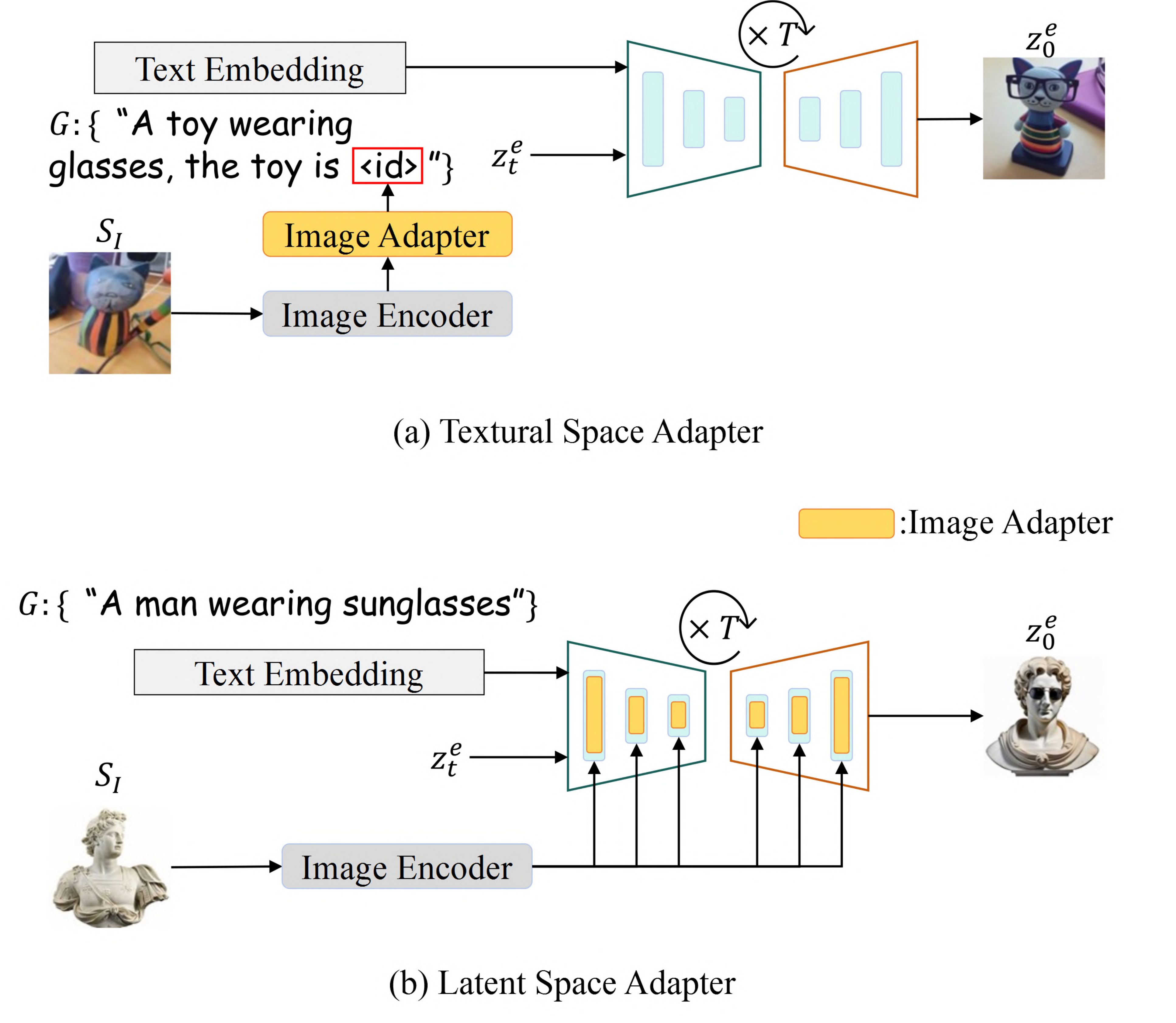}
    \caption{\textbf{\contentfreeuppercase injection schemes}. The illustrated methods are~\cite{Blip-Diffusion,IP-Adapter}. The red box in (a) indicates the insertion of subject embedding into the target embedding. The image adapter in (b) enables the communication of features from generating images and source images. }
	\label{fig:content-free-injection}
\end{figure}

\contentfreeuppercase tasks~\cite{DreamBooth,ELITE,Reversion,pops} intend to maintain the high-level semantic of source images in final result, where the common injection schemes are demonstrated in \figurename~\ref{fig:content-free-injection}. In following, we introduce the advanced methods in subject-driven and attribute-driven customization.

\subsubsection{Subject-Driven Customization}
Subject-driven customization is designed to grasp the identity of the target, and generate novel images to place it in new context. We organize related approaches based on their applicable domains.

\noindent$\bullet$
\textbf{Domain-Aware Customization}. A family of methods focus on personalization of subjects in specific domain~\cite{Taming,InstantBooth,E4T,FastComposer}, like animals~\cite{LSUN}, or human~\cite{CelebA}. Some of works~\cite{Taming,E4T,Enhance-Detail} aim for general class customization. Taming~\cite{Taming} and InstantBooth~\cite{InstantBooth} use CLIP image encoder~\cite{CLIP} to extract identity information, and introduce learnable attention layers to inject visual condition. Adapter in E4T~\cite{E4T} additionally receives noisy latent to predict separate embedding in each denoising step, which is summed with domain-specific word embedding to preserve domain properties. Meanwhile, the method employs regularization term to constraint the magnitude of residual embedding to prevent overfitting. Since regularization term may impairs the identity, ProFusion~\cite{Enhance-Detail} proposes Fusion Sampling to tackle the issue, while keeping generative diversity. Similar to score-based \editingalgorithmsmall methods~\cite{DCO,SINE}, Fusion Sampling intervenes the sampling process with multi-noise guidance for both content consistency and semantic fidelity. 

Another group of works~\cite{FastComposer,Photoverse,InstantID,Face0,Hyper-Dreambooth,DreamIdentity,PhotoMaker} concentrate on human personalization, where identity maintenance is an essential requirement. To address the concept mixing issue in multi-identity customization, which confuses characteristics of different persons, FastComposer~\cite{FastComposer} blends the image embedding and corresponding word embedding of each subject to distinguish different identities. Meanwhile, the method also introduces attention loss~\cite{Break-A-Scene}~\cite{DPL} to make each identifier attend to correct region. In order to mitigate the scarcity of images belonging to the same person in existing datasets~\cite{CelebA}~\cite{CelebA-HQ}, PhotoMaker~\cite{PhotoMaker} builds a novel human dataset through sophisticated pipeline, where the image condition is dissimilar with training target in viewpoints, expressions and so on. The improvement increases the diversity of customized results. In addition, since CLIP~\cite{CLIP} is insufficient for preserving facial details, Photoverse~\cite{Photoverse} processes image embedding through established image adapter. Other methods~\cite{InstantID}~\cite{Face0}~\cite{DreamIdentity} exploit pre-trained face model for extracting more expressive representation of human identity. Among them, InstantID~\cite{InstantID} further introduces auxiliary branched network~\cite{ControlNet} to incorporate expression information (facial keypoints) for maintaining finer details.

\noindent$\bullet$
\textbf{Domain-Agnostic Customization}. Recently, a number of studies~\cite{IP-Adapter,ELITE,Blip-Diffusion,UMM,Re-Imagen,Subject-Driven-Diffusion,Domain-Agnostic,Instruct-Imagen} learn the customization model upon large dataset~\cite{LAION-400M,OpenImage,ImageNet} for domain-agnostic customization. Some of these approaches~\cite{IP-Adapter,ELITE,Blip-Diffusion,Domain-Agnostic,Subject-Driven-Diffusion} explore the personalization of single subject. ELITE~\cite{ELITE} employs global and local mapping networks to encode coarse semantics and object details respectively, where the features are integrated into base model through attention layers. Instead of extracting image features from CLIP~\cite{CLIP}, BLIP-Diffusion~\cite{Blip-Diffusion} exploits the multimodal encoder from BLIP-2~\cite{BLIP-2}, which jointly processes source image and prompt to extract text-aligned features. Based on E4T~\cite{E4T}, Domain-Agnostic~\cite{Domain-Agnostic} enhances the pipeline for open-domain customization. It uses contrastive loss~\cite{Reversion} to make image embedding close to the nearest word embedding, while far from others with different semantics. Meanwhile, the method introduces additional hypernetwork to modulate the weights in attention layers~\cite{Custom-Diffusion} for better retention of details.

Other approaches~\cite{UMM,Subject-Diffusion,Instruct-Imagen} support multi-subject customization. UMM~\cite{UMM} introduces Text-and-Image Unified Encoder (TIUE) to embed text and image into a unified space. For each subject, TIUE replaces corresponding word embedding with subject one, while keeping others unchanged to prevent overfitting issue. Besides, Subject-Diffusion~\cite{Subject-Diffusion} improves the training strategy by taking advantage of various image information. It first processes training data~\cite{OpenImage} through off-the-shelf tools~\cite{SAM,BLIP-2,G-DINO} to get multiple labels, like object box, semantic mask and so on. Through these auxiliary information, the method localizes each subject in source image and exploits existing techniques~\cite{Break-A-Scene,FastComposer,GILGEN} for better supervision, which achieves good balance between consistency and generative diversity. In addition, Instruct-Imagen~\cite{Instruct-Imagen} further supports multiple modalities, such as mask, depth and edge maps, along with instruction guidance for more flexible customization. Specifically, it reuses the backbone network~\cite{Imagen} to extract expressive features from each image-text pair. The method then incorporates these features into base model through newly added attention modules, accomplishing multi-modal customization.

\subsubsection{Attribute-Driven Customization}
Extracting the disentangled attributes from a holistic concept is a challenging task. Similar with the approaches in subject-driven customization, ArtAdapter~\cite{ArtAdapter} leverages image encoder~\cite{VGG} and introduces auxiliary adapter to capture the style from source images. Recently, another line of research~\cite{pops,DreamCreature,Language-Informed} intends to extract the attributes in a finer approach. For example, DreamCreature~\cite{DreamCreature} first employs multi-level hierarchical clustering to distinguish each concept based on DINO features~\cite{DINO-ViT}, which corresponds to a unique word embedding. Then, it fine-tunes with entropy-based attention loss for better disentanglement of different concepts. Work from~\cite{Language-Informed} introduces a set of encoders to extract multiple attributes from image, like material and color \emph{etc}. Furthermore, it leverages BLIP-2~\cite{BLIP-2} to get text predictions under attribute-related question, and aligns the concept embedding with text embedding to alleviate entanglement. Different from these methods, pOps~\cite{pops} combines distinct concepts in CLIP image space, where the assembled embedding is used to condition DALLE-2~\cite{DALLE-2} to generate the final result.

\section{Extension in Video Editing} \label{sec:video}
\begin{figure}[!t]
	\centering
    \includegraphics[width=0.92\linewidth]{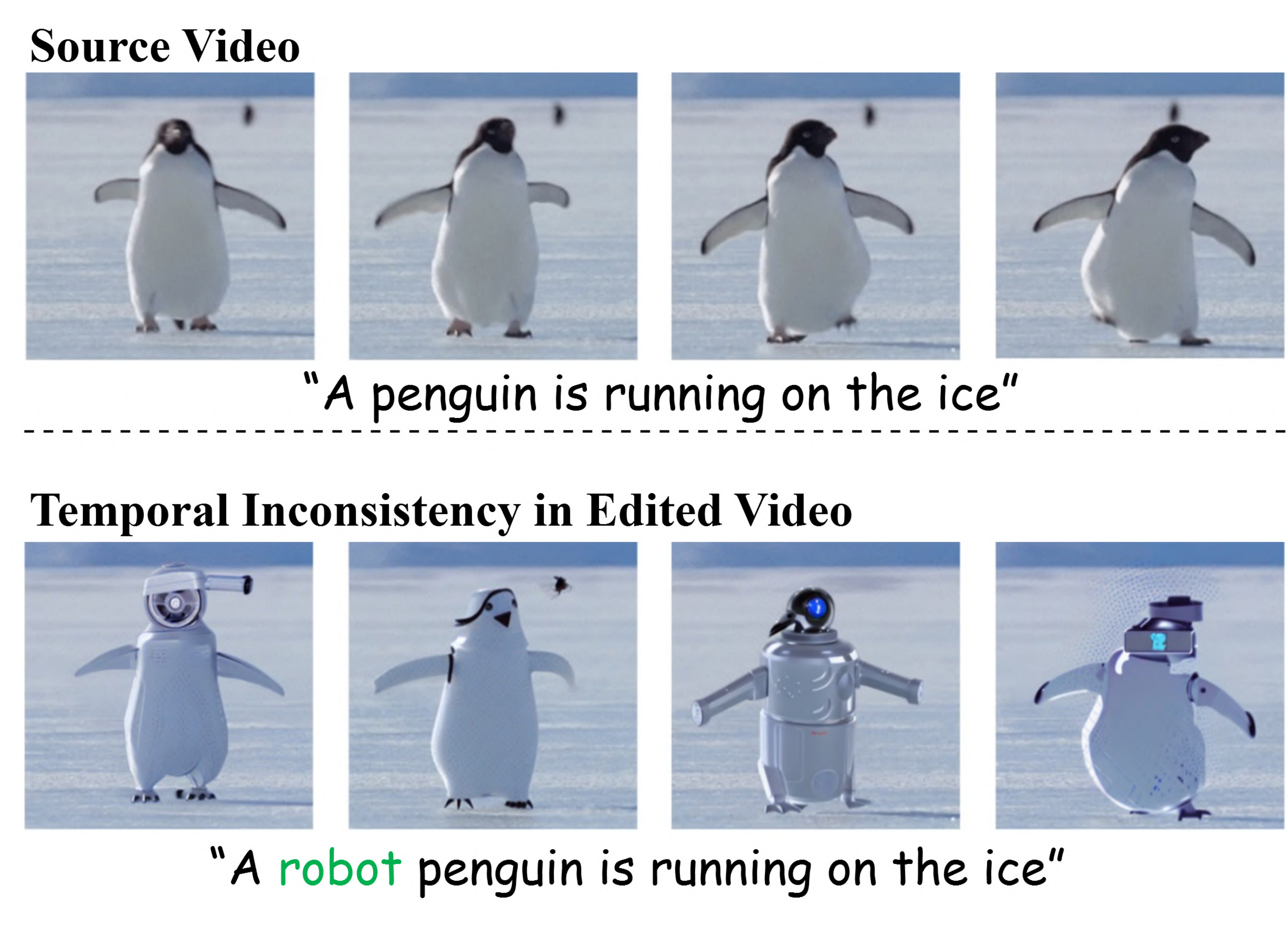}
    \caption{\textbf{Inconsistency caused by frame-by-frame editing}. The example is sourced from~\cite{Video-P2P}. }
	\label{fig:time-inconsistency}
\end{figure}

In this section, we briefly talk about the applications of image editing techniques in video domain. Since text-to-image models~\cite{StableDiffusion,Imagen,StableDiffusionXL} are trained with static images, simply exploiting advanced algorithm~\cite{MasaCtrl,P2P} to edit each frame independently often causes incoherent results, as evident in \figurename~\ref{fig:time-inconsistency}. In order to tackle the temporal inconsistency issue, some works leverage prior knowledge from pre-trained video diffusion models (VDMs)~\cite{Imagen-Video,Dreamix} or train a video editing model from scratch~\cite{Gen-1,FlowVid}. In contrast, due to the scarcity of available data, other works~\cite{TokenFlow,Text2Video-Zero,VideoControlNet,ControlVideo,Ground-A-Video} accommodate 2D algorithms to video in zero-shot or one-shot manner. In this section, we discuss several solutions for improvement of temporal coherency.

\subsection{Utilization of Motion Prior} \label{sec:motion-prior}
Some methods~\cite{Dreamix,AnyV2V} employ pre-trained VDMs for motion guidance. Dreamix~\cite{Dreamix} feeds the downscaled noisy video into cascaded VDM~\cite{Imagen-Video} along with text prompt, resulting in the edited video with temporal consistency. AnyV2V edits the first frame, and generate the whole video through image-to-video model~\cite{I2VGen-XL}. Other works~\cite{Gen-1,FlowVid} instead train an editing model from scratch. To avoid training with scarce text-video pairs, Gen-1~\cite{Gen-1} learns the model based on image conditions. Specifically, the method randomly chooses a middle frame as content guidance, while utilizing estimated depth maps~\cite{Midas} as geometry reference to enhance appearance and structure coherence respectively. During inference, Gen-1 maps the text embedding to image features through prior model to further support text guidance. FlowVid~\cite{FlowVid} accomplishes editing in an inpainting fashion. It warps the first frame to propagate pixels along time dimension through pre-trained optical flow model~\cite{UniMatch}. The method then learns to fill the meaningless region in warped video. During inference, FlowVid only edits the first frame, and refines warped video to get desired outcome.

\subsection{Cross-Frame Attention}\label{sec:cross-frame}
As discussed in Section~\ref{sec:editing-algorithm}, mutual self-attention mechanism builds correspondence among image features from different sources. Inspired by the idea, some studies~\cite{FateZero,Video-P2P,Tune-a-Video,Edit-A-Video} introduce spatial-temporal attention (ST-Attn) modules to replace original self-attention modules for cross-frame attention. Specifically, ST-Attn queries in reference frames for temporal coherency. However, the simple replacement is insufficient to obtain desired results. Therefore, a group of works~\cite{FateZero,Tune-a-Video,Edit-A-Video} boost the performance in following approaches.

\noindent$\bullet$
\textbf{One-Shot Tuning}. Since images from source video may not conform to the input distribution of T2I models, simply employing ST-Attn blocks is insufficient for temporal consistency. To tackle the issue, several works~\cite{Video-P2P,Tune-a-Video,Edit-A-Video} fine-tune original parameters to fit the video data in a one-shot manner. Tune-A-Video (TAV)~\cite{Tune-a-Video} optimizes ST-Attn modules and additional temporal self-attention layers (T-Attn) to fit source video. After tuning, the method refines temporal correspondence across frames. Following works~\cite{Video-P2P,Edit-A-Video} inherit the idea from TAV, while further optimizing ``null-text'' embedding like in NTI~\cite{NTI} to correct reconstruction error under high guidance scale\cite{classifier-free}.

\noindent$\bullet$
\textbf{Local Editing}. Ordinary ST-Attn module changes the content in unrelated region. To address the problem, some works~\cite{FateZero,Video-P2P,Edit-A-Video} inherit 2D techniques~\cite{P2P,MasaCtrl,ProxEdit} to retain irrelevant pixels in source video. Borrowing the idea from~\cite{P2P}, Edit-A-Video (EAV)~\cite{Edit-A-Video} infers the binary mask for preserving background content. Furthermore, it proposes Temporal Consistent Blending (TC Blending) to enhance temporal smoothness, which fuses masks estimated in first and former frames through cross-frame attention map. Moreover, Video-P2P\cite{Video-P2P} and it's concurrent work~\cite{Vid2Vid-Zero} further apply attention injection scheme in P2P~\cite{P2P} for finer modification. In addition, FateZero~\cite{FateZero} merges cross-frame attention maps from source and generating videos for both structure and appearance consistencies.

\subsection{Propagation of Image Information}\label{sec:propagation}
Another technique family~\cite{FLATTEN,TokenFlow,Text2LIVE,StableVideo,Rerender-A-Video,Shape-Aware-NLA,VidToMe} edits several key frames and propagate the pixel / feature information to other frames based on the inter-frame correspondence, which is identified through off-the-shelf tools~\cite{FLATTEN,Text2LIVE,Rerender-A-Video}, or feature properties~\cite{TokenFlow,Fairy}. We organize these studies based on their propagation spaces.

\noindent$\bullet$
\textbf{Propagation in Image Space}. A group of studies~\cite{Text2LIVE,Shape-Aware-NLA,Rerender-A-Video} track pixels through pre-trained models~\cite{GMFlow,NLA} and propagate edited pixels from key frames to the whole video, achieving temporal consistency. One of promising research lines~\cite{Text2LIVE,StableVideo,VidEdit} is tracking associated pixels in canonical space through Neural Layered Atlas (NLA)~\cite{NLA}. Specifically, NLA is trained on video dataset to estimate 2D atlases for each object and background, which serve as canonical representations throughout the video. It correlates pixel located at $(x,y,t)$ and corresponding pixel at $(u,v)$ in layered atlases through learned mapping networks. Here, $(x,y,t)$ and $(u,v)$ are coordinates in video and atlas spaces respectively. Based on NLA, Text2LIVE~\cite{Text2LIVE} first pre-trains a editing model for specific target with semantic and structure losses~\cite{CLIP,Stylegan-Nada}. It then edits atlas images through pre-trained model and propagates pixels to each frame based on mapping relationship. Furthermore, VidEdit~\cite{VidEdit} adopts a more subtle approach. Instead of editing the whole image, it crops the object in foreground atlas using pre-trained segmentation model~\cite{Mask-Transformer,CCL} and edits the small patch with auxiliary structure guidance~\cite{ControlNet,HED}. Through replacing the edited patch in it's original location and performing propagation, the method obtains more consistent outcome. Different from these works, StableVideo~\cite{StableVideo} edits the video frame by frame to leverage various viewpoint information from previous frames. Specifically, the method propagates pixels in previous edited image to current frame through forward and backward mapping operations. To complete the region which is missing in previous frames, StableVideo refines the image following 2D technique~\cite{SDEdit}, which inverts the image back to noise space and improves the quality through backward process. 

Fixed mapping relationship~\cite{NLA} hinders shape-aware editing, like modifying the contour of foreground object. To alleviate the issue, Shape-aware NLA~\cite{Shape-Aware-NLA} first performs shape modification on the key frame, which is fed to semantic correspondence model~\cite{Probabilistic-Warp-Consistency} to identify the deformation of each pixel. Through deformation field and original mapping network, the method corrects the mapping between video and atlas spaces, achieving more consistent result under shape variation. From another perspective, DiffusionAtlas~\cite{DiffusionAtlas} trains additional network with several objectives~\cite{DreamFusion,Shape-Aware-NLA} to refine the mapping relationship. 

\noindent$\bullet$
\textbf{Propagation in Feature Space}. The sampling process in diffusion model allows methods to rectify undesired content step by step. Therefore, some methods~\cite{FLATTEN,TokenFlow,VidToMe,Fairy} propagate diffusion features in each denoising step. TokenFlow~\cite{TokenFlow} captures motion dynamic through calculating the similarity of features from self-attention modules. It first edits key frames simultaneously based on cross-frame attention, and then propagates their image tokens through established inter-frame correspondence. Specifically, for each patch in intermediate images, the method fuses associated tokens in neighbour key frames, where the blending factor depends on temporal distance. In addition, Fairy~\cite{Fairy} argues that the cross-frame attention implicitly tracks relevant features. Therefore, it chooses several reference images from video and edits the video through anchor-based cross-frame attention, where key and value are sourced from reference frames. Inspired from Token Merging (ToMe)~\cite{ToMe}, VidToMe~\cite{VidToMe} uses bipartite soft matching algorithm to construct inter-frame correspondence, and propagates image tokens across frames through merging and unmerging operations. From another perspective, FLATTEN\cite{FLATTEN} employs pre-trained optical flow model\cite{RAFT} to build motion trajectory for each pixel. Therefore, image tokens only attend to relevant ones in the same path for cross-frame attention, mitigating the negative effect from uncorrelated patches.

\section{Future Direction} \label{sec:future}
Recent studies~\cite{MasaCtrl,Anydoor,P2P,Custom-Diffusion,IP-Adapter} have made tremendous progress in multimodal-guided image editing. However, due to the diversity and complexity of editing tasks, undesired results still persist in these methods. In this section, we talk about several unresolved issues and challenges in existing methods, suggesting possible future directions for researchers.

\noindent$\bullet$
\textbf{Challenges in \contentawarebig}. For \contentawaresmall tasks, existing methods~\cite{P2P,MasaCtrl,Blended-Latent-Diffusion,Dragon-Diffusion} are unable to handle with multiple editing scenarios and control signals. This limitation forces applications to switch the proper back-end algorithm for different tasks. Moreover, some advanced methods are unfriendly in terms of ease of use. Several approaches~\cite{P2P,Edit-Friendly,Imagic} require users adjusting pivotal parameters to get optimal results, while others need cumbersome inputs, like source and target prompts~\cite{ProxEdit,NTI}, or auxiliary masks~\cite{Drag-Diffusion,Dragon-Diffusion}.

\noindent$\bullet$
\textbf{Challenges in \contentfreebig}. For \contentfreesmall tasks, existing methods~\cite{DreamBooth,Custom-Diffusion,Reversion} struggle with lengthy test-time tuning process~\cite{TI,NeTI} and overfitting issues~\cite{Language-Drift1,DreamBooth}. Several works aim to alleviate the problem through optimizing small number of parameters~\cite{TI,Custom-Diffusion,Lora-Image} or training a customization model from scratch~\cite{ELITE,Hyper-Dreambooth}. However, they often loss finer details of personalized subject or exhibit inferior generalization ability. Besides, current methods also fall short in extracting abstract concepts from few images. Concept Decomposition~\cite{Concept-Decomposition} learns multi-level attributes of the subject, but requires cumbersome tuning process. Other approaches introduce contrastive loss~\cite{Reversion,Lego} or identify concept-related channels~\cite{ADI} to steer the gradient descent. However, they can not completely decouple the desired concept from other visual elements.

\noindent$\bullet$
\textbf{Challenges in Training-Based Image Editing}. For specific editing scenarios, some training-based methods~\cite{InstructPix2Pix,InstructDiffusion,Paint-by-Example,ControlNet} generate available training data automatically with the aid of off-the-shelf tools~\cite{P2P,SAM,BLIP-2} to alleviate laborious hand-crafted collection. Nevertheless, low-quality and incorrect samples still persist in synthetic dataset, which impair the model performance. In addition to the complex construction process of training data, since methods require to modify the original architecture for accommodating task-specific properties, it's difficult for building a versatile model to achieve multiple editing goals.

\noindent$\bullet$
\textbf{Challenges in Video~\!\&~\!3D Editing}. Image editing techniques provide valuable solutions for video~\cite{Text2LIVE,FateZero,AnyV2V,TokenFlow} and 3D~\cite{InstructNeRF2NeRF,GaussianEditor} arenas. However, due to the scarcity of available high-quality dataset in these realms, temporal or multi-view inconsistency still exists in existing works.

\section{Conclusion} \label{sec:conclusion}
In this work, we survey over three hundred papers in field of multimodal-guided image editing based on text-to-image diffusion models. First, we define the scope of editing from a more generalized perspective, including numerous uninvolved methods in previous literature. We then provide an exhaustive categorization of user guidance and editing scenarios. Subsequently, the proposed unified framework integrates existing state-of-the-art methods, which consists of two algorithm families. This framework allows users to select optimal inversion and \editingalgorithmsmall algorithms according to different purposes. In addition, we extend our discussion to video editing, introducing several major solutions for temporal inconsistency. Finally, we present open challenges in the field and suggest potential future research directions. Our work provides a comprehensive analysis of current approaches, and investigates the essence and inherent logic of editing methods from a novel perspective, filling a critical gap in the research area.

%

\ifCLASSOPTIONcaptionsoff
  \newpage
\fi

{\small
\bibliographystyle{IEEEtran}
\bibliography{egbib}
}

\end{document}